\documentclass[10pt,twocolumn,letterpaper]{article}

\usepackage{cvpr}              
\usepackage{comment}

\usepackage{url}
\usepackage{wrapfig}
\usepackage{graphicx}
\usepackage{multirow}
\usepackage{siunitx}
\usepackage{booktabs}
\usepackage{adjustbox}
\usepackage{tabularx}
\usepackage{colortbl}
\usepackage{xcolor}
\usepackage{float}

\definecolor{best}{rgb}{0.96, 0.57, 0.58}
\definecolor{second}{rgb}{0.98, 0.78, 0.57}
\definecolor{third}{rgb}{1.0, 1.0, 0.56}

\usepackage{xcolor}
\usepackage{pifont}

\newcommand{\xmark}{\textcolor{red}{\ding{55}}}

\usepackage{algorithm}
\usepackage{amsmath}
\usepackage{algpseudocode}
\usepackage{setspace} 
\usepackage{makecell}

\definecolor{cvprblue}{rgb}{0.21,0.49,0.74}
\usepackage[pagebackref,breaklinks,colorlinks,allcolors=cvprblue]{hyperref}

\def\paperID{***} 
\def\confName{CVPR}
\def\confYear{2026}
 
\title{Stereo World Model: Camera-Guided Stereo Video Generation}

\author{
Yang-Tian Sun\textsuperscript{1}
\quad
Zehuan Huang\textsuperscript{2$*$}
\quad
Yifan Niu\textsuperscript{2}
\quad
Lin Ma\textsuperscript{3}
\\
Yan-Pei Cao\textsuperscript{2}
\quad
Yuewen Ma\textsuperscript{3}
\quad
Xiaojuan Qi\textsuperscript{1$\dagger$}
\\[0.5em]
\textsuperscript{1} The University of Hong Kong
\quad
\textsuperscript{2} VAST
\quad
\textsuperscript{3} ByteDance Pico
}

\begin{document}


\twocolumn[{%
\maketitle
\vspace{-3em}
\begin{center}
    \centering
    \captionsetup{type=figure}
    \includegraphics[width=1\linewidth, trim=0 0 0 0, clip]{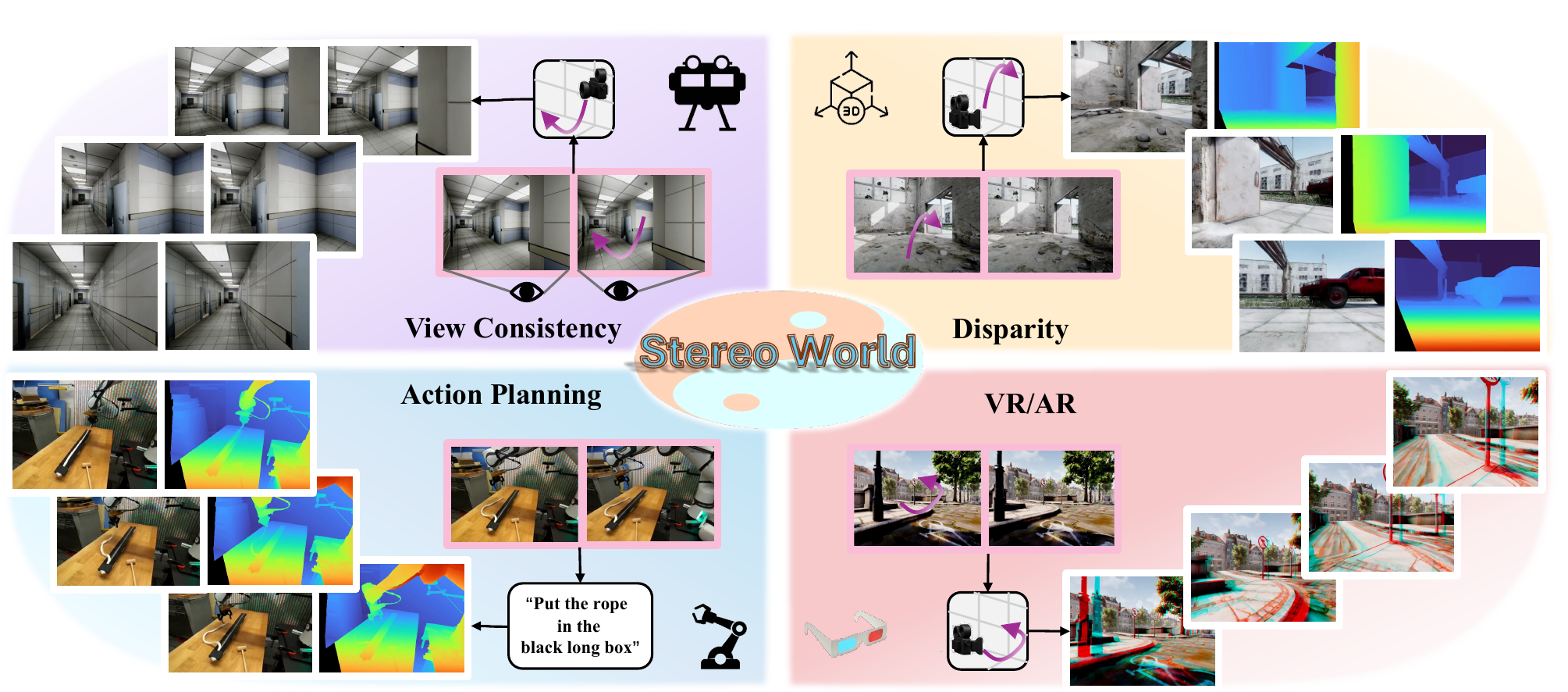}
    \vspace{-3mm}
    \captionof{figure}{We introduce \textbf{StereoWorld}, a stereo world model capable of performing exploration based on given binocular images, generating view-consistent stereo videos with intrinsic geometric understanding. StereoWorld can be applied to downstream tasks like VR/AR visualization as well as action planning in embodied intelligence.  Project: \url{https://sunyangtian.github.io/StereoWorld-web/}.}
\label{fig:teaser}
\end{center}
\vspace{1em}
}]

\if TT\insert\footins{\noindent\footnotesize{
* Project Lead,  $\dagger$Corresponding Author \\
}}\fi

\begin{abstract} 
We present \textbf{StereoWorld}, a camera-conditioned stereo world model that jointly learns appearance and binocular geometry for end-to-end stereo video generation.Unlike monocular RGB or RGBD approaches, StereoWorld operates exclusively within the RGB modality, while simultaneously grounding geometry directly from disparity. 
To efficiently achieve consistent stereo generation, our approach introduces two key designs: (1) a unified camera-frame RoPE that augments latent tokens with camera-aware rotary positional encoding, enabling relative, view- and time-consistent conditioning while preserving pretrained video priors via a stable attention initialization; and (2) a stereo-aware attention decomposition that factors full 4D attention into 3D intra-view attention plus horizontal row attention, leveraging the epipolar prior to capture disparity-aligned correspondences with substantially lower compute. Across benchmarks, \textbf{StereoWorld} improves stereo consistency, disparity accuracy, and camera-motion fidelity over strong  monocular-then-convert pipelines, achieving more than 3$\times$ faster generation with an additional $5\%$ gain in viewpoint consistency. Beyond benchmarks, StereoWorld enables end-to-end binocular VR rendering without depth estimation or inpainting, enhances embodied policy learning through metric-scale depth grounding, and is compatible with long-video distillation for extended interactive stereo synthesis.
\end{abstract}    
\vspace{-6mm}
\section{Introduction}

Learning a generative world model-- \ie, predicting future observations conditioned on actions and camera motion-- has become increasingly important for interactive perception and embodied intelligence. Modern world models~\cite{Team2025AetherGU, Zhou2025StableVC, sun2024dimensionx, müller2025gen3c3dinformedworldconsistent} predominantly use monocular video representations and achieve strong results in controllable video synthesis. Yet monocular observations impose fundamental geometric limits: depth is implicit, scale is ambiguous, and geometric consistency must be inferred rather than observed, which accumulates 3D errors under long-horizon camera trajectories and constrains applications where accurate geometry is critical (e.g., embodied intelligence and navigation). RGB-D world models~\cite{chen2025deepverse, Huang2025VoyagerLA} introduce an auxiliary depth channel, but predicted depth is scene-dependent and still scale-ambiguous, often requiring ad-hoc normalization and remaining unstable across domains~\cite{guizilini2023towards}. 

\begin{figure}[t]
    \centering
    \includegraphics[width=1\linewidth, trim=0 0 0 0,clip]{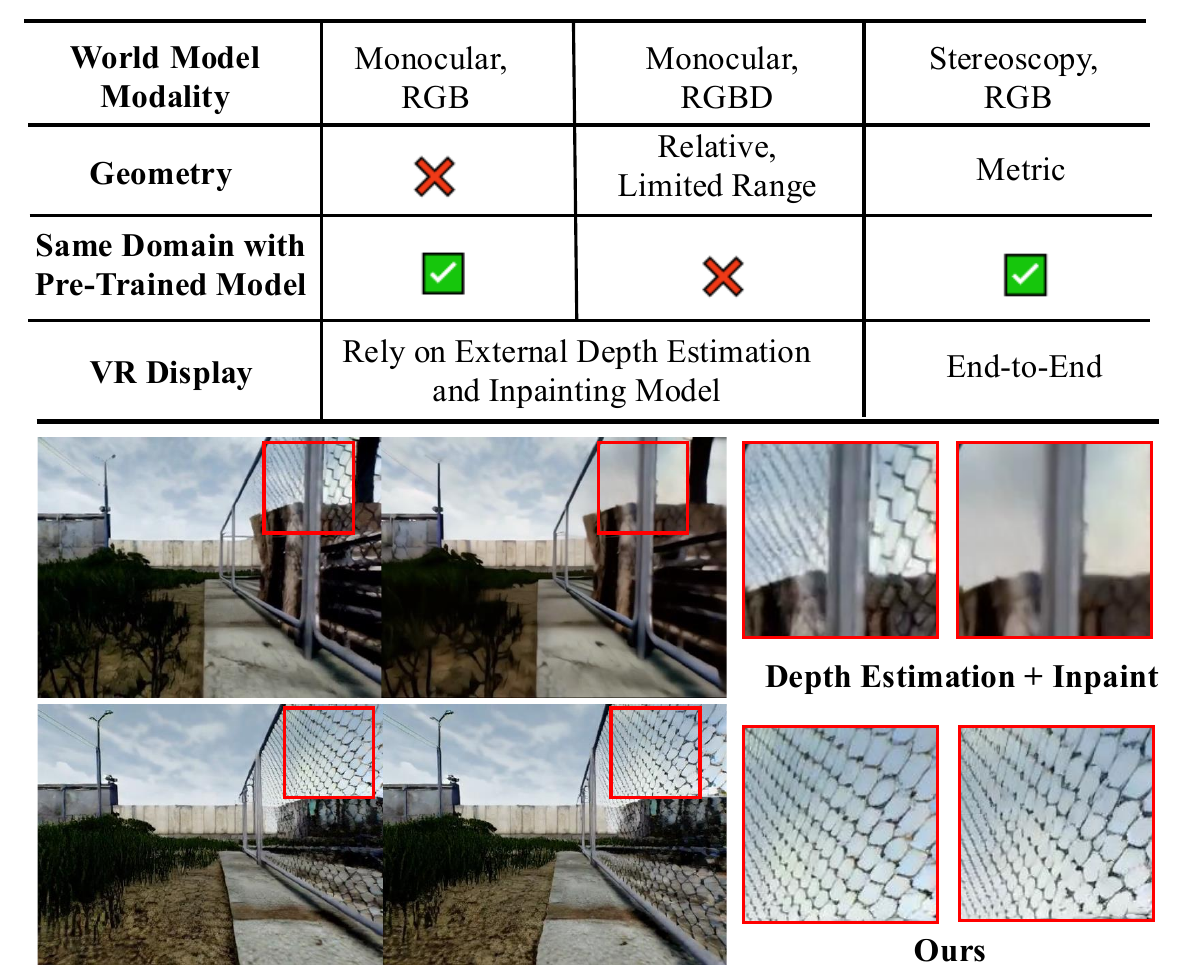}
    \vspace{-3mm}
    \caption{World Model Comparison. StereoWorld incorporates metric-scale geometry, producing output modalities that are more compatible with pretrained models. Moreover, it can be applied end-to-end for VR visualization, ensuring better consistency of fine-grained details between the left and right views.}
    \vspace{-5mm}
    \label{fig:world_model_comparison}
\end{figure}

In contrast, stereo vision -- the dominant perceptual mechanism in many biological systems~\cite{howard2012perceiving, nityananda2017stereopsis}-- provides direct, robust geometric cues to 3D scene structure. This motivates us to study a \textbf{stereo world model} that grounds geometry in binocular observations rather than inferring depth from monocular motion or relying on imperfect depth predictors ({see Fig.~\ref{fig:disp_comparison}}).
Compared to monocular world models, a stereo-conditioned system jointly learns the coupled evolution of appearance and geometry under camera motion and actions; compared to RGB-D systems, it avoids producing and stabilizing explicit metric depth maps while retaining strong geometric signals. The result is consistent, metric-scale perception well suited to VR/AR rendering and embodied navigation, as illustrated in Fig.~\ref{fig:world_model_comparison}.

Building a \textbf{stereo world model} remains non-trivial. First, the predictions must remain consistent across both binocular views and time while generalizing over varying intrinsics, extrinsics, and baselines-- calling for a unified, view- and time-aware camera embedding. Ray-map concatenation~\cite{gao2024cat3d, Team2025AetherGU} encodes absolute coordinates tied to a specific frame, which can entangle viewpoint and scene layout and make relative cross-view generalization (across changing baselines or poses) harder; a relative camera formulation is preferable. 
Second, naive stereo extensions of monocular transformers incur prohibitive compute: self-attention scales quadratically with tokens, and full 4D spatiotemporal cross-view attention quickly becomes infeasible. Third, pretrained video diffusion backbones are highly sensitive to positional-encoding changes, so injecting view-control signals risks wiping out learned priors. 

To address these challenges, we introduce \textbf{StereoWorld}, the first camera-conditioned stereo world model. Our approach is built around two key designs. First, we propose a \textit{unified camera-frame RoPE strategy} that expands the latent token space and augments it with camera-aware rotary positional encoding, enabling joint reasoning across time and binocular views {while minimally modifying} the pretrained backbone’s RoPE space. This formulation effectively encodes relative camera relationships, naturally supports scenes with varying intrinsics and baselines, while preserves pretrained video priors,  
facilitating stable and efficient adaptation to stereo video modeling. Second, we design a \textit{stereo-aware attention mechanism} that decomposes full 4D attention into 3D intra-view attention plus horizontal row attention, leveraging the epipolar prior that stereo correspondences concentrate along scanlines. This achieves strong stereo consistency while dramatically reducing computation. Together, these components allow \textbf{StereoWorld} to learn appearance and geometry jointly, delivering end-to-end binocular video generation with accurate camera control and disparity-aligned 3D structure.

Experiments demonstrate that \textbf{StereoWorld} delivers significant improvements in stereo consistency (Fig.~\ref{fig:stereo_generation_comparison}), disparity accuracy (Fig.~\ref{fig:disp_comparison}), and camera motion fidelity (Fig.~\ref{fig:camera_traj}) over monocular world models. For instance, compared with the SOTA method augmented by post-hoc stereo conversion, our approach achieves a 3$\times$ improvement in generation speed, while also delivering an approximately 5\% gain in viewpoint consistency (see Tab.~\ref{tab:stereo_video_comparison}).
Beyond benchmarks, \textbf{StereoWorld} unlocks practical applications: (i) direct binocular VR rendering without depth estimation or inpainting pipelines (see Sec.~\ref{sec:vr}); (ii) improved spatial awareness for embodied agents through metric-scale geometry grounding ({see Sec.~\ref{sec:embodiedAI}}); and (iii) compatibility with long-range monocular video generation methods~\cite{yin2025causvid, huang2025selfforcing} via distillation to support extended interactive stereo scene synthesis (see Sec.~\ref{sec:distillation}). To our knowledge, this is the first system to realize end-to-end, camera-conditioned stereo world modeling, opening a path toward geometry-aware generative world representations.
Our contributions are summarized as follows: 
\begin{itemize}
\item  We introduce the first \emph{camera-conditioned stereo world model} that jointly learns appearance and binocular geometry, producing view-consistent stereo videos under explicit camera trajectories or action controls. 
\item We expand latent tokens with a camera-aware rotary positional encoding (without altering the backbone’s original RoPE), enabling \emph{relative}, \emph{unified} conditioning across time and binocular views while preserving pretrained video priors via a stable attention initialization. 
\item We decompose full 4D spatiotemporal attention into \emph{3D intra-view attention} plus \emph{horizontal row attention} for cross-view fusion, leveraging the epipolar prior to cut computation substantially while maintaining disparity-aligned correspondence. 
\item Our approach delivers superior quantitative and qualitative results. It enables \textit{end-to-end} VR rendering with improved viewpoint consistency, provides potential geometry-grounded benefits for embodied policy learning, and extends naturally to long-video generation. 
\end{itemize}

\section{Related Work}

\textbf{Camera-Controlled Video Generation.}
With advances in text-to-video models \citep{svd, videocrafter2, cogvideox, snapvideo, hunyuanvideo}, recent work increasingly explores adding conditional signals for controllable generation \citep{dragnuwa, sparsectrl, makeyourvideo, fu20243dtrajmaster}. Among these, camera-controlled video generation \citep{direct_a_video, vd3d, cami2v, zheng2025vidcraft3} aims to explicitly regulate viewpoints via camera parameters.
Notable methods include AnimateDiff \citep{animatediff}, which uses motion LoRAs \citep{lora} to model camera motion; MotionCtrl \citep{motionctrl}, which injects 6DoF extrinsics into diffusion models; and CameraCtrl \citep{He2024CameraCtrlEC}, which designs a dedicated camera encoder for improved control. CVD \citep{CVD} extends control to multi-sequence settings through cross-video synchronization, while AC3D \citep{ac3d} systematically studies camera motion representations for better visual fidelity. Several training-free methods have also emerged \citep{hou2024training, hu2024motionmaster, ling2024motionclone}, , further broadening the landscape of camera-controllable video synthesis. These methods pave the way for world modeling.

\vspace{0.1in}\noindent\textbf{Stereo Video Generation.}
Recently, a growing number of studies~\cite{wang2024stereodiffusion, dai2024svg, zhao2024stereocrafter, zhang2024spatialme, shi2024immersepro, sun2024splatter, shi2024stereocrafter} have focused on converting monocular videos into stereo videos. Most of these approaches rely on pre-existing depth estimation results, followed by warping and inpainting operations in the latent space. Some methods, like StereoDiffusion~\cite{wang2024stereodiffusion} and SVG~\cite{dai2024svg}  adopt a training-free paradigm, performing inpainting through optimization based on pretrained image or video diffusion priors. While works like StereoCrafter~\citep{zhao2024stereocrafter}, SpatialMe~\citep{zhang2024spatialme}, StereoConversion~\citep{mehl2024stereo},  ImmersePro~\citep{shi2024immersepro} construct large-scale stereo video datasets to train feed-forward networks capable of directly completing the warped videos.

However, such approaches cannot be directly applied to explorable stereo world model generation. A straightforward solution might involve extending the outputs of a monocular world model using the aforementioned techniques. Nonetheless, these methods depend heavily on video depth estimation and warping, making them non–end-to-end, computationally inefficient, and susceptible to error accumulation—particularly in fine-detail regions (such as the wire fence illustrated in Fig~\ref{fig:world_model_comparison}).

\vspace{0.1in}\noindent \textbf{Multi-View Video Generation.}
Multi-view generation has also emerged as a rapidly evolving research direction. CAT3D~\cite{gao2024cat3d} enables novel view synthesis from single- or multi-view images by combining multi-view diffusion with NeRFs. SV4D~\cite{xie2024sv4d} extend Stable Video Diffusion (SVD)~\cite{blattmann2023stable} into Stable Video 4D (SV4D), which reconstructs a 4D scene from a single input video; however, their method is limited to a foreground animated object and does not model the background. Similar approaches, such as Generative Camera Dolly~\cite{van2024generative}, CAT4D~\cite{wu2025cat4d} and SynCamMaster~\cite{bai2024syncammaster}, also explore view synthesis across large camera baselines. Nevertheless, these methods primarily target novel view generation and are not directly applicable to stereo video generation.
\section{Stereo World Model}

\begin{figure*}[t]
    \centering
    \includegraphics[width=0.95\textwidth]{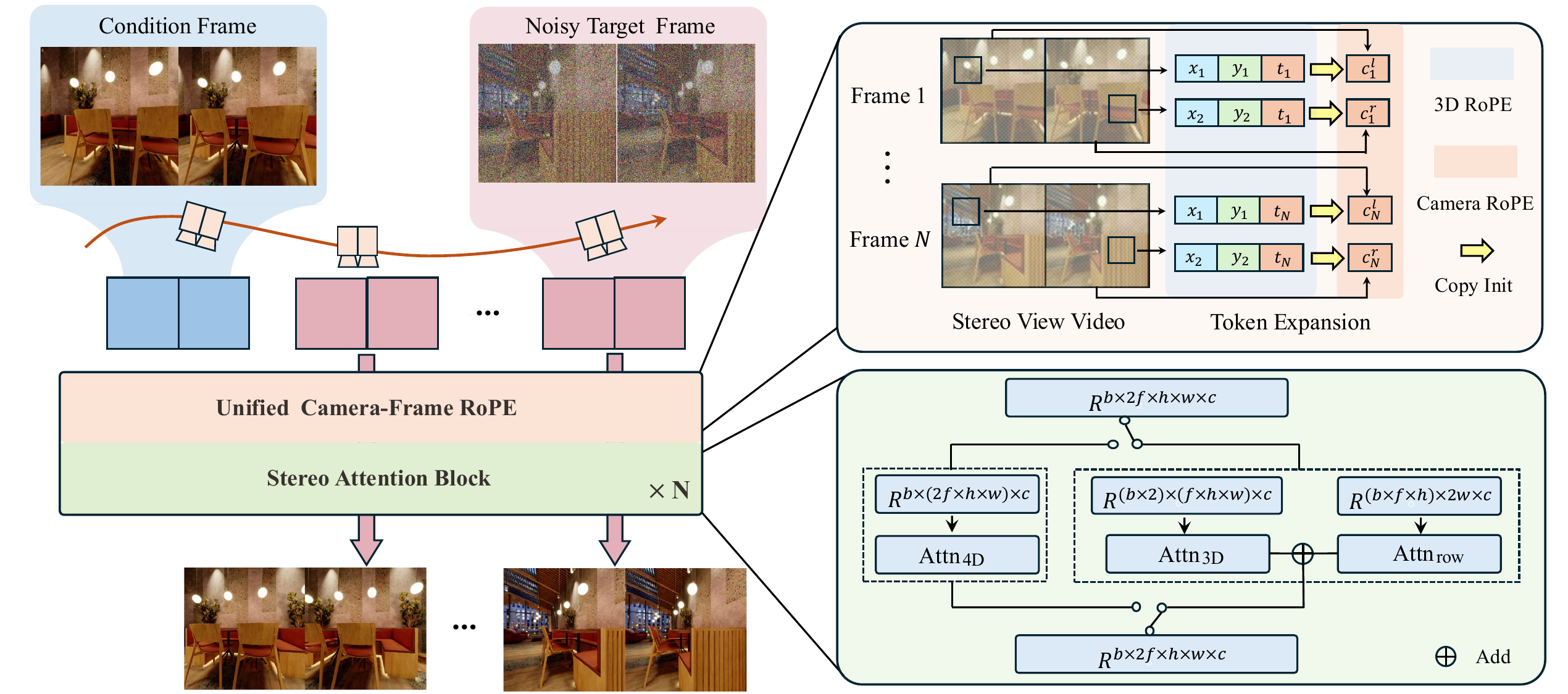}
    \vspace{-3mm}
    \caption{Illustration of StereoWorld. Given a pair of stereo images and a conditional camera trajectory, StereoWorld first encodes conditional and noisy video latents from different viewpoints and timesteps using a unified camera–frame RoPE representation. It then performs denoising through a DiT equipped with stereo attention, ultimately producing the final stereo video.}
    \vspace{-3mm}
    \label{fig:pipeline}
\end{figure*}

Given a rectified stereo pair  \((\mathbf{I}_{\text{left}}, \mathbf{I}_{\text{right}}) \in \mathbb{R}^{3 \times H \times W}\) with baseline $b$ and and a scene prompt $\mathbf{c}$,  our goal is to synthesize a stereo video conditioned on an action specified as a camera trajectory  $\{\texttt{cam}_t\} := \{(\mathbf{K}_t \in \mathbb{R}^{3\times3}, \mathbf{T}_t \in \mathbb{R}^{4\times4}), t \in (1,2,\cdots,N)\} $ where $\mathbf{K}$ and $\mathbf{T}$ are the intrinsic and extrinsic respectively, and  $N$ denotes the number of actions. The generated sequences should (i) remain temporally smooth while following the prescribed camera motion, and (ii) be left-right consistent at every timestep. 
To this end, building upon a pre-trained video diffusion model (Sec.~\ref{sec:pre-trained vdm}), we propose \textbf{StereoWorld} with two key components (Fig.~\ref{fig:pipeline}): (a) a \textbf{unified camera-frame positional embedding} strategy that expands the backbone’s latent token space and augments it with camera-aware RoPE, minimally perturbing pretrained priors (Sec.~\ref{sec:unified_camera_frame}); and (b) a \textbf{stereo-aware attention mechanism}  (Sec. \ref{sec:stereo_attention}) that decomposes cross-view fusion into 3D intra-view attention plus horizontal row attention, balancing computational efficiency with accurate epipolar (disparity-aligned) correspondence.

\subsection{Pre-trained Video Diffusion Model}
\label{sec:pre-trained vdm}
Our work builds on a pretrained video diffusion model and repurposes it for stereo world modeling, enabling us to leverage the strong spatiotemporal priors and visual fidelity provided by large-scale video pretraining. Specifically, we adopt a latent diffusion model ~\cite{blattmann2023align} consisting of a 3D Variational Autoencoder (VAE) ~\cite{kingma2013auto} and a Transformer-based diffusion model (DiT)~\cite{peebles2023scalable}.
The VAE encoder $\mathcal{E}$ compresses the video ($\mathbf{V} \in \mathbb{R}^{F\times H \times W \times 3}$) into a compact spatiotemporal latent representation:
\begin{equation}
    \mathbf{z} = \mathcal{E}(\mathbf{V}) \in \mathbb{R}^{f\times h \times w \times c}.
    \label{eq:encoder}
\end{equation}
The DiT is then trained in this latent space, progressively denoising noisy latent variables into video latents following the rectified flow formulation~\cite{esser2024scaling}. Once trained, the model can generate samples from pure noise via iterative denoising. After denoising, the VAE decoder  $\mathcal{D}$ 
reconstructs the latents back into the pixel domain. 
In our stereo setting, a stereo video $\{\mathbf{V}_\text{left}, \mathbf{V}_\text{right} \} \in \mathbb{R}^{F \times H \times W \times 3}$ is encoded in a viewpoint-agnostic manner using Eq.~\eqref{eq:encoder}, producing latent representations  $\{\mathbf{z}_\text{left}, \mathbf{z}_\text{right}\}$.



\paragraph{Rotary Positional Encoding and Attention}

Vanilla RoPE~\cite{su2024roformer} encodes relative positions by rotating the query and key vectors before dot-product attention. For a 1D sequence, the attention matrix is defined as: 
\begin{equation}
\small
\mathbf{A}_{t_1, t_2} 
= (\mathbf{q}_{t_1} \mathbf{R}_{t_1}(d) ) (\mathbf{k}_{t_2} \mathbf{R}_{t_2}(d))^{\top}
= \mathbf{q}_{t_1} 
\mathbf{R}_{\Delta t}(d) 
\mathbf{k_{(t_2)}},
\label{eq:rope}
\end{equation}
where $\Delta t=t_1-t_2$, $\mathbf{q}_{t_1}$, $\mathbf{k}_{t_2}$ are the query and key embeddings at positions $t_1$ and $t_2$ and $\mathbf{R}_{\Delta_t}(d)$ is the relative rotation matrix acting on each 2D subspace of the $d$-dimensional embedding. The relative rotation matrix $\mathbf{R}_{\Delta_t}(d) = \exp(\Delta t  \theta_n \mathrm{i}) \in \mathbb{R}^{d \times d}$, where $\mathrm{i}$ is the imaginary unit, and $\theta_n$ is the frequency of rotation applied to a specific $n$-th pair of $d$ dimensions ($n = 0, \dots, d/2-1)$, enables the model to capture relative positional relationships directly within attention.

For video, recent RoPE variants (e.g., M-RoPE in Qwen2-VL~\cite{wang2024qwen2}) preserve the inherent 3D structure by factorizing rotations along time and space. Let positions be $(t,x,y)$. The attention term becomes:
{
\begin{equation}
\mathbf{A}_{(t_1, x_1, y_1), (t_2, x_2, y_2)} 
= \mathbf{q}_{(t_1, x_1, y_1)} 
\mathbf{R}_{\Delta t, \Delta x, \Delta y}(d) 
\mathbf{k}^{\top}_{(t_2, x_2, y_2)},
\label{eq:3drope}
\end{equation}
}
where $\Delta t = t_1 - t_2, \Delta x = x_1 - x_2, \Delta y = y_1 - y_2$, and $\mathbf{R}_{\Delta t, \Delta x, \Delta y} = \mathbf{R}_{\Delta t} \mathbf{R}_{\Delta x} \mathbf{R}_{\Delta y}$. The rotations
$\mathbf{R}_{\Delta t}$, $\mathbf{R}_{\Delta x}$, and $\mathbf{R}_{\Delta y}$ act on \emph{disjoint} 2D subspaces of the $d$-dimensional feature, so they commute and compose multiplicatively.
In practice (e.g., Wan~\cite{wan2025}-style implementations), the feature dimension $d$ is partitioned evenly across $t$, $x$, and $y$, with independent 1D RoPEs applied per axis and then composed as above. 

\subsection{Unified Camera-Frame RoPE}
\label{sec:unified_camera_frame}
Fine-tuning a pretrained DiT video diffusion model into a stereo world model requires injecting camera conditioning -- including stereo cameras with varying baselines and dynamic camera motions -- while minimizing disruption to the pretrained prior.

A common approach concatenates Pl{\"u}kcer Ray encodings~\cite{zhang2024cameras} onto the input feature channels.
However, similar to early positional encoding methods~\cite{vaswani2017attention}, this approach relies on absolute coordinates, making it sensitive to the choice of reference frame. To mitigate this limitation, recent methods such as GTA~\cite{miyato2023gta} and PRoPE~\cite{li2025cameras} model relative camera positions, yielding improved generalization. Specifically, PRoPE replaces
$\mathbf{R}_{\Delta t, \Delta x, \Delta y}$ in Eq.~\eqref{eq:3drope} with $\mathbf{R}_{\Delta t, \Delta x, \Delta y} ^{\Delta \texttt{cam}}$,
where
\begin{align}
\mathbf{R}_{\Delta t, \Delta x, \Delta y} ^{\Delta \texttt{cam}} (d) =& \mathbf{R}_{t_1, x_1, y_1} ^{\texttt{cam}_{t_1}} (d) (\mathbf{R}_{t_2, x_2, y_2} ^{\texttt{cam}_{t_2}} (d))^{\top}, \\
\mathbf{R}_{t_j, x_j, y_j} ^{\texttt{cam}_{t_j}}(d) =& \begin{bmatrix}
    \mathbf{I}_{d/8} \otimes {\mathbf{P}}_{j} & \mathbf{0} \\
    \mathbf{0}  & \mathbf{R}_{t_j, x_j, y_j}(d/2) 
    \end{bmatrix}, \label{eq:prope_R}\\
{\mathbf{P}}_{j} =& \begin{bmatrix} \boldsymbol{K}_j & \mathbf{0} \\ \mathbf{0} & 1 \end{bmatrix} \boldsymbol{T}_j, \quad \boldsymbol{K}_j, \boldsymbol{T}_j = \texttt{cam}_{t_j}. \nonumber
\end{align}
Here $j \in \{1,2\}$, $\otimes$ is the Kronecker product, and $\mathbf{I}_{d/8} \in \mathbb{R}^{d/8 \times d/8} $ is the identity matrix.
However, when fine-tuning a pretrained model (e.g., Wan~\cite{wan2025}), directly modifying the original positional encoding with Eq.~\eqref{eq:prope_R} can significantly disrupt the model’s learned prior, because the DiT’s attention weights, normalization statistics, and token bases are co-adapted to the original RoPE frequencies and axis partitioning. 

To address this, we propose injecting camera positional encodings by expanding the token dimension, rather than altering the original encoding scheme. Concretely, we extend the original self-attention layer by increasing its feature dimension, i.e.
\begin{equation}
    \mathbf{\tilde{q}}_{(t,x,y)} = \begin{bmatrix} \mathbf{q}_{(t,x,y)} \\ \mathbf{q_\texttt{cam}}_{(t,x,y)} \end{bmatrix} \in \mathbb{R}^{d+d_c},
\end{equation}
Here $d_c$ is the expanded dimension for camera RoPE. The same expansion is also applied to $\mathbf{k}$. Hence the rotary matrix in Eq.~\eqref{eq:prope_R} can be extended to $ \mathbb{R}^{(d+d_c) \times (d+d_c)}$:
\begin{align}
\mathbf{\tilde{R}}_{t, x, y} ^{\texttt{cam}_{t}}(d+d_c) = 
\begin{bmatrix}  
\mathbf{R}_{\Delta t, \Delta x, \Delta y}(d) & \mathbf{0} \\
    \mathbf{0}  & \mathbf{I}_{d_c/4} \otimes {\mathbf{P}}_{t} 
    \end{bmatrix},
\end{align}
leading to our \emph{unified camera-frame RoPE}: 
\begin{align}
\mathbf{\tilde{R}}_{\Delta t, \Delta x, \Delta y} ^{\Delta \texttt{cam}} (d^\prime) = 
\mathbf{\tilde{R}}_{t_1,  x_1, y_1} ^{\texttt{cam}_{t_1}}&(d^\prime)
(\mathbf{\tilde{R}}_{t_2,  x_2, y_2} ^{\texttt{cam}_{t_2}}(d^\prime))^{\top}, \label{eq:UniRoPE}
\end{align}
where $d^\prime=d+d_c$. In this setup, the first $d\times d$ block of the matrix remains identical to that in Eq.~\eqref{eq:3drope}, which aligns with the pretrained prior. For the newly added $d_c\times d_c$block, we experiment with two different initialization strategies for the expanded layer corresponding to $\mathbf{q}_{\texttt{cam}}$ and $\mathbf{k}_{\texttt{cam}}$.

We experiment with two initialization schemes for the new subspace ($\mathbf{q}_{\texttt{cam}}$ and $\mathbf{k}_{\texttt{cam}}$): 
\begin{itemize}
\item \textbf{Zero Init} ensures that the model’s initial output remains identical to that of the pretrained model. However, this initialization makes training more challenging, as the camera conditioning signal is difficult to activate effectively.
\item \textbf{Copy Init} initializes the new subspace with temporal attention weights. Since camera and temporal embeddings operate at the frame level, this provides a strong starting point while minimally affecting pretrained behavior. 
\end{itemize} 

{In contrast to PRoPE~\cite{li2025cameras}, our unified camera–frame RoPE expands the token dimension rather than reparameterizing RoPE, preserving the pretrained positional subspace and adding an orthogonal, camera-conditioned channel. Empirically (Fig.~\ref{fig:camera_ablation}), this yields more stable training, faster convergence. 




\subsection{Stereo-Aware Attention}
\label{sec:stereo_attention}
With the unified camera-frame representation, camera positional encodings for each viewpoint are injected into the stereo video latents ${\mathbf{z}_{\text{left}}, \mathbf{z}_{\text{right}}}$, modeling relationships between arbitrary token pairs as 
$\mathbf{\tilde{q}}_{(t_1,x_1,y_1)} \mathbf{\tilde{R}}_{\Delta t, \Delta x, \Delta y} ^{\Delta cam} (d^\prime) \mathbf{\tilde{k}}_{(t_2,x_2,y_2)}$. This unified formulation allows our method to seamlessly accommodate multi stereo video datasets with varying baselines and intrinsic parameters, as demonstrated in Tab.~\ref{tab:training_data}.

With this representation, a naive stereo generator concatenates left–right tokens along the sequence dimension and applies full joint attention over features $f^{in} \in \mathbb{R}^{b\times2f\times h\times w \times c}$, yielding a 4D attention ({$\text{Attn}_\text{4D}$}) that couples spatial, temporal, and viewpoint dependencies. However, because attention cost grows quadratically with the number of tokens, this approach is computationally prohibitive for video synthesis. 


Observing that in rectified stereo pairs the epipolar lines align horizontally, we exploit this geometry to design a more efficient \emph{stereo-aware attention}. The 4D attention is decomposed into:
(a) intra-view 3D attentions ($\text{Attn}_\text{3D}$) capturing spatial–temporal dynamics, and
(b) cross-view attentions computed only among horizontally aligned tokens at the same timestep ($\text{Attn}_\text{row}$).
As illustrated in Fig.~\ref{fig:pipeline}, the final output aggregates both components:

\begin{equation}
f^{\text{out}} = \text{Attn}_\text{3D}(f^{\text{in}}) + \text{Attn}_\text{row}(f^{\text{in}}).
\end{equation}



With this design, the overall computational complexity is reduced from $\mathcal{O}((2f \cdot h \cdot w)^2)$ to $\mathcal{O}(2 \cdot (f \cdot h \cdot w)^2) + f \cdot h \cdot (2w)^2)$.
We report a comparison of the performance differences between these two attention mechanisms in Tab.~\ref{tab:attention_ablation}, which demonstrates the efficiency and effectiveness of the proposed decoupled attention scheme.





\section{Experiment}

\begin{table}[ht]
  \vspace{-3.5mm}
  \centering
    \caption{Training Data Information.}
  \vspace{-3.5mm}
  \adjustbox{width={\linewidth},keepaspectratio}{
    \begin{tabular}{l|cccc}
        \bottomrule
            {\textbf{Dataset}} & {Sample Num.} & {Baseline / $m$} & {Motion} & {Domain}\\
            \hline 
            {Stereo4D~\cite{Jin2024Stereo4DLH}} & {11718} & {0.063} & {Dynamic} & {Realistic}  \\
            {TartanAir~\cite{Wang2020TartanAirAD}} & {6433} & {0.25} & {Static} & {Synthetic}  \\
            {TartanAirGround~\cite{patel2025tartanground}} & {58168} & {0.25} & {Static} & {Synthetic}  \\
            {DynamicReplica~\cite{karaev2023dynamicstereo}} & {1686} & {Varying} & {Dynamic} & {Synthetic}  \\
            {VKitti~\cite{Cabon2020VirtualK2}} & {230} & {Varying} & {Dynamic} & {Synthetic}  \\
        \bottomrule
    \end{tabular}
    }
    \label{tab:training_data}
  \vspace{-5.5mm}
\end{table}

\begin{figure*}[t]
    \centering
    \setlength{\fboxrule}{0.5pt}
    \setlength{\fboxsep}{-0.01cm}
    \setlength\tabcolsep{0pt}
    \begin{spacing}{1}
    \begin{tabular}{p{0.04\linewidth}<{\centering}p{0.16\linewidth}<{\centering}p{0.16\linewidth}<{\centering}p{0.16\linewidth}<{\centering}p{0.16\linewidth}<{\centering}p{0.16\linewidth}<{\centering}p{0.16\linewidth}<{\centering}}
    
    & Left View & Right View & Left View& Right View & Left View & Right View \\

    \rotatebox{90}{ \hspace{1mm} \small{Aether~\cite{Team2025AetherGU}}} &
    \includegraphics[width=1\linewidth, trim=0 0 0 0,clip]{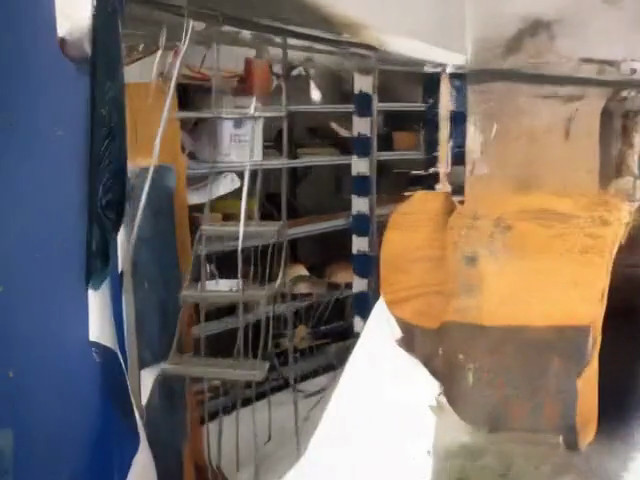} &
    \includegraphics[width=1\linewidth, trim=0 0 0 0,clip]{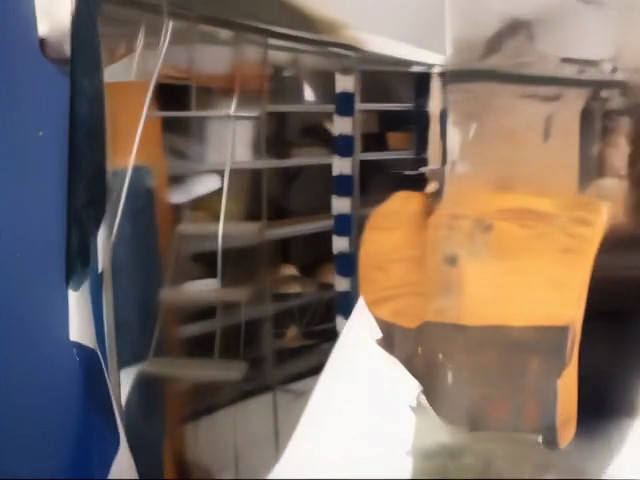} &
    \includegraphics[width=1\linewidth, trim=0 0 0 0,clip]{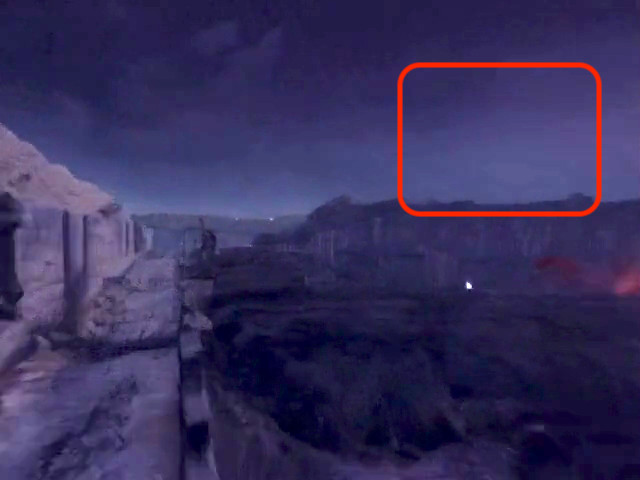} &
    \includegraphics[width=1\linewidth, trim=0 0 0 0,clip]{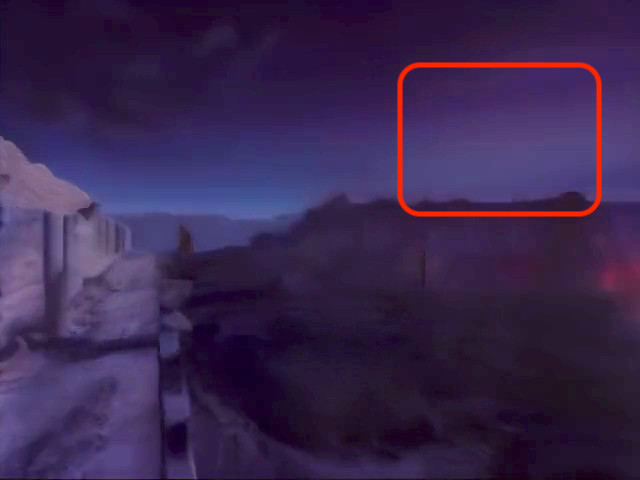} &
    \includegraphics[width=1\linewidth, trim=0 0 0 0,clip]{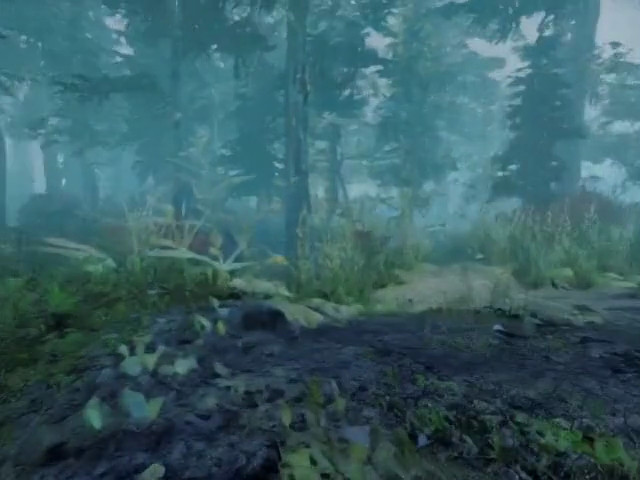} &
    \includegraphics[width=1\linewidth, trim=0 0 0 0,clip]{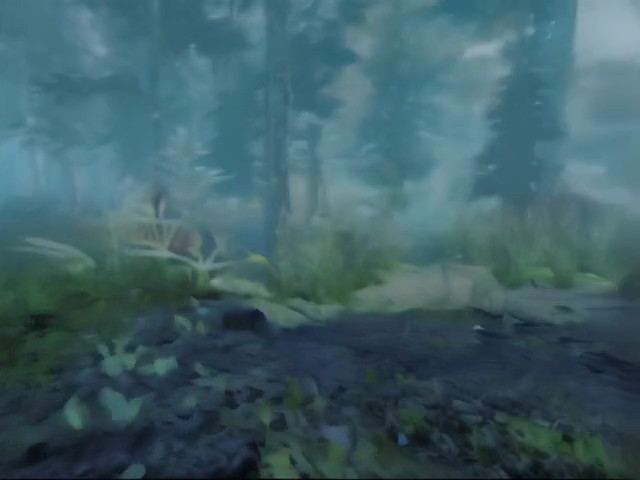}
    \\
    \specialrule{0em}{0pt}{-15pt} \\
    \rotatebox{90}{ \hspace{-1mm} \small{DeepVerse~\cite{chen2025deepverse}}} &
    \includegraphics[width=1\linewidth, trim=0 0 0 0,clip]{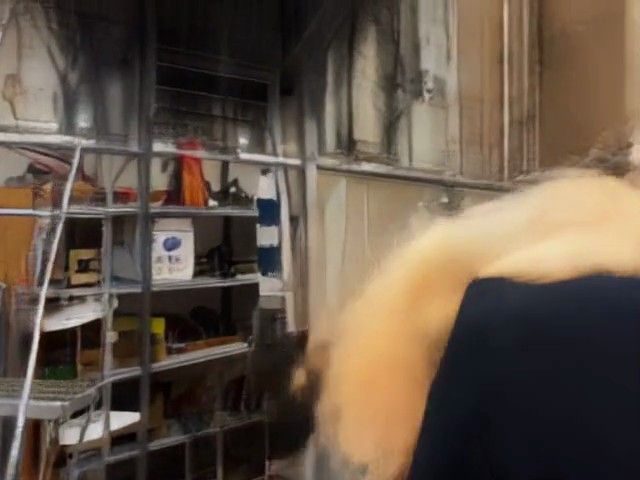} &
    \includegraphics[width=1\linewidth, trim=0 0 0 0,clip]{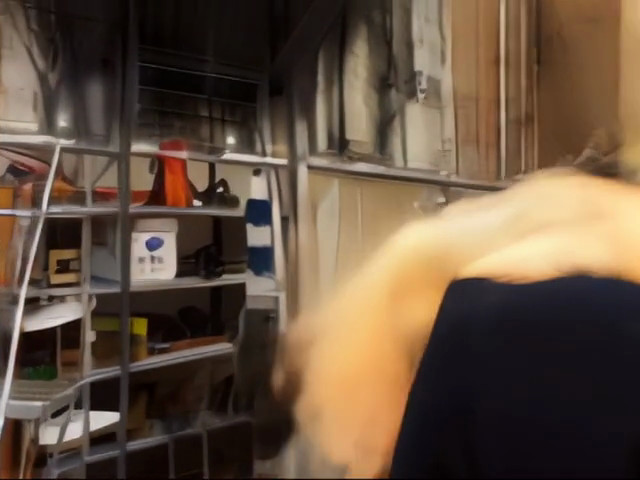} &
    \includegraphics[width=1\linewidth, trim=0 0 0 0,clip]{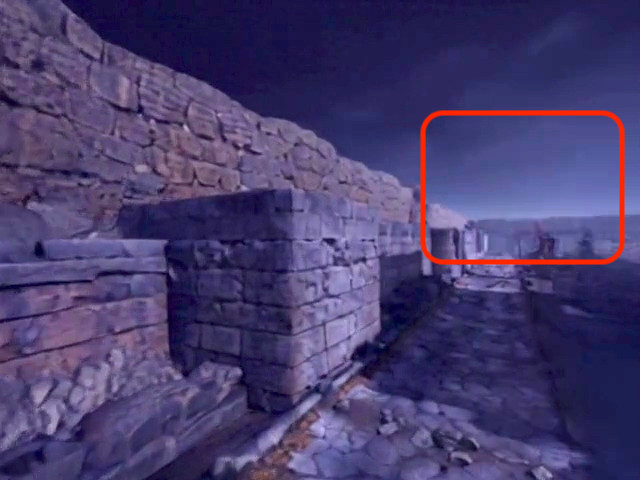} &
    \includegraphics[width=1\linewidth, trim=0 0 0 0,clip]{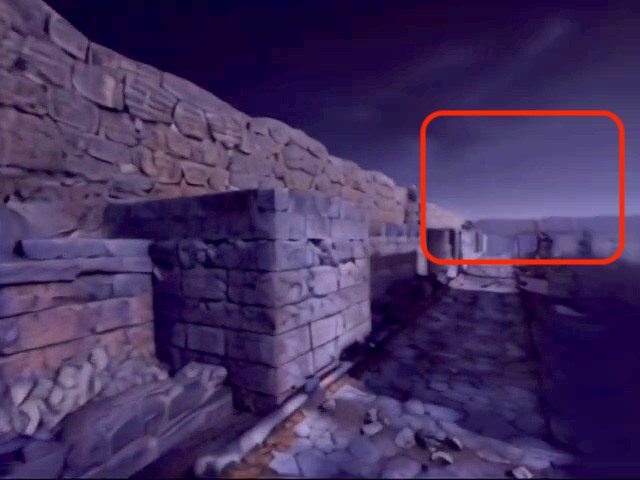} &
    \includegraphics[width=1\linewidth, trim=0 0 0 0,clip]{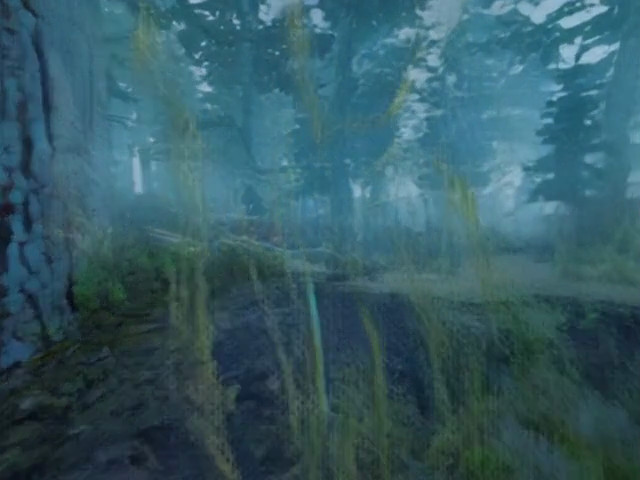} &
    \includegraphics[width=1\linewidth, trim=0 0 0 0,clip]{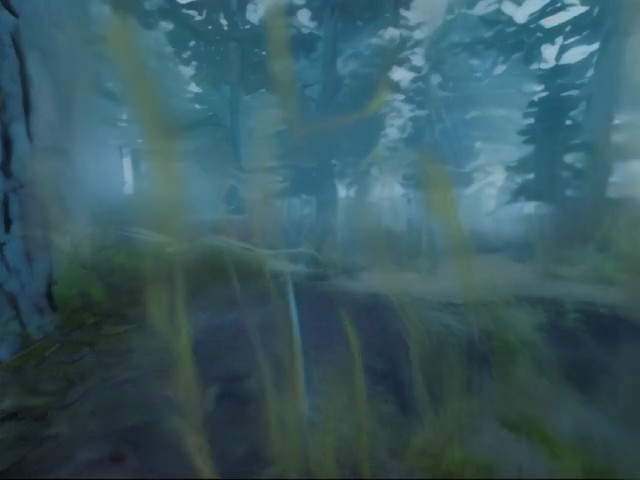}
    \\
    \specialrule{0em}{0pt}{-15pt} \\
    \rotatebox{90}{ \hspace{3mm} \small{SEVA~\cite{Zhou2025StableVC}}} &
    \includegraphics[width=1\linewidth, trim=0 0 0 0,clip]{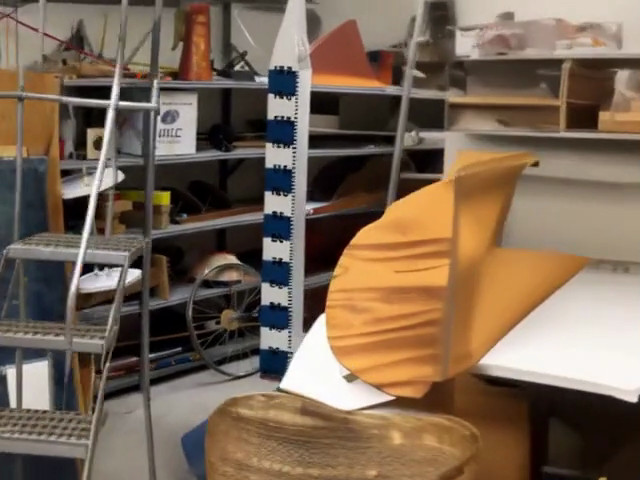} &
    \includegraphics[width=1\linewidth, trim=0 0 0 0,clip]{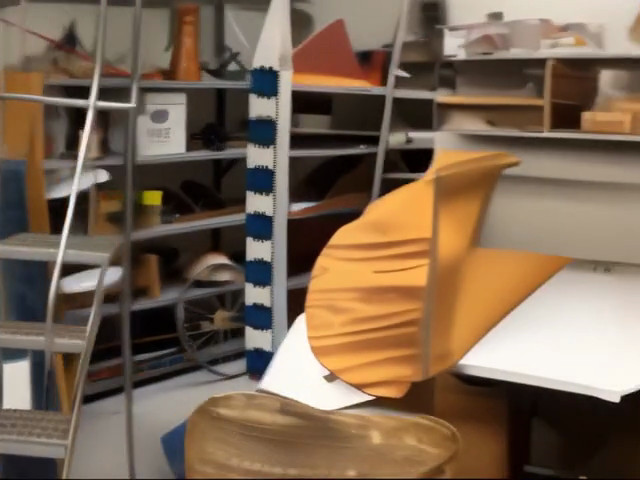} &
    \includegraphics[width=1\linewidth, trim=0 0 0 0,clip]{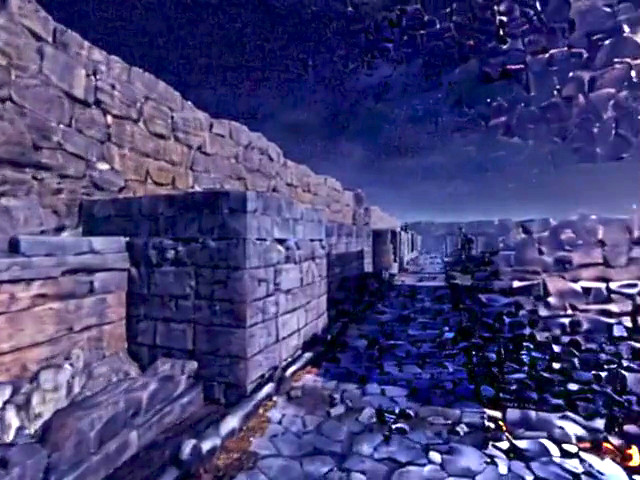} &
    \includegraphics[width=1\linewidth, trim=0 0 0 0,clip]{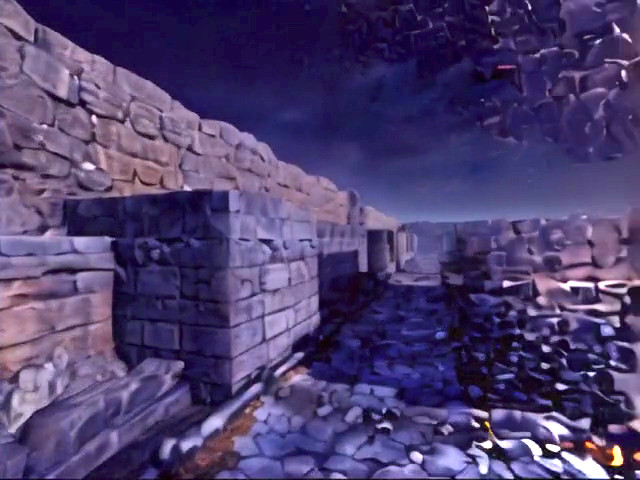} &
    \includegraphics[width=1\linewidth, trim=0 0 0 0,clip]{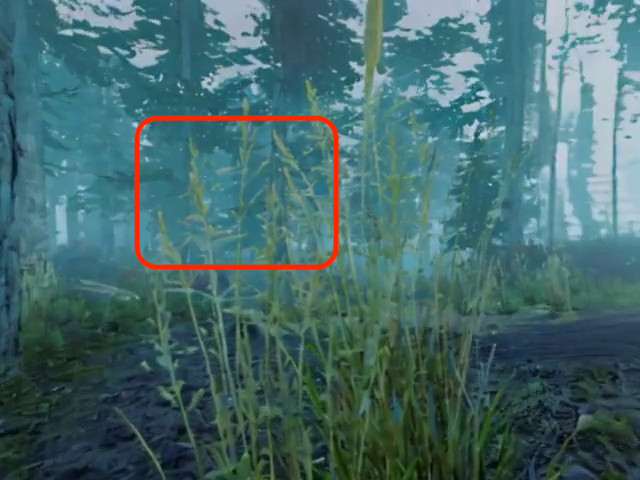} &
    \includegraphics[width=1\linewidth, trim=0 0 0 0,clip]{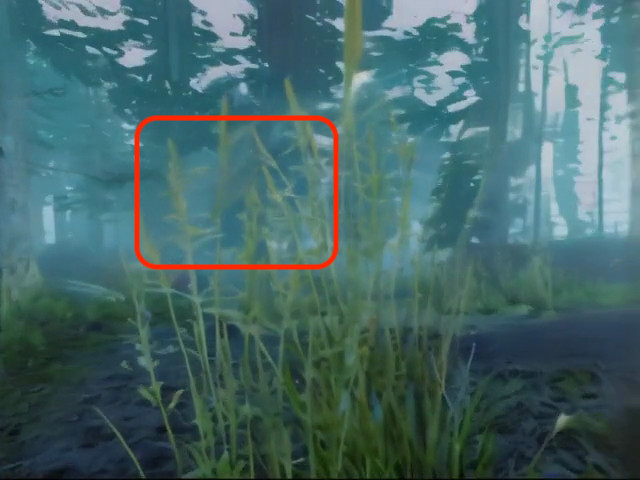}
    \\
    \specialrule{0em}{0pt}{-15pt} \\
    \rotatebox{90}{ \hspace{-3mm} \small{ViewCrafter~\cite{yu2024viewcrafter}}} &
    \includegraphics[width=1\linewidth, trim=0 0 0 0,clip]{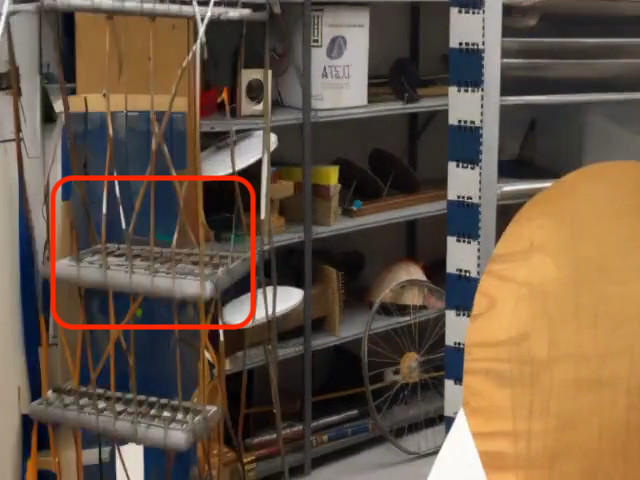} &
    \includegraphics[width=1\linewidth, trim=0 0 0 0,clip]{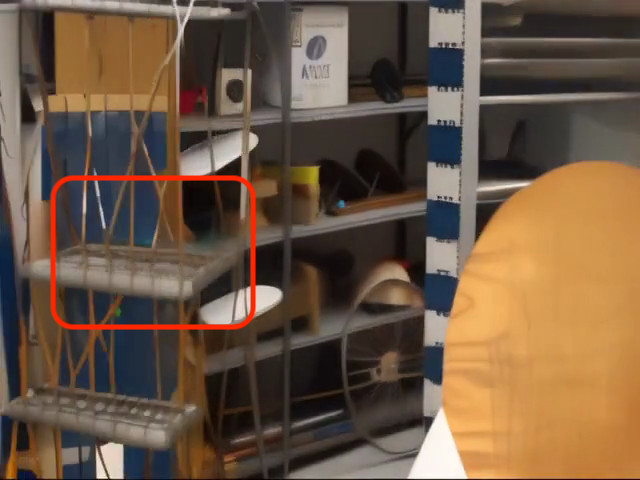} &
    \includegraphics[width=1\linewidth, trim=0 0 0 0,clip]{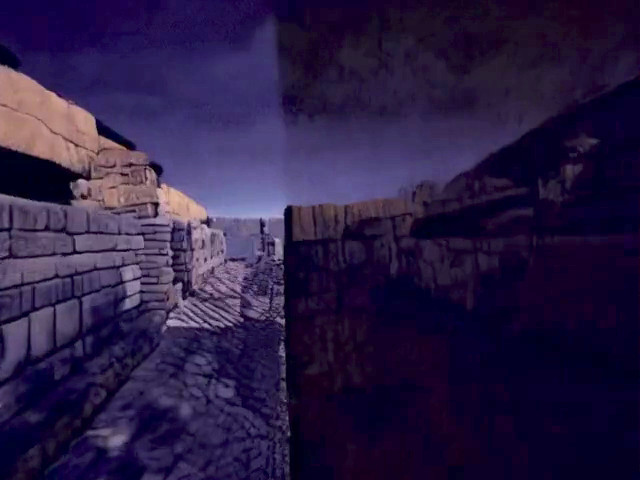} &
    \includegraphics[width=1\linewidth, trim=0 0 0 0,clip]{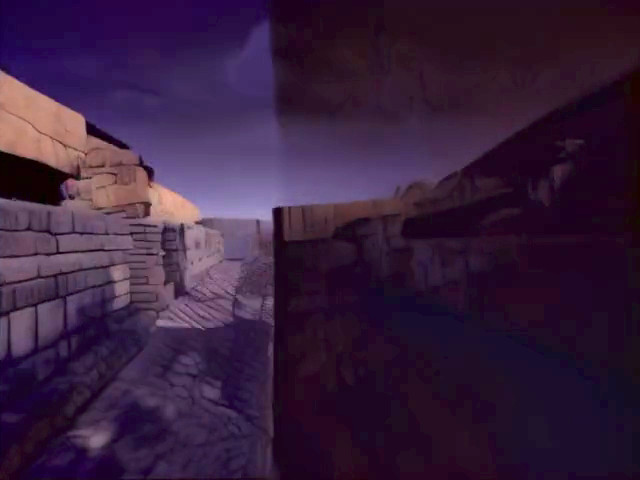} &
    \includegraphics[width=1\linewidth, trim=0 0 0 0,clip]{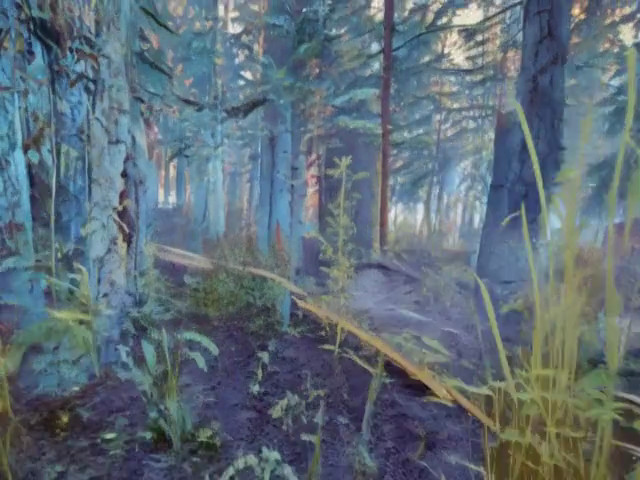} &
    \includegraphics[width=1\linewidth, trim=0 0 0 0,clip]{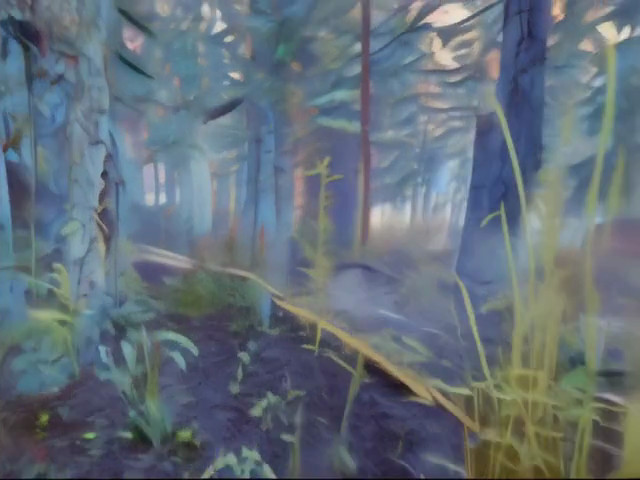}
    \\
    \specialrule{0em}{0pt}{-15pt} \\
    \rotatebox{90}{ \hspace{5mm} \small{Ours}} &
    \includegraphics[width=1\linewidth, trim=0 0 0 0,clip]{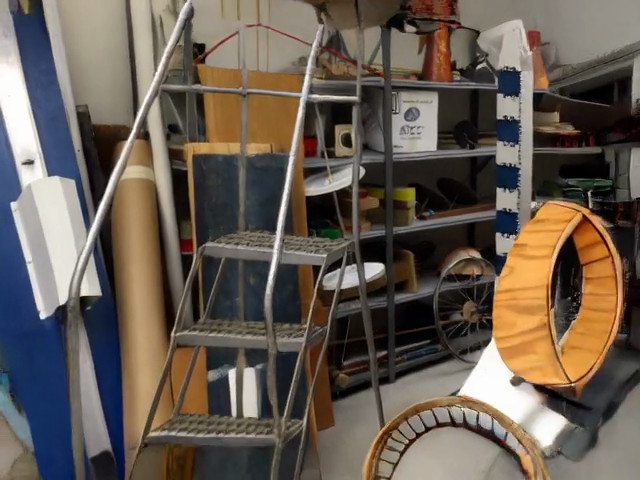} &
    \includegraphics[width=1\linewidth, trim=0 0 0 0,clip]{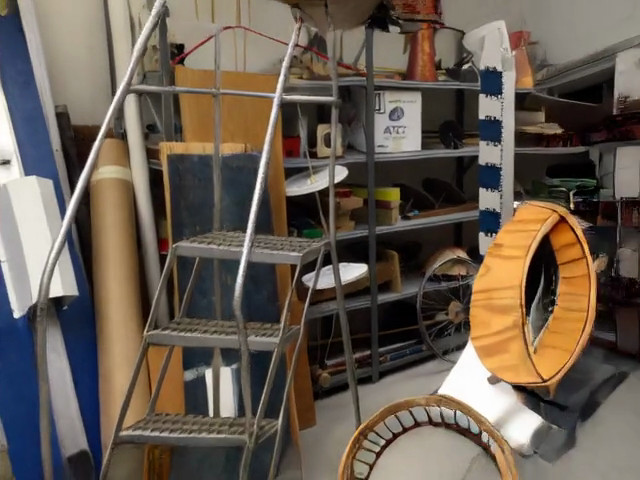} &
    \includegraphics[width=1\linewidth, trim=0 0 0 0,clip]{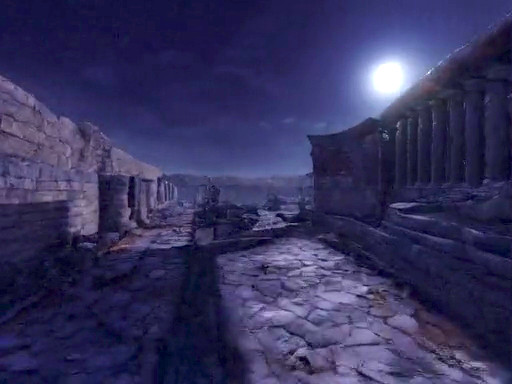} &
    \includegraphics[width=1\linewidth, trim=0 0 0 0,clip]{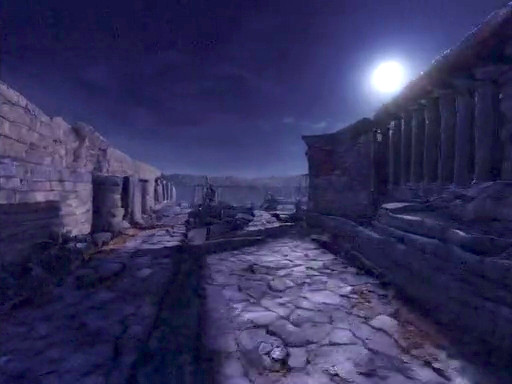} &
    \includegraphics[width=1\linewidth, trim=0 0 0 0,clip]{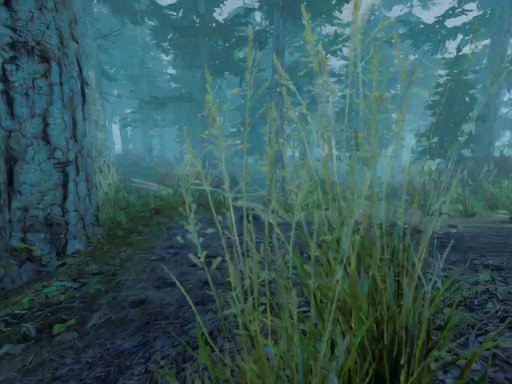} &
    \includegraphics[width=1\linewidth, trim=0 0 0 0,clip]{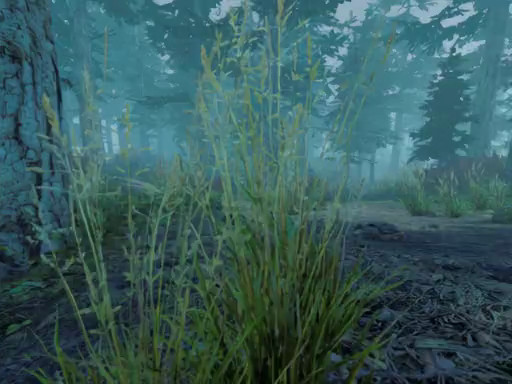}

    \end{tabular}
    \end{spacing}
    \vspace{-3mm}
    \caption{ Stereo video generation comparison with SOTA methods augmented by post-hoc stereo conversion. Our method directly generates stereo video in an end-to-end manner, enabling better preservation of inter-view detail consistency and tonal coherence.}
    \vspace{-3mm}
    \label{fig:stereo_generation_comparison}
\end{figure*}

\subsection{Implementation Details}
We implement StereoWorld based on the video generation model Wan2.2-TI2V-5B~\cite{wan2025}. The model is trained on a mixed dataset list in Tab.~\ref{tab:training_data}. Each video clip contains 49 frames, and is cropped and resized to 480$\times$640 before feeding to the network. We train StereoWorld using AdamW optimizer~\cite{Loshchilov2017DecoupledWD} for 20k steps, with batch size of 24, on 24 NVIDIA H20 GPUS.
The learning rate is set to 1e-4. 

\subsection{Benchmark Datasets and Metric}
\paragraph{Evaluation Datasets.} We construct the evaluation set with 435 stereo images sampled from FoundationStereo~\cite{wen2025stereo}(\textit{Synthetic}), UnrealStereo4K~\cite{Tosi2021SMDNetsSM}(\textit{Synthetic}), TartanAir Testset(\textit{Synthetic}) and Middlebury~\cite{Scharstein2014HighResolutionSD}(\textit{Realistic}), covering both indoor and outdoor scenes, and versatile textures and various baselines. For each stereo image, we use Qwen2.5-VL~\cite{qwen2.5-VL} to caption the scene and sample a random camera trajectory. 

\paragraph{Evaluation Metrics.} 
StereoWorld is evaluated on camera accuracy, left-right view synchronization, visual quality and FPS. For \textit{camera accuracy}, we extract camera poses from the generated videos, computing both rotation and translation errors (RotErr and TransErr). \textit{View synchronization} is measured using image matching technique GIM \citep{gim} to count the number of matching pixels exceeding a confidence threshold (Mat. Pix.). We further measure cross-domain alignment using the FVD-V score from SV4D \citep{sv4d} and the average CLIP similarity between corresponding source and target frames at each timestep, denoted CLIP-V \citep{CVD}.
For \textit{visual quality}, we evaluate fidelity, text coherence, and temporal consistency using Fréchet Image Distance (FID) \citep{fid}, Fréchet Video Distance (FVD) \citep{fvd}, CLIP-T, and CLIP-F, respectively, following~\cite{Bai2025ReCamMasterCG}.
We also benchmark our method using the standard VBench metrics \cite{vbench}.


\subsection{Stereo Video Comparison}
\paragraph{Baselines.}

\begin{table*}[ht]
  \vspace{-3.5mm}
  \centering
    \caption{Comparison of stereo video with SOTA methods on visual quality, camera accuracy, view synchronization and FPS.}
    \vspace{-3mm}
  \adjustbox{width={\linewidth},keepaspectratio}{
    \begin{tabular}{l|c|cccc|cc|ccc|c}
    \bottomrule
         \multirow{2}{*}{\textbf{Method}} & \multirow{2}{*}{\textbf{Modality}} & \multicolumn{4}{c|}{Visual Quality} & \multicolumn{2}{c|}{Camera Accuarcy} & \multicolumn{3}{c|}{View Synchronization} & \multirow{2}{*}{FPS$\uparrow$}  \\

          & & {{\small{FID}}}$\downarrow$ & {\small{{FVD}}}$\downarrow$ & {\small{CLIP-T}}$\uparrow$ & {\small{CLIP-F}}$\uparrow$ 
          & {\small{{RotErr}}}$\downarrow$ & {\small{TransErr}}$\downarrow$ 
          & {\small{{Mat. Pix.(K)}}}$\uparrow$ & {\small{FVD-V}}$\downarrow$ & {\small{{CLIP-V}}}$\uparrow$ 
          &  \\
        \hline
        {Voyager~\cite{Huang2025VoyagerLA}}  & {RGBD}    
        & {226.97} & \cellcolor{third}{170.37} & {24.85} & {97.03} & {1.34} & {0.25} & {4.26} & {55.45} & {91.41} & {0.03} \\
        {DeepVerse~\cite{chen2025deepverse}} & {RGBD}     
        & \cellcolor{third}{191.32} & {176.72} & {24.59} & \cellcolor{third}{97.31} & {1.51} & \cellcolor{third}{0.16} & {4.48} & \cellcolor{third}{33.50} & \cellcolor{third}{93.86} & \cellcolor{second}{0.35} \\
        {Aether~\cite{Team2025AetherGU}} & {RGBD}    
        & \cellcolor{second}{185.72} & \cellcolor{second}{152.97} & \cellcolor{third}{24.93} & {97.14} & {1.50} & \cellcolor{second}{0.13} & {4.35} & {42.07} & {93.71} & {0.11} \\
        {SEVA~\cite{Zhou2025StableVC}} & {RGB}   
        & {195.70} & {170.92} & {24.77} & \cellcolor{best}{98.11} & \cellcolor{second}{1.09} & {0.51} & \cellcolor{third}{4.49} & \cellcolor{second}{31.10} & \cellcolor{second}{94.73} & {0.10} \\
        {ViewCrafter~\cite{yu2024viewcrafter}} & {RGB}  
        & {211.89} & {185.76} & \cellcolor{second}{25.02} & {96.15} & \cellcolor{third}{1.24} & {0.20} & \cellcolor{second}{4.49} & {42.10} & {93.51} & \cellcolor{third}{0.13} \\
        \hline
        \textit{Ours Monocular} & \textit{RGB}   
        & \textit{126.83} & \textit{96.87} & \textit{24.97} & \textit{97.12} & \textit{1.36} & \textit{0.14} & {\xmark} & {\xmark} & {\xmark} & {\xmark} \\
        {Ours Stereo} & {RGB}  
        & \cellcolor{best}{111.36} & \cellcolor{best}{83.04} & \cellcolor{best}{25.74} & \cellcolor{second}{97.55} & \cellcolor{best}{1.01} & \cellcolor{best}{0.11} & \cellcolor{best}{4.56} & \cellcolor{best}{22.00} & \cellcolor{best}{97.50} & \cellcolor{best}{0.49} \\
    \bottomrule
    \end{tabular}}
    \label{tab:stereo_video_comparison}
\end{table*}

\begin{table}[ht]
  \vspace{-3.5mm}
  \centering
    \caption{Comparison of stereo video on Vbench metrics.}
    \vspace{-3mm}
  \adjustbox{width={\linewidth},keepaspectratio}{
		\begin{tabular}{l|cccc}
			\bottomrule
                {\textbf{Method}} & \makecell[c]{Aesthetic\\Quality $\uparrow$} & \makecell[c]{Imaging\\Quality $\uparrow$} & \makecell[c]{Temporal\\Flickering $\uparrow$} &  \makecell[c]{Background\\Consistency $\uparrow$}\\
                \hline 
                {Voyager~\cite{Huang2025VoyagerLA}} & {38.23} & {59.32} & \cellcolor{best}{94.55} & \cellcolor{second}{92.81}  \\
                {DeepVerse~\cite{chen2025deepverse}} & {38.71} & {60.11} & \cellcolor{second}{94.52} & \cellcolor{third}{92.61}  \\
                {Aether~\cite{Team2025AetherGU}} & {39.02} & {60.26} & {93.63} & {92.46}  \\
                {SEVA~\cite{Zhou2025StableVC}} & \cellcolor{second}{40.60} & \cellcolor{second}{64.28} & {93.49} & \cellcolor{best}{93.01}  \\
                {ViewCrafter~\cite{yu2024viewcrafter}} & \cellcolor{third}{40.31} & \cellcolor{third}{61.90} & {90.63} & {91.45}  \\
                Ours & \cellcolor{best}{44.27} & \cellcolor{best}{66.51} & \cellcolor{third}{93.63} & {92.42}  \\
			\bottomrule
		\end{tabular}
    
    }
  \vspace{-5.5mm}
\end{table}

StereoWorld is the first stereo video generation model. To demonstrate the advantages of simultaneous stereo-view generation, we first use a series of state-of-the-art camera-controlled video generation methods to obtain a monocular video, and then extend them into stereo videos using StereoCrafter~\cite{zhao2024stereocrafter}. StereoCrafter is a warp-inpainting video generation model. Therefore, for RGBD generation models~\cite{Huang2025VoyagerLA, chen2025deepverse, Team2025AetherGU}, we directly use the generated depth to warp the video into another view; for RGB generation models~\cite{Zhou2025StableVC, yu2024viewcrafter}, we first use DepthCrafter~\cite{Hu2024DepthCrafterGC} for video depth estimation, and then perform the warping. 
Compared to these multi-stage pipelines, StereoWorld achieves more efficient generation as an end-to-end model, as shown in the``FPS'' column of Tab~\ref{tab:stereo_video_comparison}.

In addition, since the training data used by different models are not well aligned, we also trained a monocular version (``\textit{Ours Monocular}'') of our method under the same settings as the stereo version for comparison, in order to better demonstrate the advantages brought by stereo generation.

\subsubsection{Stereo View Consistency}
Fig.~\ref{fig:stereo_generation_comparison} presents a visual comparison between our method and the baseline approaches on the stereo video generation task. The comparison methods, which rely on additional depth estimation and view inpainting models, often suffer from misaligned details between the left and right views (e.g., the plants in the third column) or exhibit slight color inconsistencies between the two views (e.g., the sky in the second column).
In contrast, our method generates stereo videos end-to-end, effectively avoiding these artifacts and ensuring better view consistency. The results in the "\emph{View Synchronization}`` column of Tab~\ref{tab:stereo_video_comparison} further validate this observation.

\subsubsection{Camera Trajectory}
Our method also achieves superior alignment between the generated results and the conditioned camera parameters. In contrast, warp-based world models~\cite{Huang2025VoyagerLA} often suffer from inaccurate depth estimation or insufficient geometric cues when the viewpoint change is large, leading to misaligned camera conformity. Meanwhile, discrete action-based world models~\cite{chen2025deepverse} lack fine-grained camera control. Benefiting from the unified camera–frame RoPE, our approach effectively models relative relationships between cameras, enabling more precise and continuous camera control. We estimate the camera poses of the generated videos using VGGT~\cite{wang2025vggt} and compare them with the conditioned camera inputs to quantify accuracy. As shown in the “Camera Accuracy” column of Tab.~\ref{tab:stereo_video_comparison}, our method achieves the highest precision. Furthermore, Fig~\ref{fig:camera_traj} visualizes the camera trajectory comparisons, clearly illustrating that our model better preserves the intended camera motion.

\subsubsection{Disparity}
Fig.~\ref{fig:disp_comparison} compares the disparity maps generated by our method with those produced by other RGB-D world models. As shown, existing RGB-D approaches often exhibit artifacts where texture patterns from the RGB outputs are inadvertently transferred into the predicted disparity — for instance, in the third column of Voyager~\cite{Huang2025VoyagerLA} and the first column of Aether~\cite{Team2025AetherGU}.
In contrast, our method effectively mitigates this issue by generating stereo image pairs first and then estimating disparity~\cite{wen2025stereo} from them, leading to cleaner and more geometrically consistent results. 
Moreover, disparity in our setting can be transferred to \emph{metric depth} directly. It is also worth noting that, unlike these comparison methods, our model is trained without any depth supervision, relying solely on binocular image signals.

    

\begin{figure}[t]
    \centering
    \includegraphics[width=1\linewidth, trim=0 0 0 0,clip]{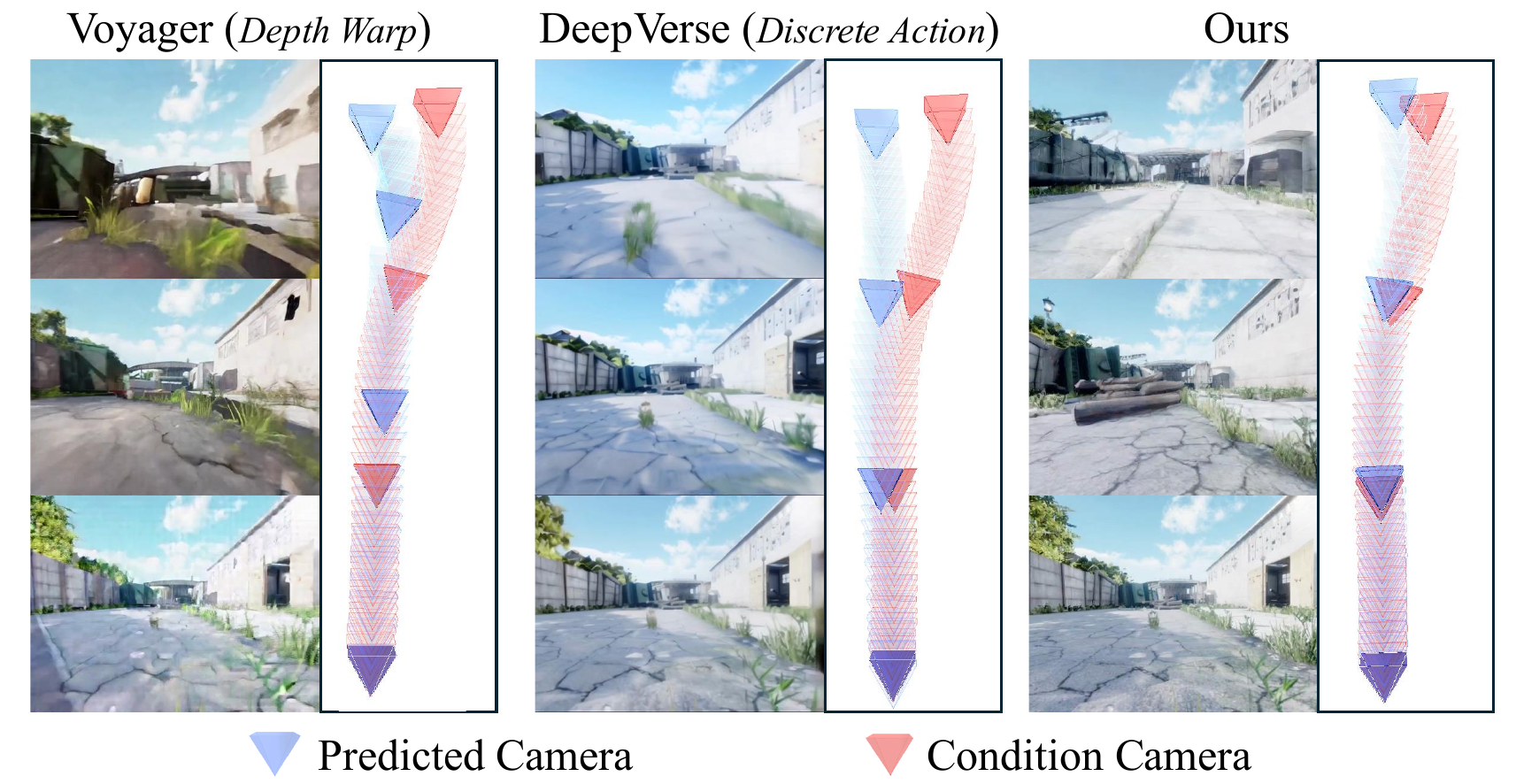}
    \vspace{-5mm}
    \caption{Visualization of camera trajectory comparison from methods with different camera conditioning types.}
    \vspace{-5mm}
    \label{fig:camera_traj}
\end{figure}
\begin{figure*}[t]
    \centering
    \setlength{\fboxrule}{0.5pt}
    \setlength{\fboxsep}{-0.01cm}
    \setlength\tabcolsep{0pt}
    \begin{spacing}{1}
    \begin{tabular}{p{0.04\linewidth}<{\centering}p{0.16\linewidth}<{\centering}p{0.16\linewidth}<{\centering}p{0.16\linewidth}<{\centering}p{0.16\linewidth}<{\centering}p{0.16\linewidth}<{\centering}p{0.16\linewidth}<{\centering}}
    
    & Left View & Disparity  & Left View & Disparity & Left View & Disparity \\

    \rotatebox{90}{ \hspace{-1mm} \small{Voyager~\cite{Huang2025VoyagerLA}}} &
    \includegraphics[width=1\linewidth, trim=0 0 0 0,clip]{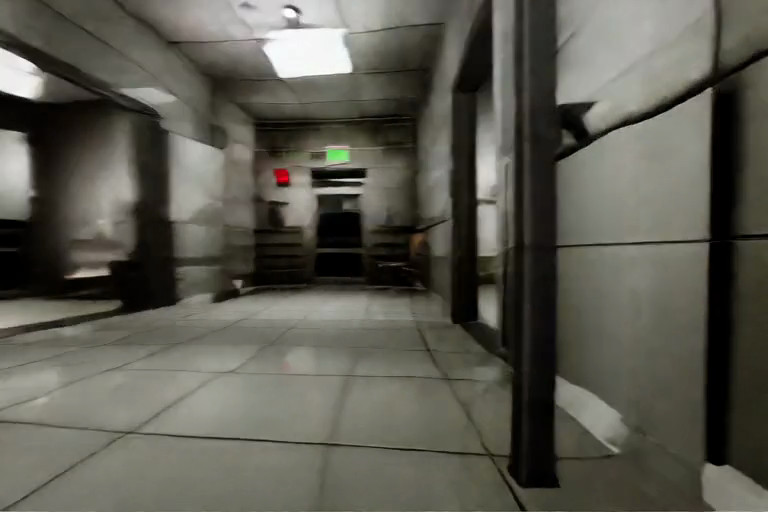} &
    \includegraphics[width=1\linewidth, trim=0 0 0 0,clip]{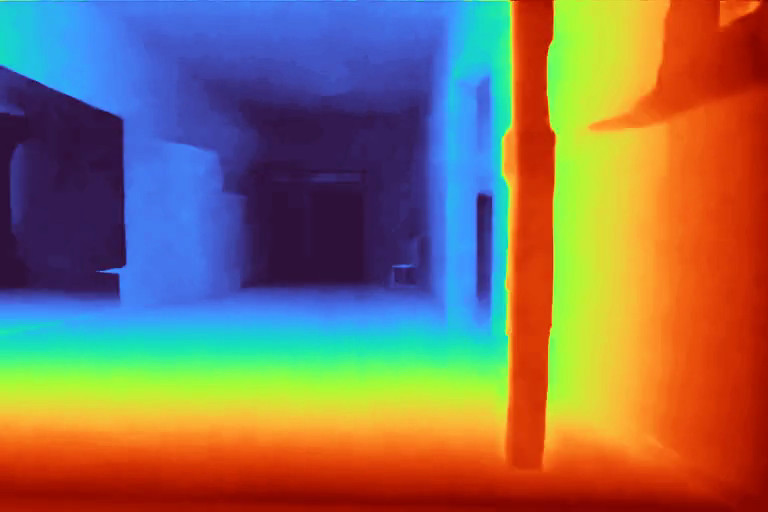} &
    \includegraphics[width=1\linewidth, trim=0 0 0 0,clip]{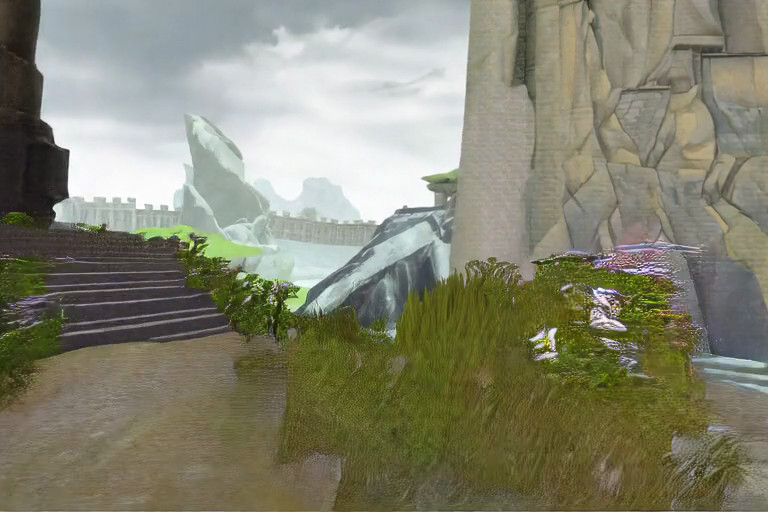} &
    \includegraphics[width=1\linewidth, trim=0 0 0 0,clip]{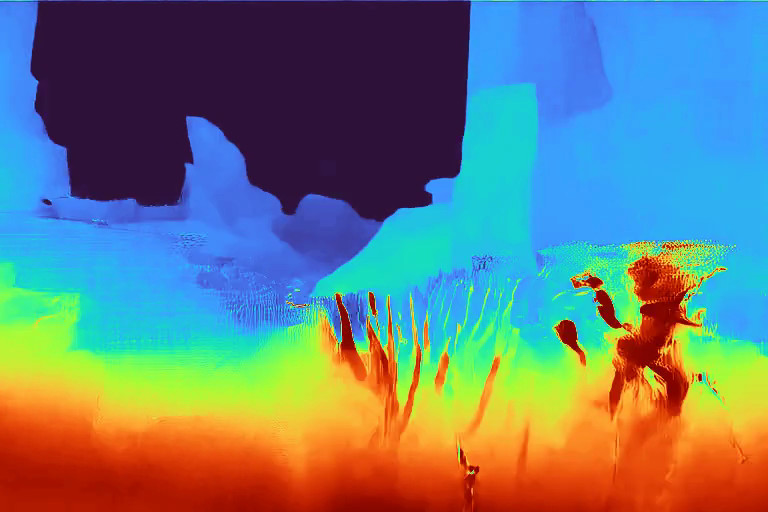} &
    \includegraphics[width=1\linewidth, trim=0 0 0 0,clip]{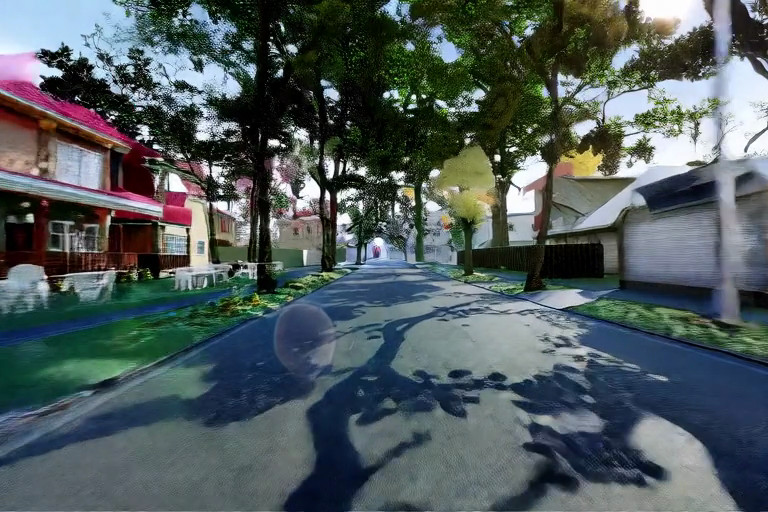} &
    \includegraphics[width=1\linewidth, trim=0 0 0 0,clip]{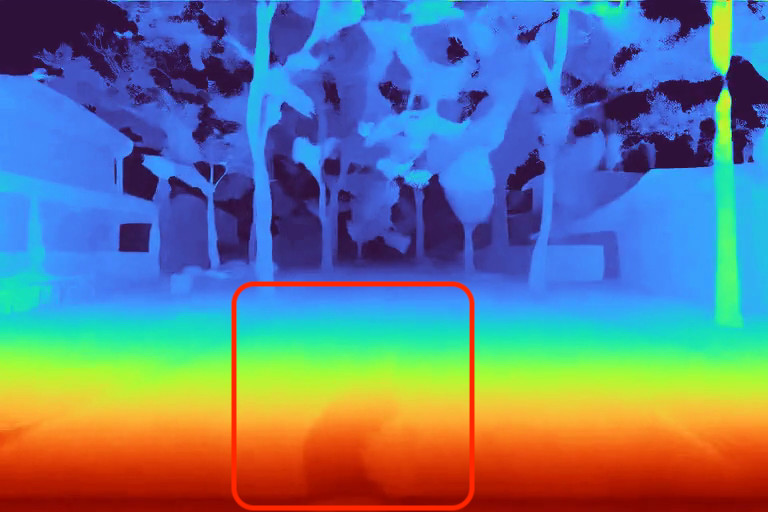}
    \\
    \specialrule{0em}{0pt}{-15pt} \\
    \rotatebox{90}{ \hspace{-1mm} \small{Aether~\cite{Team2025AetherGU}}} &
    \includegraphics[width=1\linewidth, trim=0 0 0 0,clip]{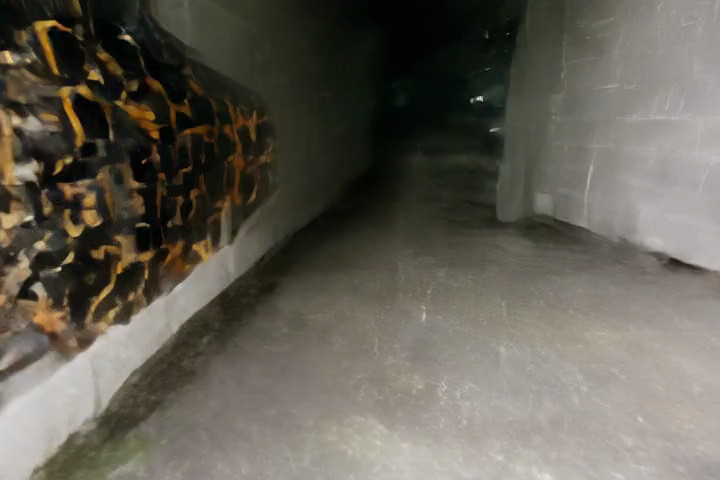} &
    \includegraphics[width=1\linewidth, trim=0 0 0 0,clip]{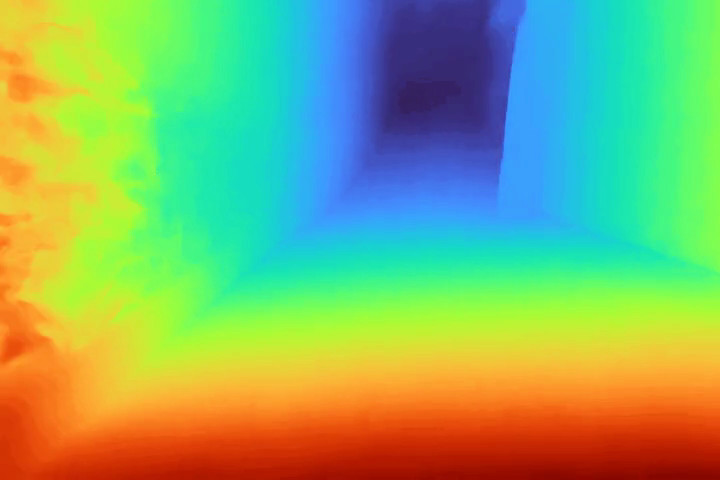} &
    \includegraphics[width=1\linewidth, trim=0 0 0 0,clip]{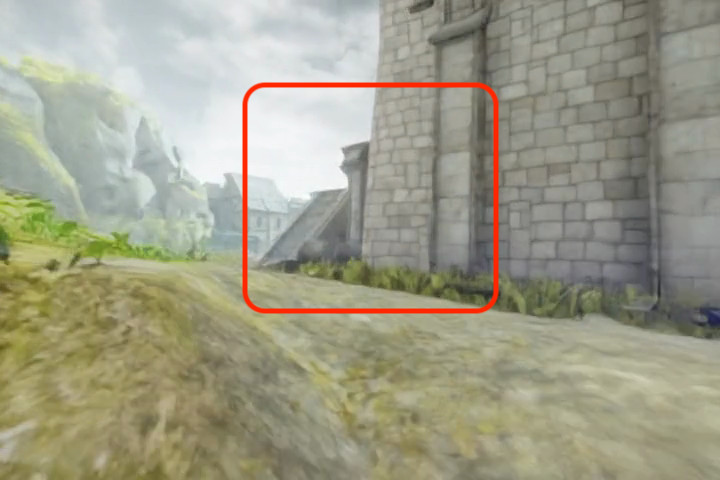} &
    \includegraphics[width=1\linewidth, trim=0 0 0 0,clip]{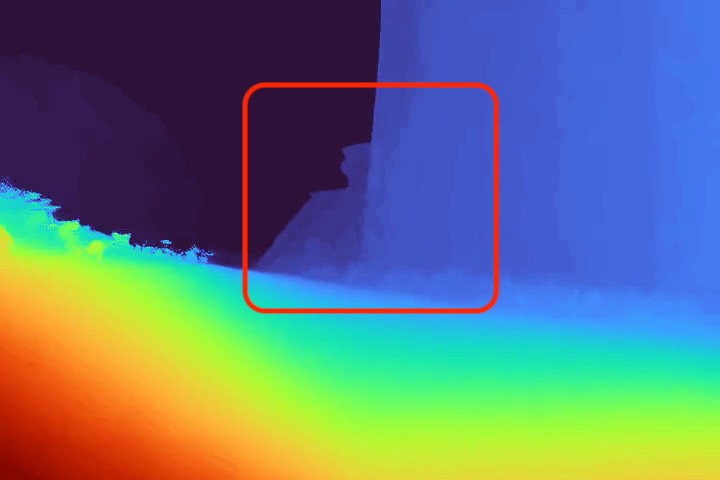} &
    \includegraphics[width=1\linewidth, trim=0 0 0 0,clip]{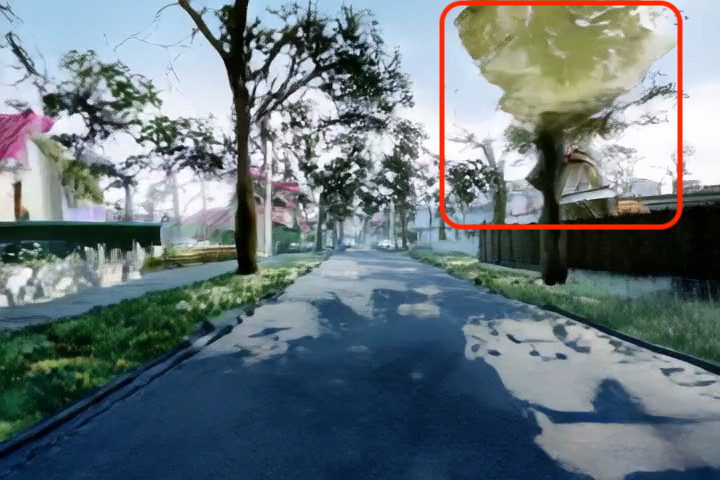} &
    \includegraphics[width=1\linewidth, trim=0 0 0 0,clip]{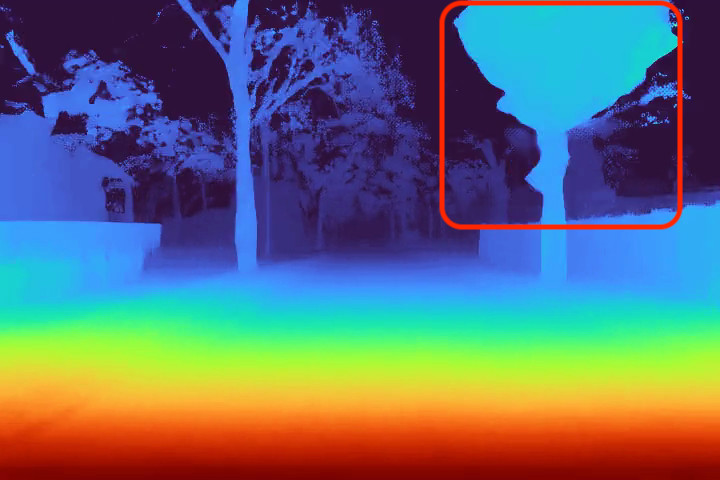}
    \\
    \specialrule{0em}{0pt}{-15pt} \\
    \rotatebox{90}{ \hspace{5mm} \small{Ours}} &
    \includegraphics[width=1\linewidth, trim=0 0 0 0,clip]{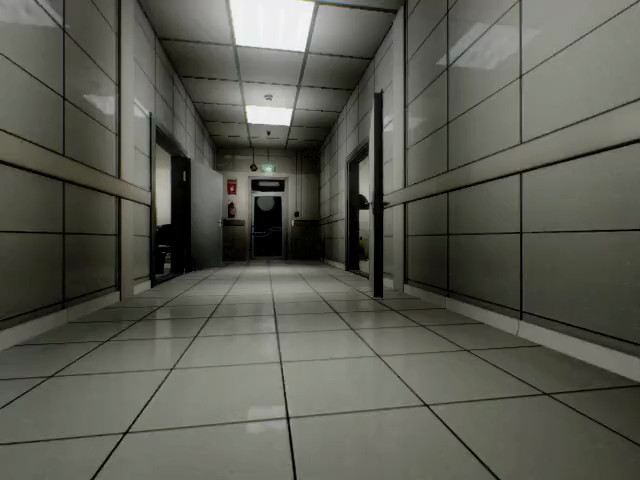} &
    \includegraphics[width=1\linewidth, trim=0 0 0 0,clip]{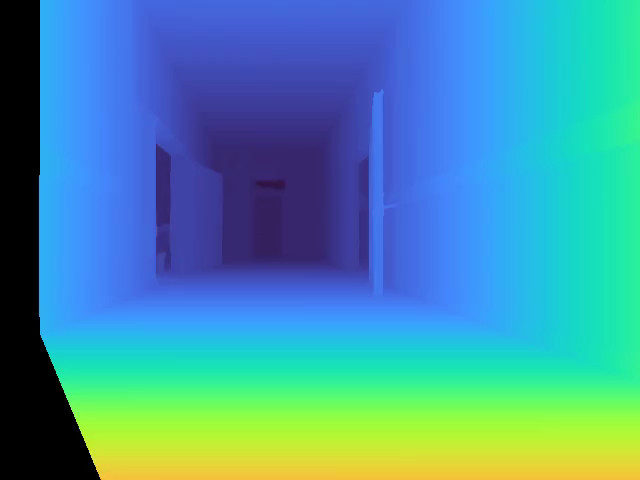} &
    \includegraphics[width=1\linewidth, trim=0 0 0 0,clip]{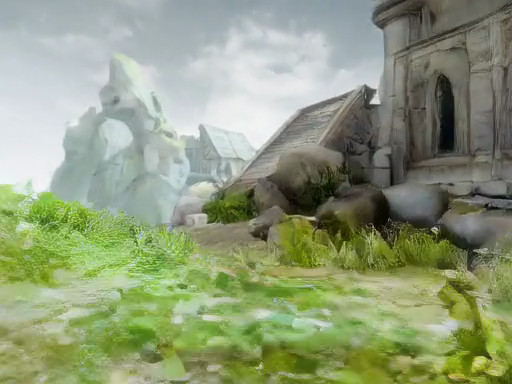} &
    \includegraphics[width=1\linewidth, trim=0 0 0 0,clip]{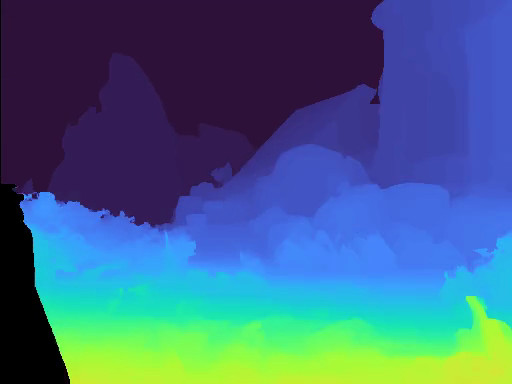} &
    \includegraphics[width=1\linewidth, trim=0 0 0 0,clip]{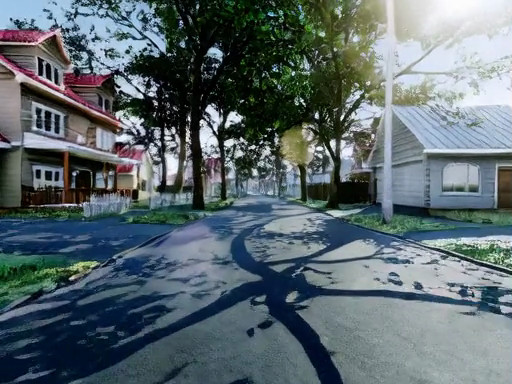} &
    \includegraphics[width=1\linewidth, trim=0 0 0 0,clip]{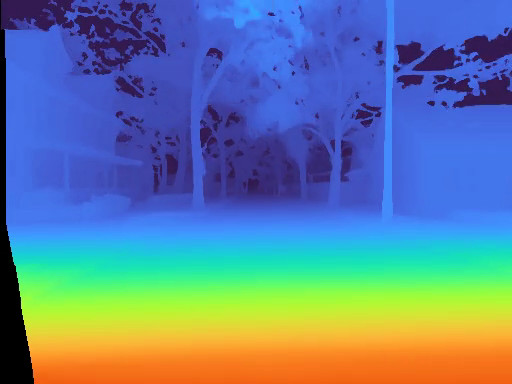}

    \end{tabular}
    \end{spacing}
    \vspace{-3mm}
    \caption{ Stereo disparity comparison. Notably, our approach does not rely on any depth supervision during training.}
    \vspace{-2mm}
    \label{fig:disp_comparison}
\end{figure*}

\subsection{Application}

\subsubsection{Virtual Reality Display}
\label{sec:vr}
Our method can be directly applied to VR/AR with professional head-mounted display devices. We visualize several red–blue anaglyph images in Fig.~\ref{fig:teaser} as examples of the generated stereo outputs. In addition, we conducted tests on a VR headset and performed a user study, the results of which are provided in the supplementary materials. 

\subsubsection{Embodided Scenarios}
\label{sec:embodiedAI}
To demonstrate the potential of our approach, we further evaluated it in embodied scenarios. Specifically, we fine-tuned our model on the binocular robotic arm dataset from DROID~\cite{khazatsky2024droid}. The trained model can generate corresponding stereo manipulation videos conditioned on a given text prompt, while also accurately recovering \emph{metric-scale} depth from the generated results. We illustrate the results in Fig.~\ref{fig:teaser} and supplementary materials.

\subsubsection{Long Video Distillation}
\label{sec:distillation}
Our method can also serve as a bidirectional attention base model for stereo video generation in an interactive causal manner, similar to Self-Forcing~\cite{huang2025selfforcing}. Specifically, we distill the diffusion sampling process into four steps and convert the model into a causal attention mechanism while maintaining a key–value (KV) cache. The distilled model is capable of generating 10-second stereo videos, with the generation speed improved from 0.49 FPS to 5.6 FPS. Additional technical details regarding long-video distillation are provided in the supplementary materials.

\subsection{Ablation}

\paragraph{Camera Injection.} We compare different camera conditioning strategies on TartanAir dataset~\cite{Wang2020TartanAirAD}, reported in Tab.~\ref{tab:camera_injection_ablation} and Fig.~\ref{fig:camera_ablation}. Among them, Ours (Zero Init) preserves the pretrained model’s prior and achieves relatively high visual quality. However, because the weights are initialized to zeros, the learning of camera conditioning becomes more difficult, leading to lower camera accuracy. The {Pl{\"u}kcer Ray~\cite{zhang2024cameras}} approach, which relies on absolute coordinates, shows limited generalization capability and suffers a performance drop. Compared with {PRoPE~\cite{li2025cameras}} , our method better preserves the pretrained model prior, achieving superior results in both visual fidelity and camera conformity.
\begin{table}[t]
  \vspace{-3.5mm}
  \centering
    \caption{Ablation on camera injection strategies.}
    \vspace{-3mm}
  \adjustbox{width={\linewidth},keepaspectratio}{
    \begin{tabular}{l|cccc|cc}
    \bottomrule

         \multirow{2}{*}{\textbf{Method}}  & \multicolumn{4}{c}{Visual Quality} & \multicolumn{2}{c}{Camera Accuracy}  \\

          & {\textit{\small{FID}}}$\downarrow$ & {\small{{FVD}}}$\downarrow$ & {\small{CLIP-T}}$\uparrow$ & {\small{CLIP-F}}$\downarrow$ & {\small{\textit{RotErr}}}$\downarrow$ & {\small{TransErr}}$\uparrow$   \\
        
        \hline
        {Pl{\"u}kcer Ray~\cite{zhang2024cameras}}   
        & {142.46} & {130.39} & {24.90} & {95.65}  & {1.52} & {0.21} \\
        {PRoPE~\cite{li2025cameras}}   
        & {144.45} & {128.32} & {25.33} & {96.83}  & {1.33} & {0.18} \\
        {Ours Zero Init}
        & {131.07} & {96.62} & {25.49} & {97.21}  & {1.81} & {0.24} \\
        {Ours Copy Init}  
        & \textbf{122.41} & \textbf{93.17} & \textbf{25.54} & \textbf{97.26}  & \textbf{1.16} & \textbf{0.15} \\
    \bottomrule
    \end{tabular}}
    \label{tab:camera_injection_ablation}
  \vspace{-1.5mm}
\end{table}
\begin{table}[t]
  \centering
    \caption{Ablation on attention scheme.}
    \vspace{-3mm}
  \adjustbox{width={\linewidth},keepaspectratio}{
    \begin{tabular}{l|cc|cc|cc}
    \bottomrule
         \multirow{2}{*}{\textbf{Method}}  & \multicolumn{2}{c|}{Visual Quality} & \multicolumn{2}{c|}{View Synchronization} & \multicolumn{2}{c}{Efficiency}  \\
          & {{\small{CLIP-T}}}$\uparrow$ & {\small{{CLIP-V}}}$\uparrow$ & {\small{Mat. Pix.(K)}}$\uparrow$ & {\small{CLIP-V}}$\uparrow$ &  {\small{FLOPs($\times 10^{10}$)}}$\downarrow$ & {\small{FPS}}$\uparrow$ \\
        
        \hline
        {4D Attn}   
        & {25.74} & {97.55} & {4.51} & {97.50} & {3.11} & {0.34} \\
        {Stereo Attn}  
        & {25.43} & {97.05} & {4.52} & {96.63} & {1.56} & {0.49} \\
    \bottomrule
    \end{tabular}}
    \label{tab:attention_ablation}
  \vspace{-5.5mm}
\end{table}

\vspace{-6mm}
\paragraph{Attention Scheme.} We also compare the impact of different attention mechanisms on the results. As shown in Tab.~\ref{tab:attention_ablation}, although 4D Attention achieves slightly better visual quality, the performance of Stereo Attention is largely comparable—and even surpasses it in terms of view consistency. Meanwhile, FLOPs and FPS are improved by approximately 50\%, demonstrating the efficiency of our design. The detailed FLOPs calculations are provided in the supplementary materials.

    



\begin{figure}[t]
    \vspace{-3mm}
    \centering
    \includegraphics[width=1\linewidth, trim=0 0 0 0,clip]{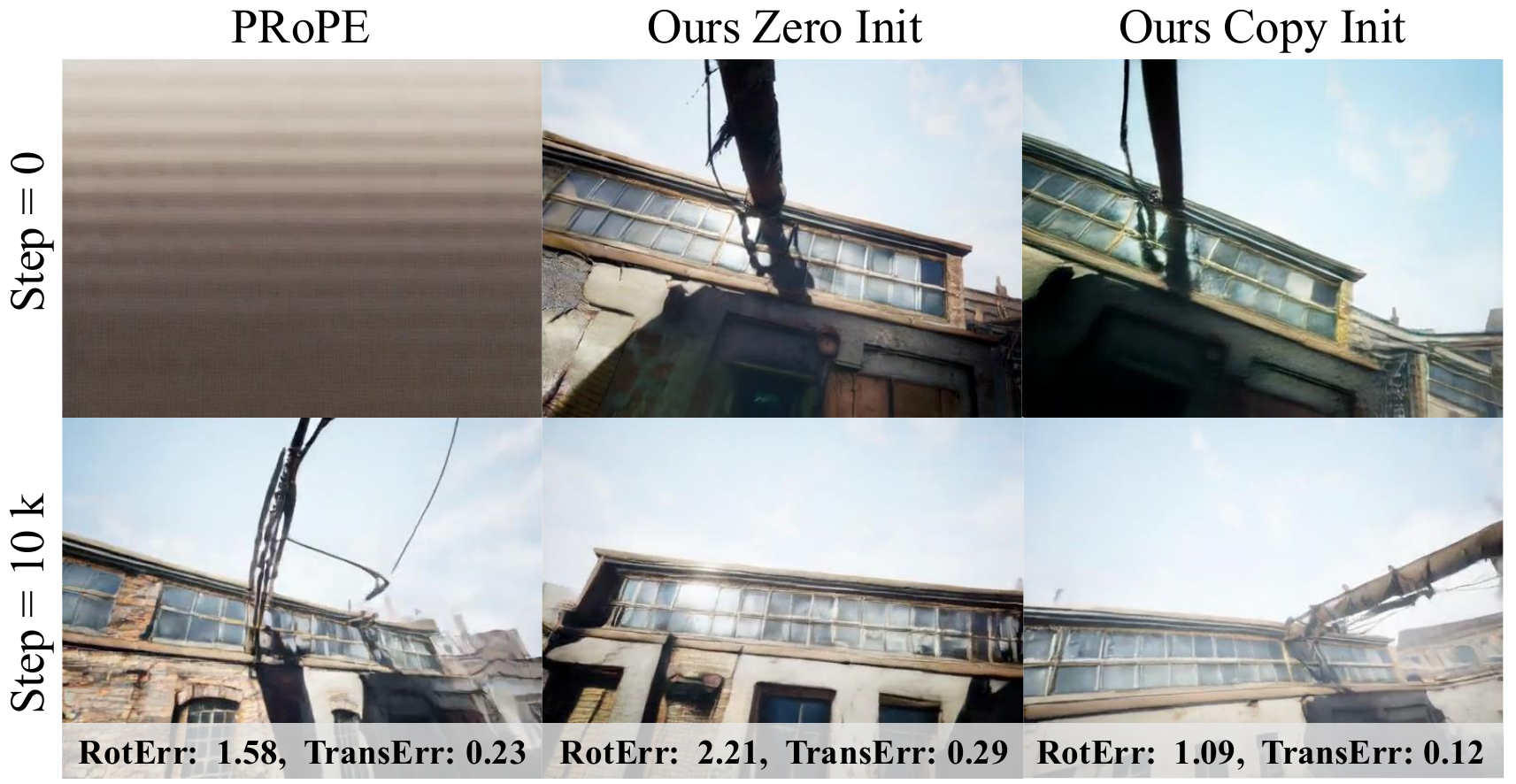}
      \vspace{-5mm}
    \caption{Comparison of different camera-condition strategies.}
    \vspace{-5mm}
    \label{fig:camera_ablation}
\end{figure}

\section{Conclusion and Discussion}
This paper presents \textbf{StereoWorld}, a camera-conditioned stereo vision model that jointly modeling binocular visual appearance, while supporting explicit geometry grounding. By employing a unified camera-frame Rotary Position Embedding (RoPE), the model encodes relative camera parameters effectively, with minimal interference to pre-trained priors. Furthermore, we introduce a stereo-aware attention mechanism that exploits the inherent horizontal epipolar constraints in stereo videos to reduce computational complexity.
Experimental results demonstrate that \textbf{StereoWorld} achieves more efficient and view-consistent outcomes in stereo video generation, with strong potential for downstream applications such as virtual reality, embodied AI, and long-horizon video synthesis.

Despite these advances, stereo video generation remains more computationally demanding than its monocular counterparts, and the scarcity of large-scale stereo datasets further limits model scalability. We provide a detailed discussion of these limitations and potential future research directions in the supplementary materials.


{
    \small
    \bibliographystyle{ieeenat_fullname}
    \bibliography{main}
}
\appendix
\renewcommand{\thesection}{S\arabic{section}} 
\renewcommand{\thesubsection}{\thesection.\arabic{subsection}} 

\clearpage
\setcounter{page}{1}
\maketitlesupplementary

\section{Experiment}

\subsection{Dataset Construction}
The datasets used for training are summarized in Tab.1 of the main paper. For Stereo4D~\cite{Jin2024Stereo4DLH}, we filtered out videos in which the camera remained static, exhibited minimal motion, or suffered from excessive jitter, as such samples are unsuitable for camera-conditioned training. Each video was divided into 49-frame clips, which were then cropped and resized to a uniform resolution of $480 \times 640$. For each clip, we used the left-eye video to generate caption annotations. All training data were accompanied by metric-scale camera parameters.

For the test set, we selected approximately 280 video clips from the processed TartanAirGround~\cite{patel2025tartanground} video clips, sampled at intervals of 200. In addition, we used the UnrealStereo4K~\cite{Tosi2021SMDNetsSM} and Middlebury~\cite{Scharstein2014HighResolutionSD} stereo image datasets, for which we generated a set of random camera trajectories to conduct out-of-domain evaluations (approximately 160 clips). Each camera trajectory was composed of both translation and rotation components. The translation sampling range along the z-axis was $[-20m, -4m] \cup [4m, 20m]$, and the rotation sampling range around the y-axis was  $[-150^\circ, -50^\circ] \cup [50^\circ,150^\circ]$.

\subsection{Stereo Attention FLOPs}
For each attention head, let $L$ be the sequence length or number of query tokens, $d$ be the head dimension, a vanilla full attention head costs:
\begin{equation}
    \text{FLOPs}_\text{full} = 4L^2d.
\end{equation}

In our experiment, the input feature has the shape $f \in \mathbb{R}^{b \times 2f \times h \times w \times c}$. As for 4D Attention, $L = 2f \times h \times w$, we have
\begin{equation}
    \text{FLOPs}_{\text{Attn}_\text{4D}} = 16bf^2h^2w^2d.
\end{equation}
While for the stereo attention, we have
\begin{align}
    \text{FLOPs}_{\text{Attn}_\text{3D}} = 8bf^2h^2w^2d,
\end{align}
\begin{align}
    \text{FLOPs}_{\text{Attn}_\text{row}} = 4bfhw^2d.
\end{align}

Supposing we use $b=1$, $f=13$, $h=15$, $w=20$, $d=128$, we can calculate that $\text{FLOPs}_{\text{Attn}_\text{4D}} \approx 3.115 \times 10^{10}$, while in comparison, the stereo attention costs $\text{FLOPs}_{\text{Attn}_\text{3D}} + \text{FLOPs}_{\text{Attn}_\text{row}} = 1.561 \times 10^{10}$. Hence the stereo attention block reduces multiply-adds by a factor about $\mathbf{2} \times$.


\section{Application}
\begin{figure}[t]
    \centering
    \includegraphics[width=1\linewidth, trim=0 0 0 0,clip]{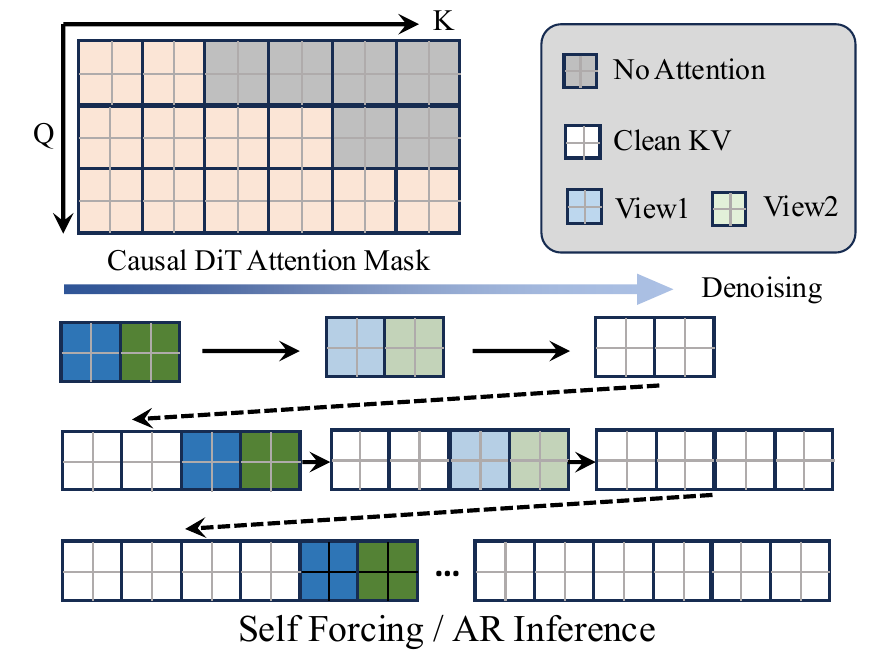}
    \caption{Attention mask configuration in distillation process.}
    \label{fig:distillation}
\end{figure}

\subsection{VR/AR Display}
\begin{figure*}[t]
    \centering
    \setlength{\fboxrule}{0.5pt}
    \setlength{\fboxsep}{-0.01cm}
    \setlength\tabcolsep{0pt}
    \begin{spacing}{1}
    \begin{tabular}{p{0.04\linewidth}<{\centering}p{0.16\linewidth}<{\centering}p{0.16\linewidth}<{\centering}p{0.16\linewidth}<{\centering}p{0.16\linewidth}<{\centering}p{0.16\linewidth}<{\centering}p{0.16\linewidth}<{\centering}}
    
    & Left View & Right View & Anaglyph & Right View & Left View & Anaglyph \\
    \rotatebox{90}{ \hspace{7mm} \small{t1}} &
    \includegraphics[width=1\linewidth, trim=0 0 0 0,clip]{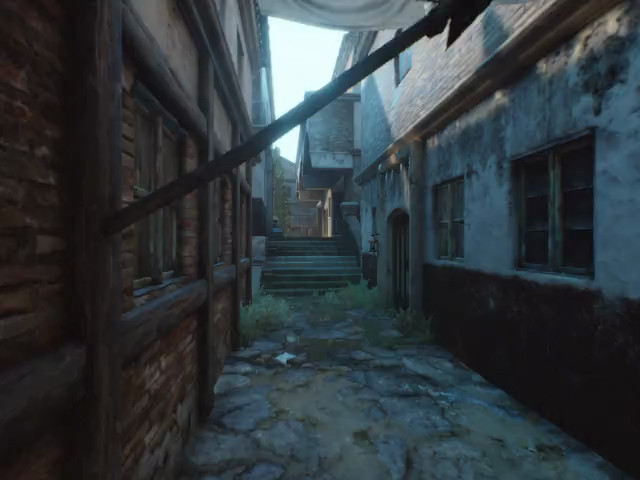} &
    \includegraphics[width=1\linewidth, trim=0 0 0 0,clip]{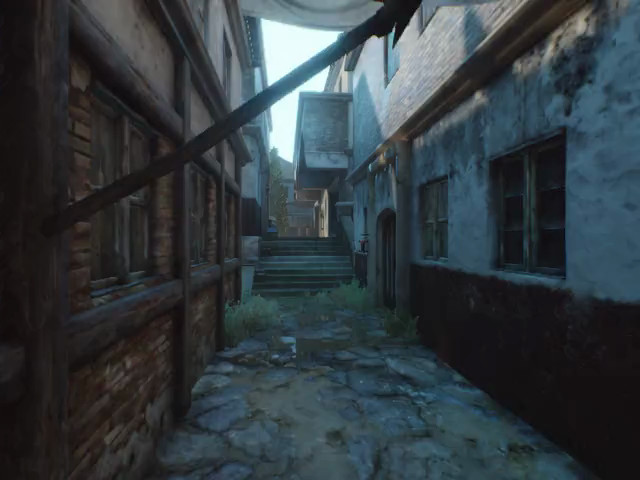} &
    \includegraphics[width=1\linewidth, trim=0 0 0 0,clip]{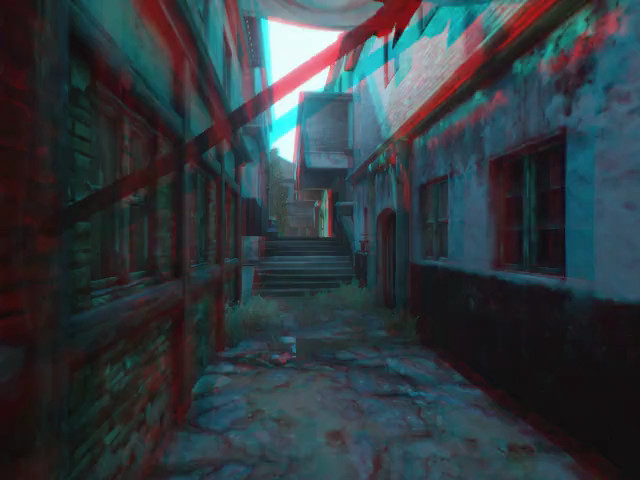} &
    \includegraphics[width=1\linewidth, trim=0 0 0 0,clip]{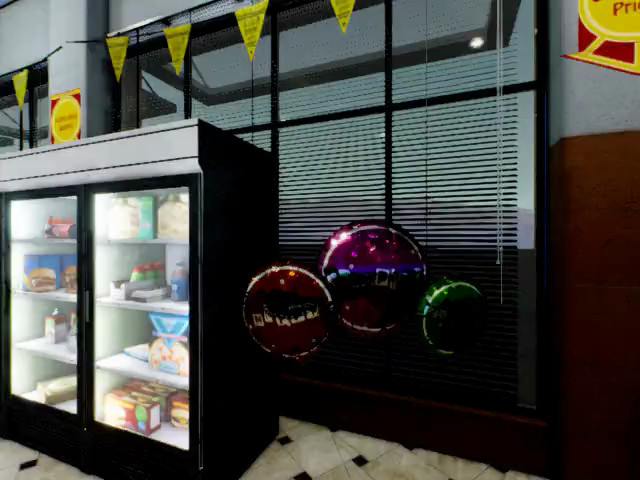} &
    \includegraphics[width=1\linewidth, trim=0 0 0 0,clip]{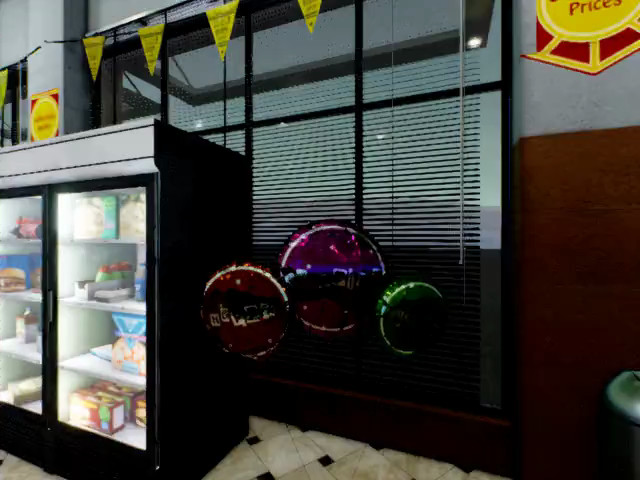} &
    \includegraphics[width=1\linewidth, trim=0 0 0 0,clip]{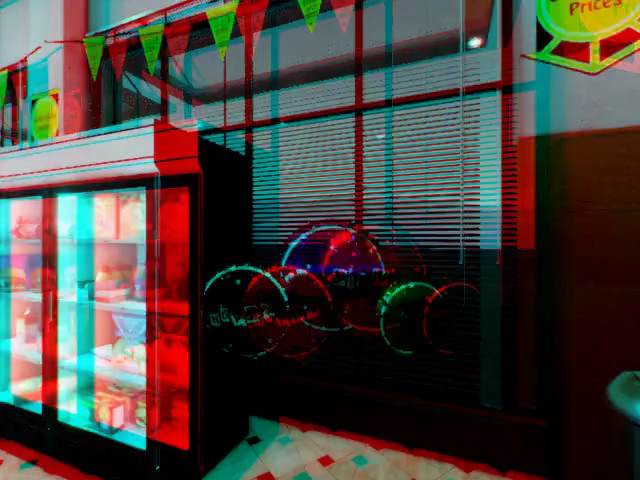}
    \\
    \rotatebox{90}{ \hspace{7mm} \small{t2}} &
    \includegraphics[width=1\linewidth, trim=0 0 0 0,clip]{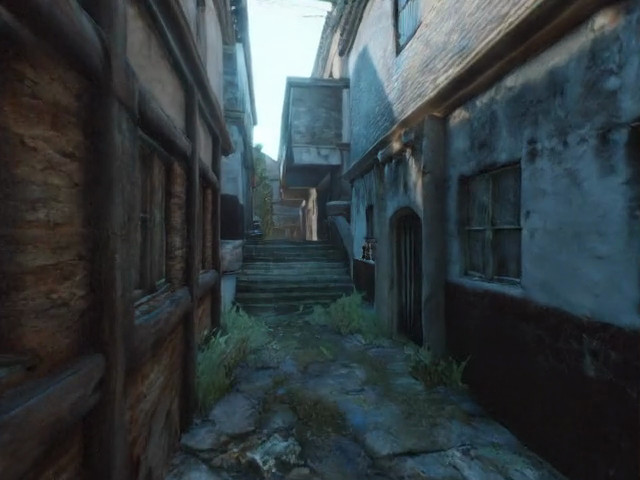} &
    \includegraphics[width=1\linewidth, trim=0 0 0 0,clip]{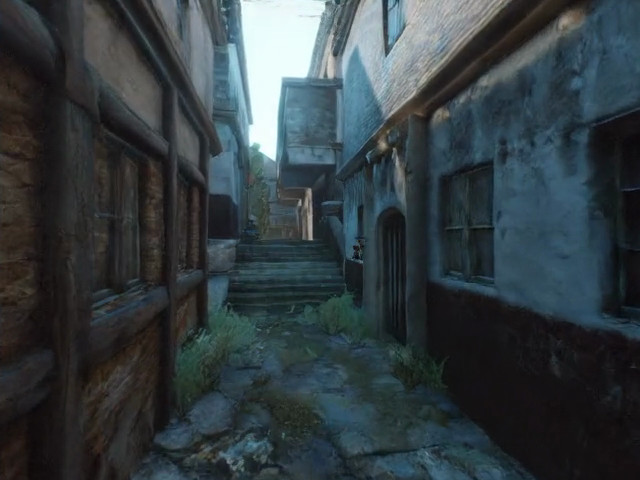} &
    \includegraphics[width=1\linewidth, trim=0 0 0 0,clip]{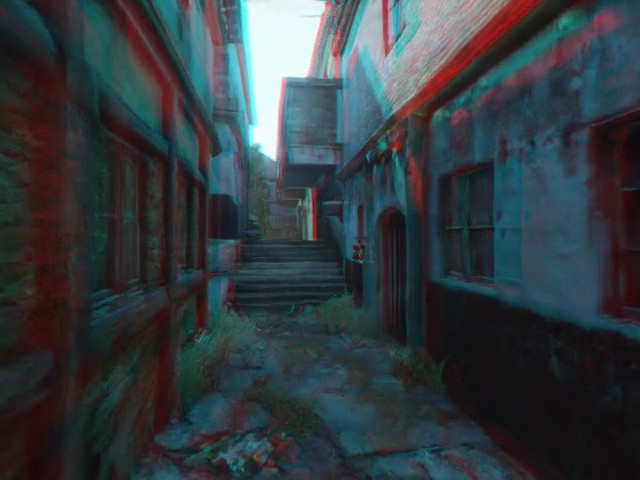} &
    \includegraphics[width=1\linewidth, trim=0 0 0 0,clip]{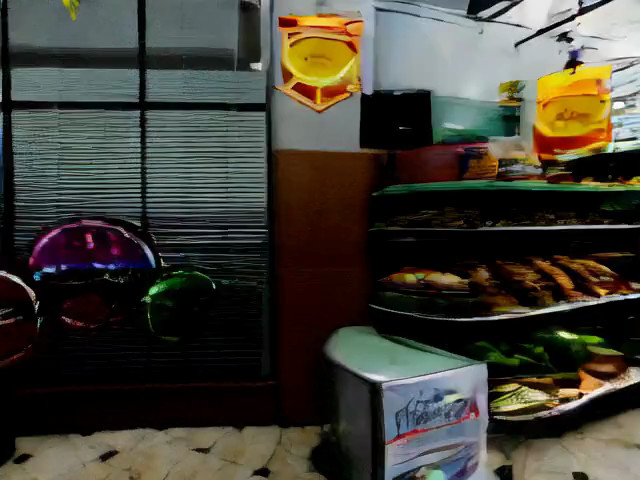} &
    \includegraphics[width=1\linewidth, trim=0 0 0 0,clip]{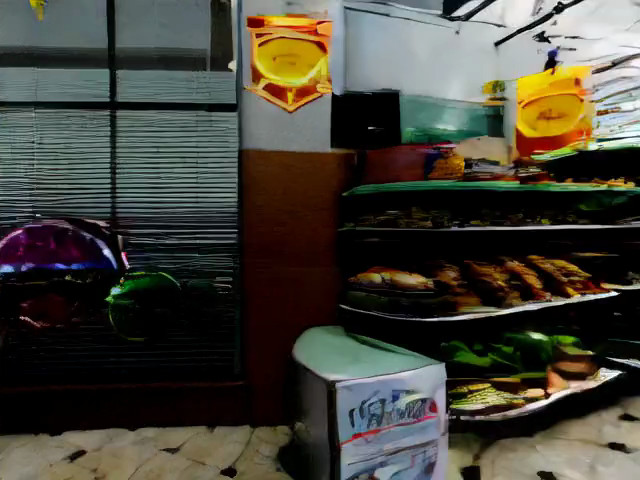} &
    \includegraphics[width=1\linewidth, trim=0 0 0 0,clip]{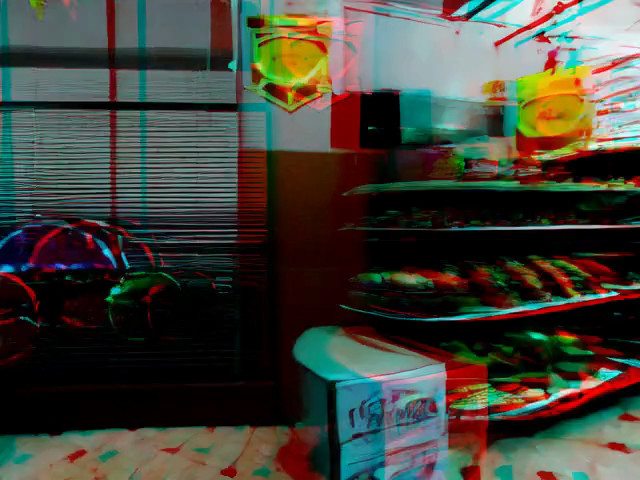}
    \\
    \rotatebox{90}{ \hspace{7mm} \small{t3}} &
    \includegraphics[width=1\linewidth, trim=0 0 0 0,clip]{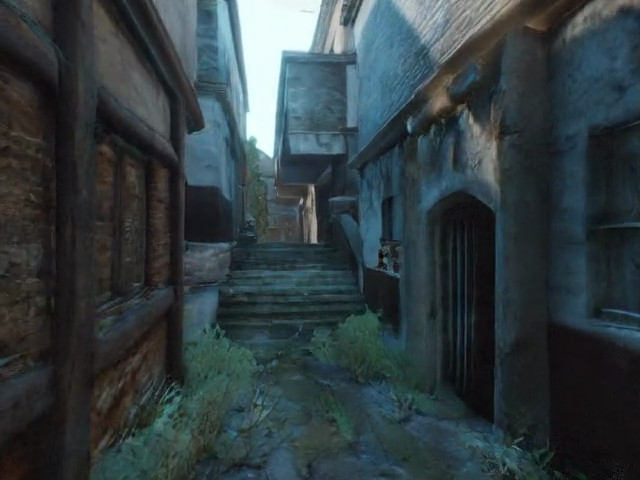} &
    \includegraphics[width=1\linewidth, trim=0 0 0 0,clip]{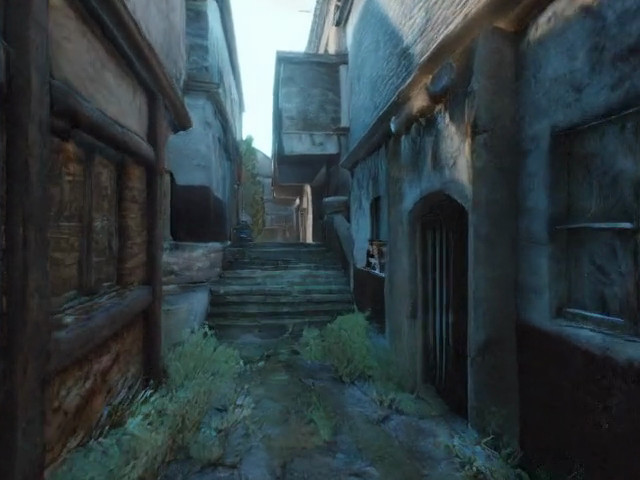} &
    \includegraphics[width=1\linewidth, trim=0 0 0 0,clip]{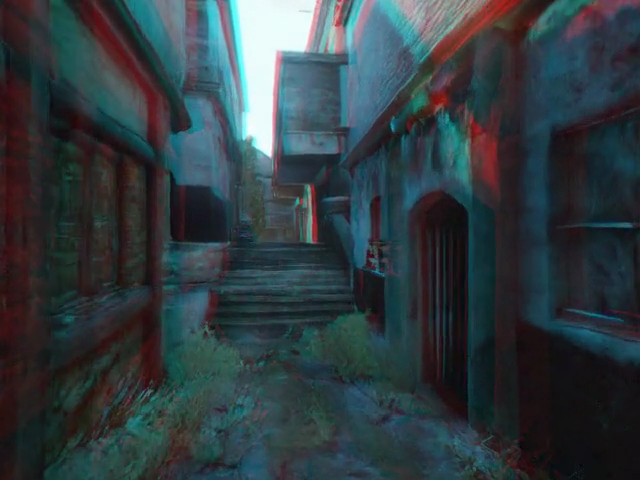} &
    \includegraphics[width=1\linewidth, trim=0 0 0 0,clip]{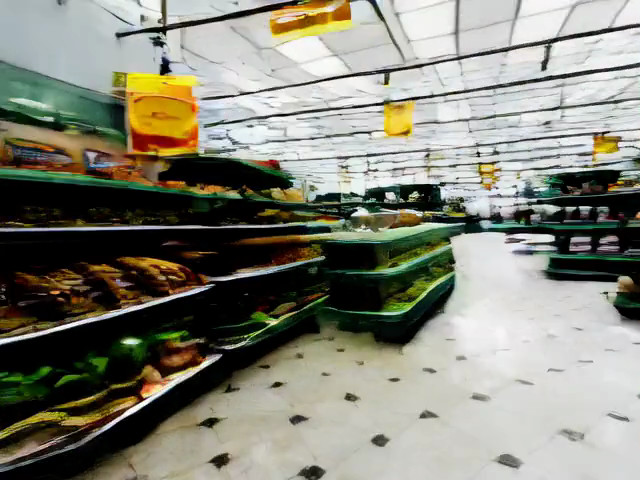} &
    \includegraphics[width=1\linewidth, trim=0 0 0 0,clip]{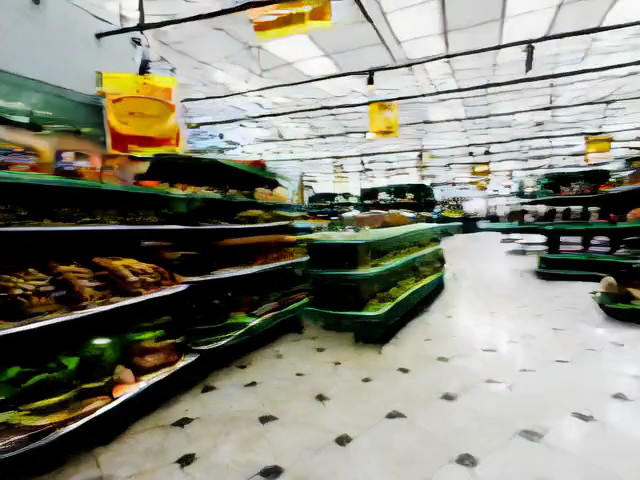} &
    \includegraphics[width=1\linewidth, trim=0 0 0 0,clip]{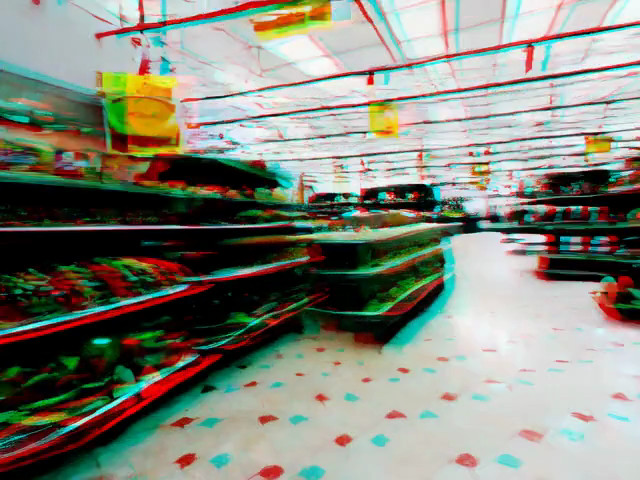}
    \\
    & Left View & Right View & Anaglyph & Right View & Left View & Anaglyph \\
    \rotatebox{90}{ \hspace{7mm} \small{t1}} &
    \includegraphics[width=1\linewidth, trim=0 0 0 0,clip]{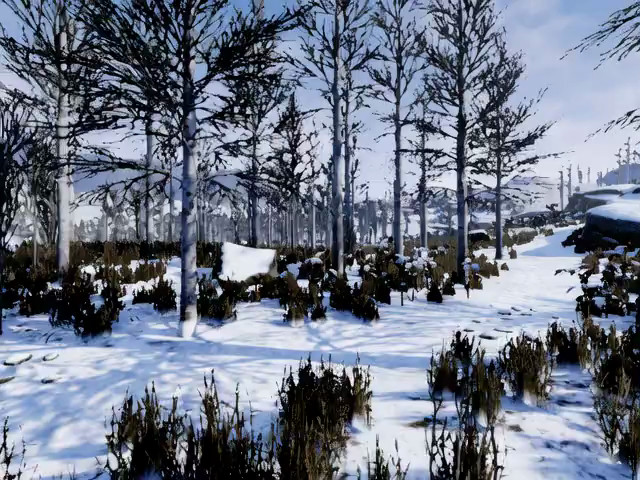} &
    \includegraphics[width=1\linewidth, trim=0 0 0 0,clip]{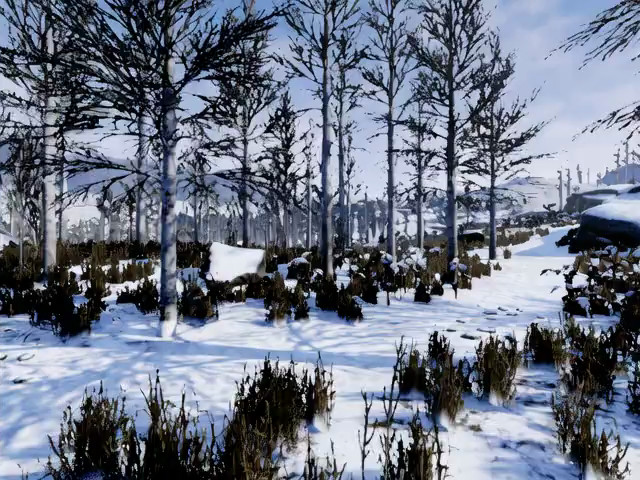} &
    \includegraphics[width=1\linewidth, trim=0 0 0 0,clip]{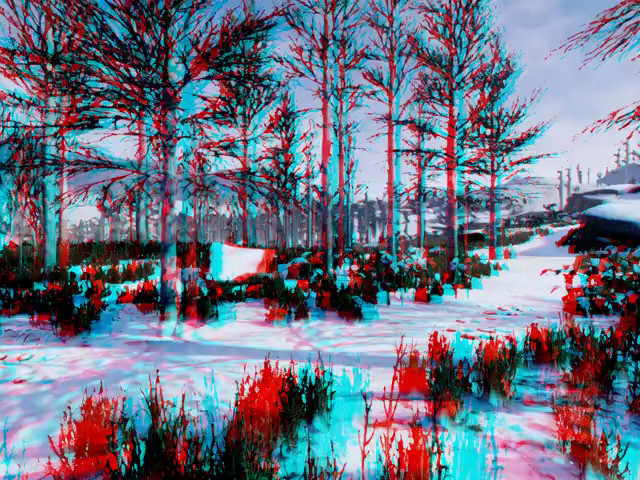} &
    \includegraphics[width=1\linewidth, trim=0 0 0 0,clip]{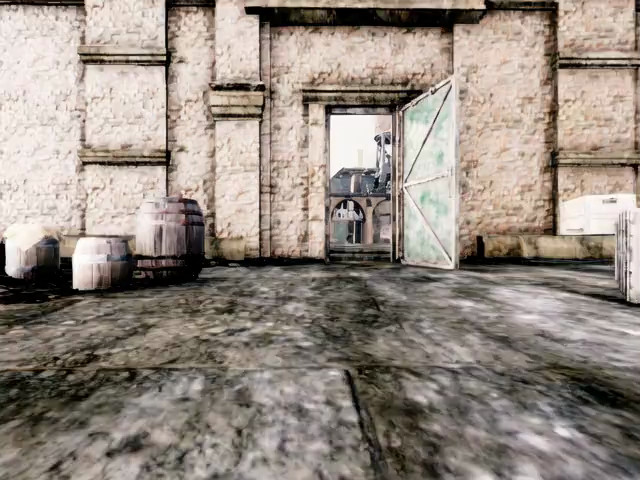} &
    \includegraphics[width=1\linewidth, trim=0 0 0 0,clip]{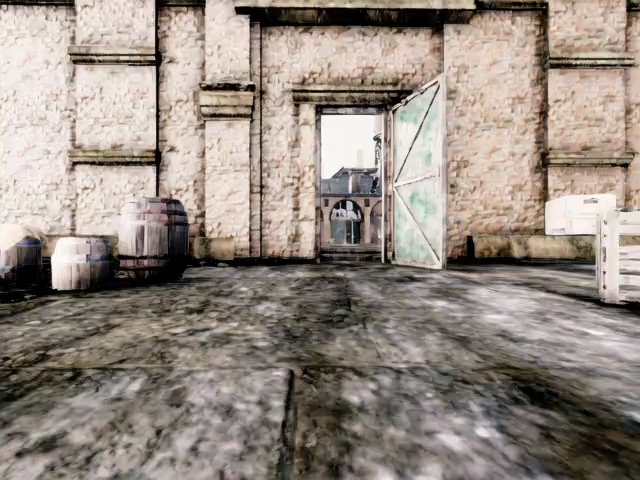} &
    \includegraphics[width=1\linewidth, trim=0 0 0 0,clip]{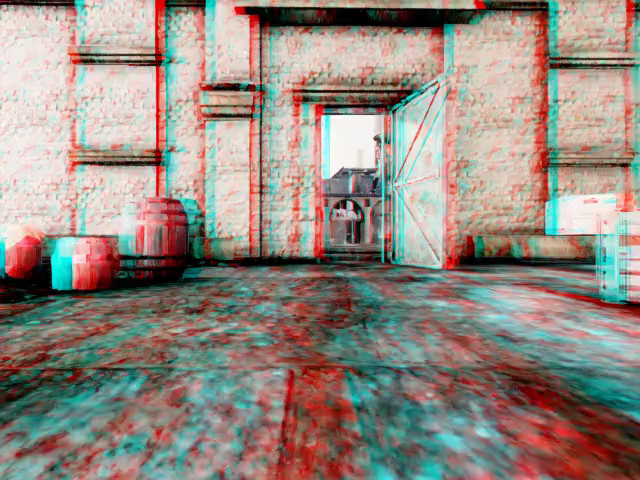}
    \\
    \rotatebox{90}{ \hspace{7mm} \small{t2}} &
    \includegraphics[width=1\linewidth, trim=0 0 0 0,clip]{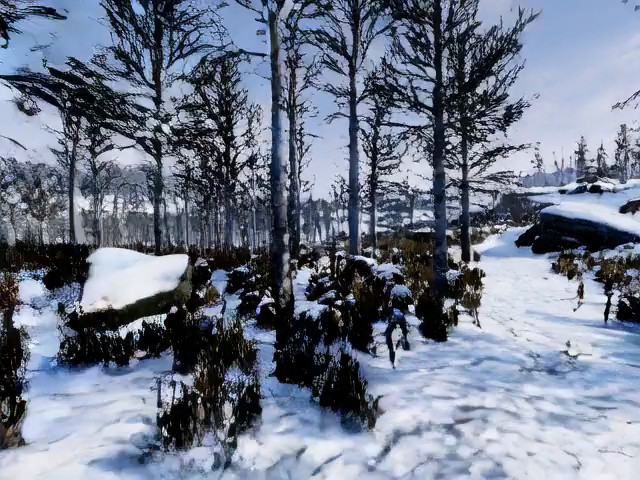} &
    \includegraphics[width=1\linewidth, trim=0 0 0 0,clip]{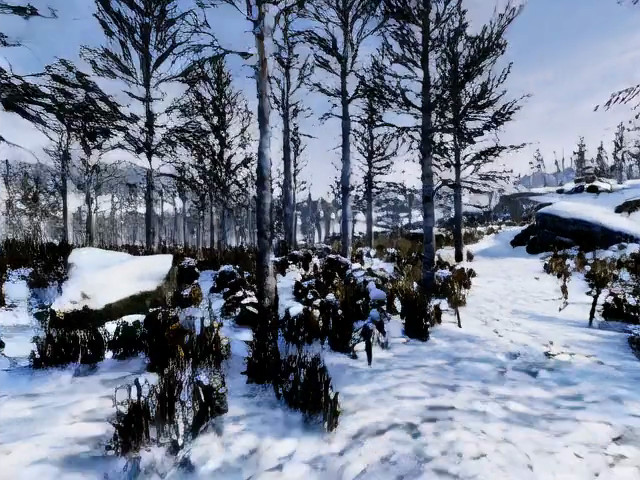} &
    \includegraphics[width=1\linewidth, trim=0 0 0 0,clip]{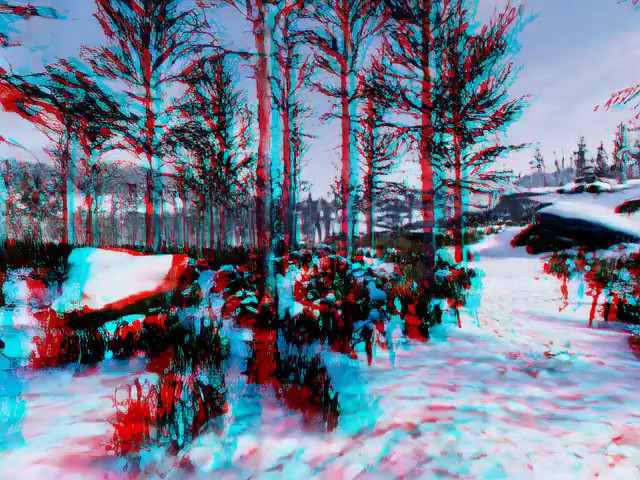} &
    \includegraphics[width=1\linewidth, trim=0 0 0 0,clip]{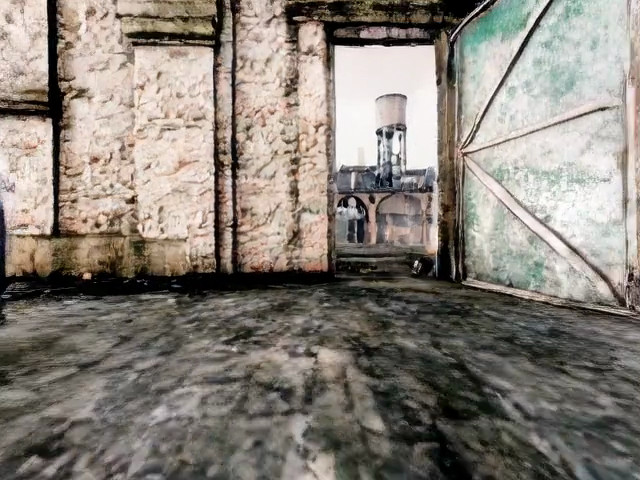} &
    \includegraphics[width=1\linewidth, trim=0 0 0 0,clip]{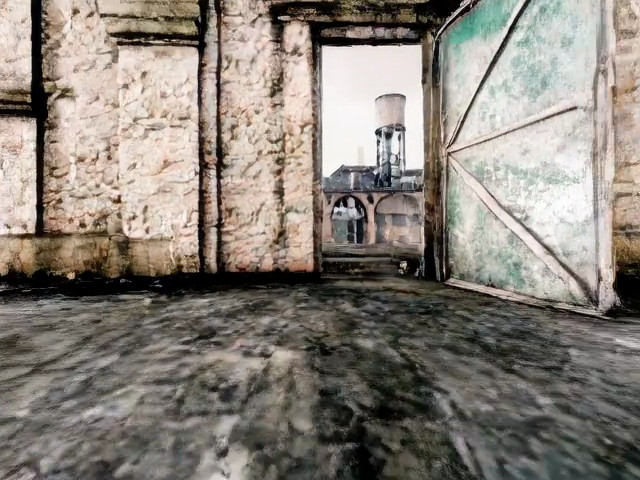} &
    \includegraphics[width=1\linewidth, trim=0 0 0 0,clip]{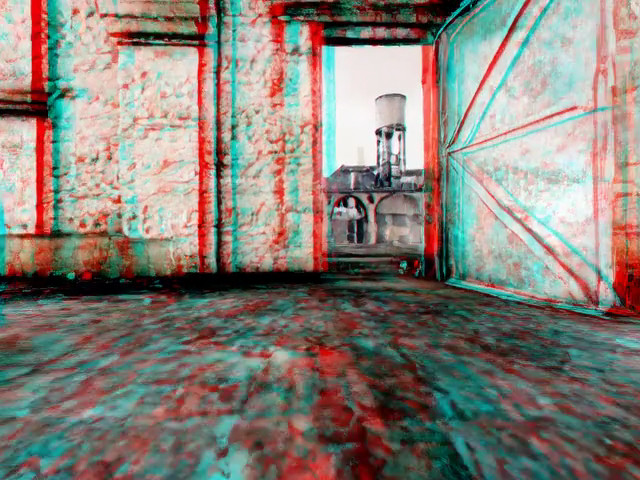}
    \\
    \rotatebox{90}{ \hspace{7mm} \small{t3}} &
    \includegraphics[width=1\linewidth, trim=0 0 0 0,clip]{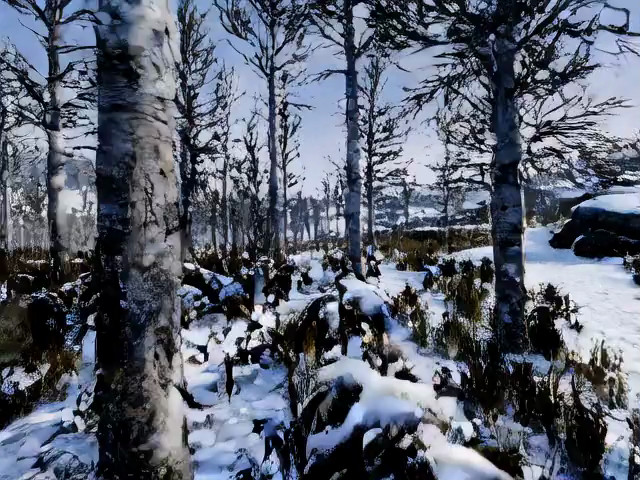} &
    \includegraphics[width=1\linewidth, trim=0 0 0 0,clip]{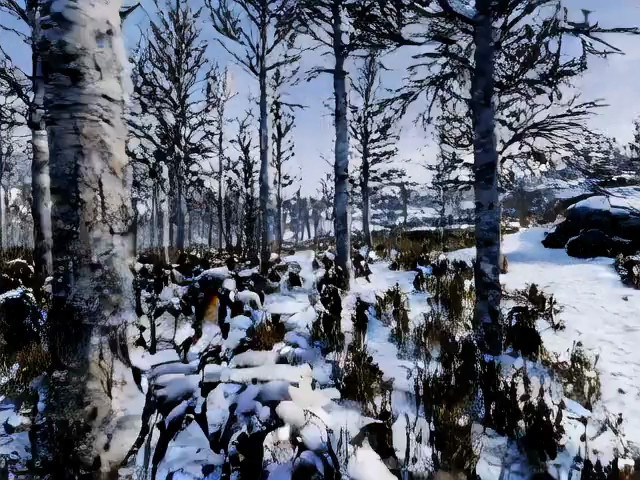} &
    \includegraphics[width=1\linewidth, trim=0 0 0 0,clip]{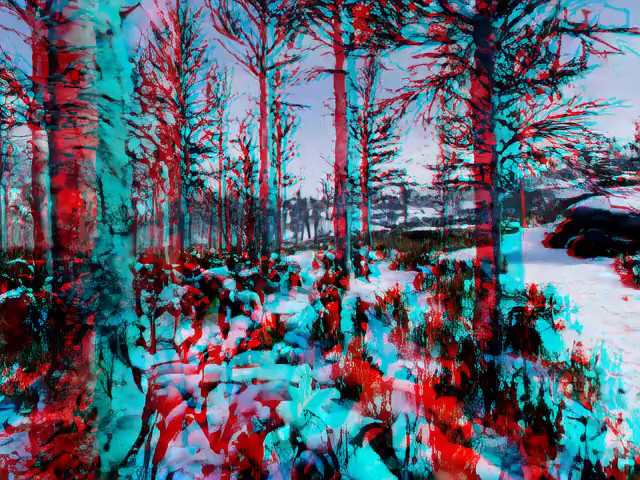} &
    \includegraphics[width=1\linewidth, trim=0 0 0 0,clip]{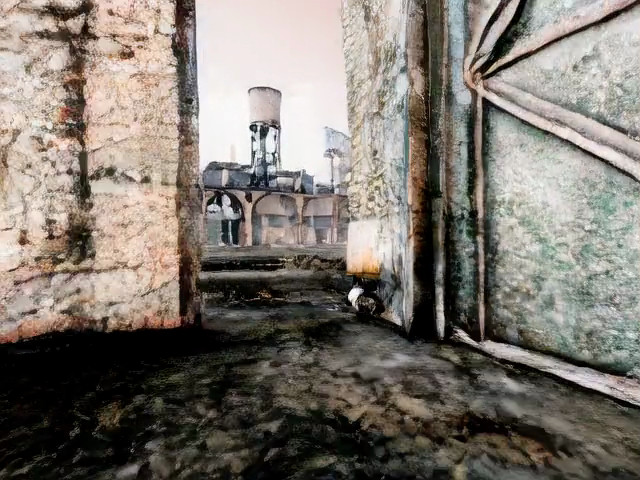} &
    \includegraphics[width=1\linewidth, trim=0 0 0 0,clip]{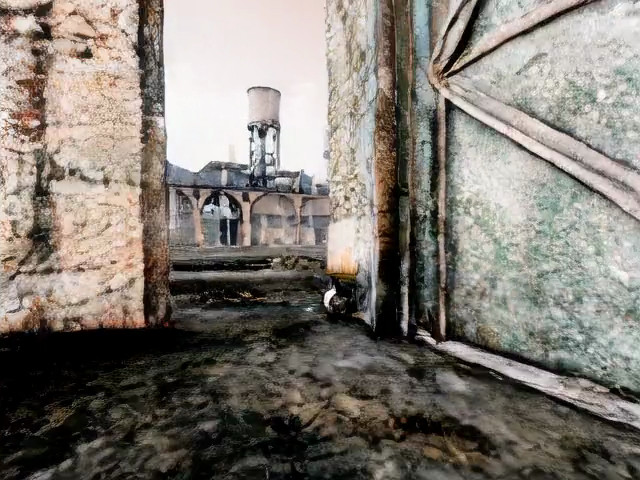} &
    \includegraphics[width=1\linewidth, trim=0 0 0 0,clip]{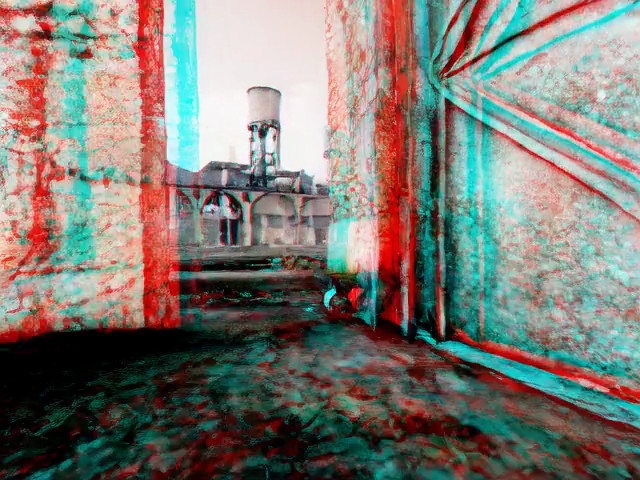}
    \\
    & Left View & Right View & Anaglyph & Right View & Left View & Anaglyph \\
    \rotatebox{90}{ \hspace{7mm} \small{t1}} &
    \includegraphics[width=1\linewidth, trim=0 0 0 0,clip]{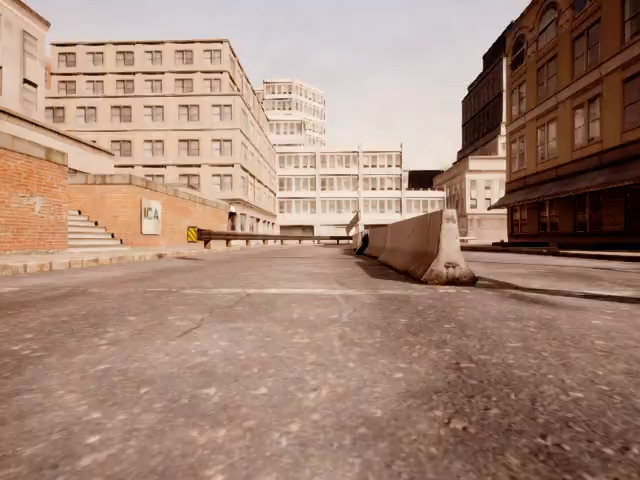} &
    \includegraphics[width=1\linewidth, trim=0 0 0 0,clip]{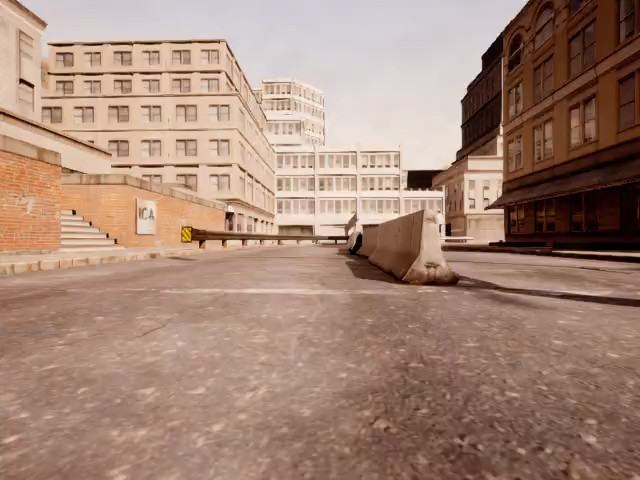} &
    \includegraphics[width=1\linewidth, trim=0 0 0 0,clip]{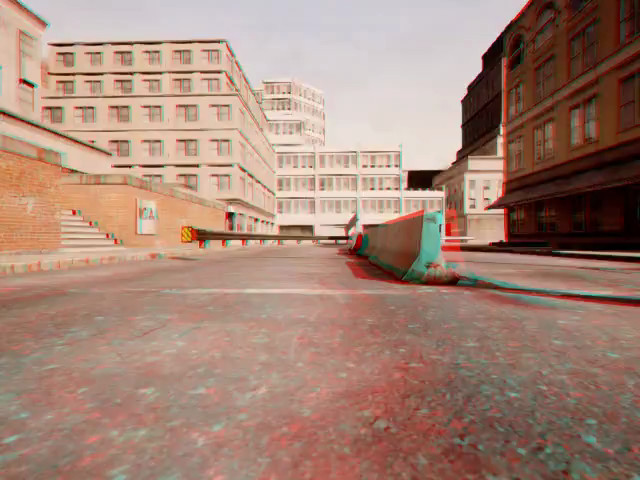} &
    \includegraphics[width=1\linewidth, trim=0 0 0 0,clip]{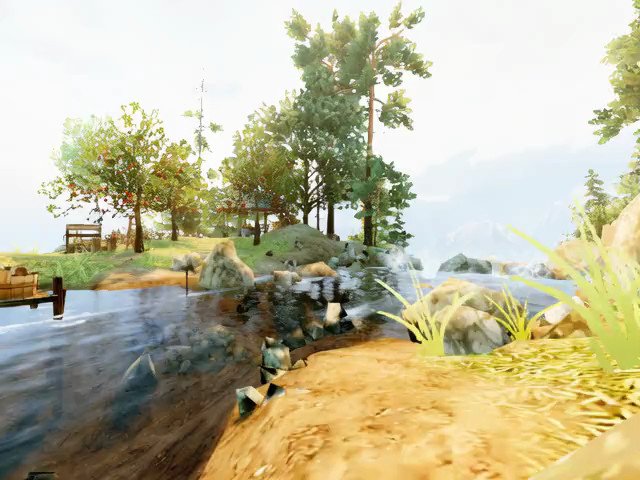} &
    \includegraphics[width=1\linewidth, trim=0 0 0 0,clip]{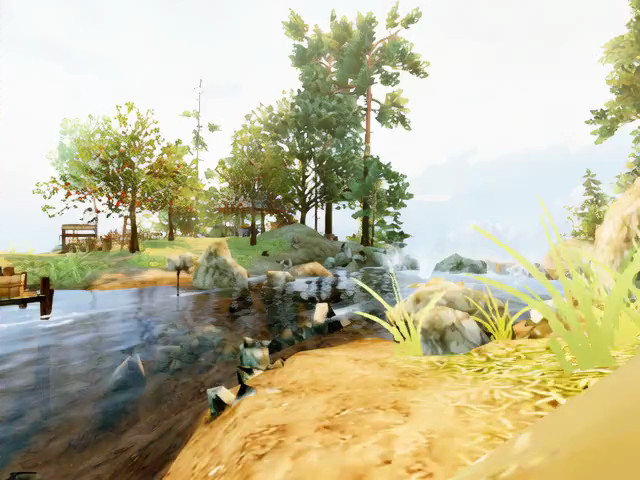} &
    \includegraphics[width=1\linewidth, trim=0 0 0 0,clip]{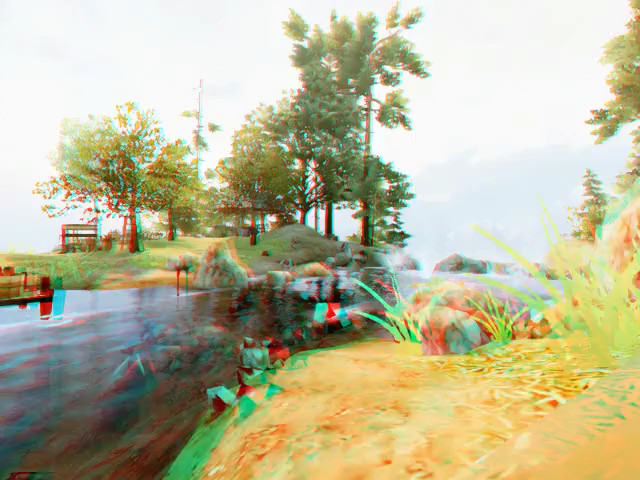}
    \\
    \rotatebox{90}{ \hspace{7mm} \small{t2}} &
    \includegraphics[width=1\linewidth, trim=0 0 0 0,clip]{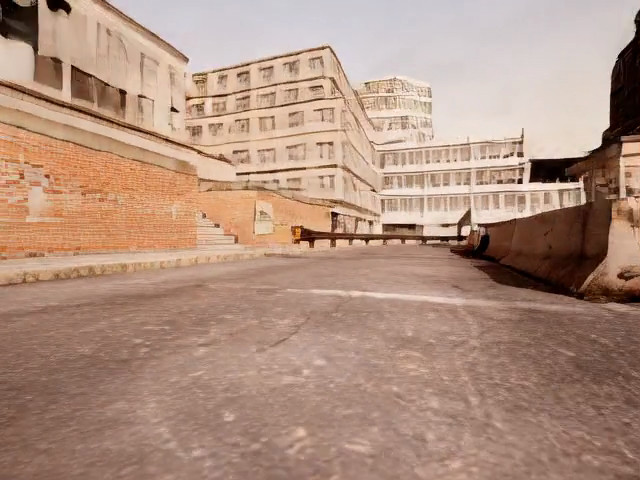} &
    \includegraphics[width=1\linewidth, trim=0 0 0 0,clip]{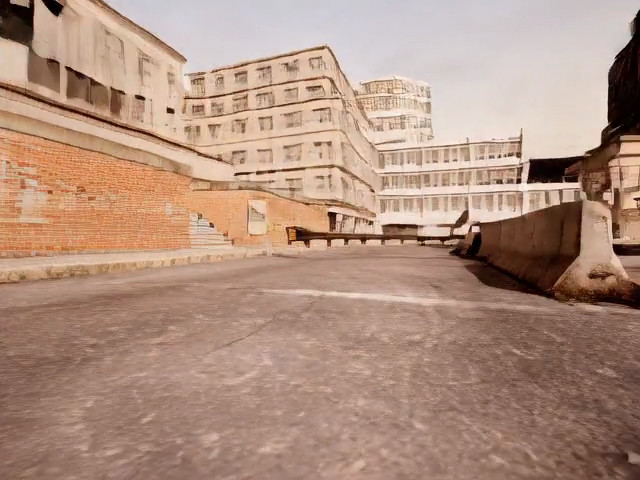} &
    \includegraphics[width=1\linewidth, trim=0 0 0 0,clip]{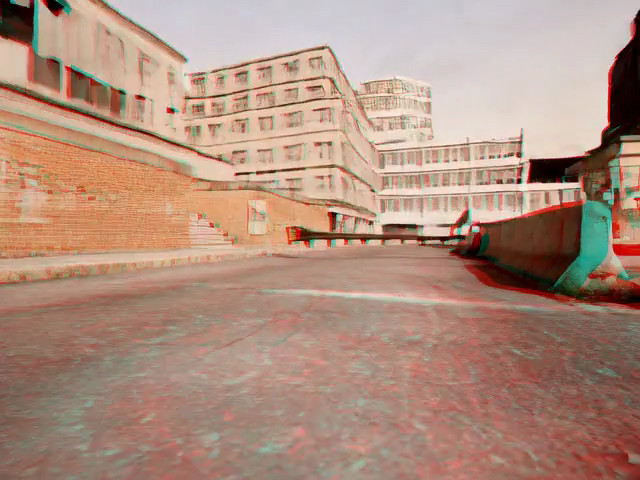} &
    \includegraphics[width=1\linewidth, trim=0 0 0 0,clip]{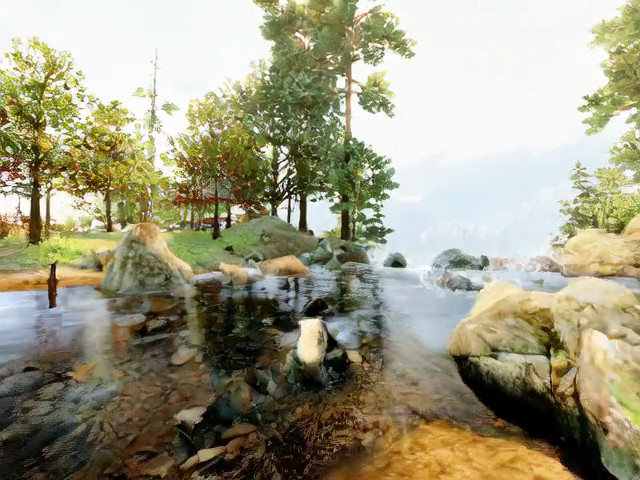} &
    \includegraphics[width=1\linewidth, trim=0 0 0 0,clip]{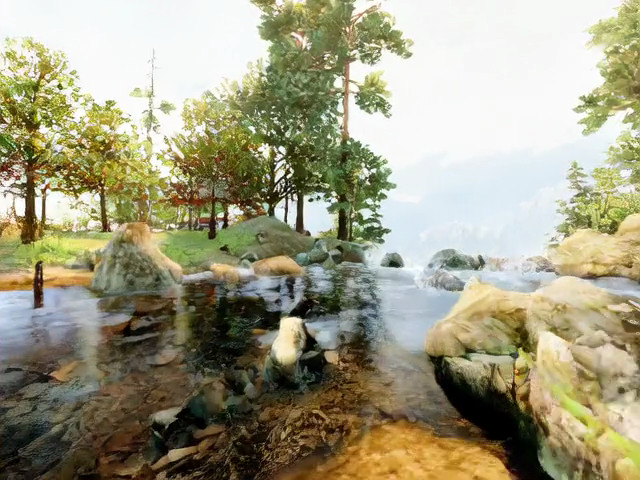} &
    \includegraphics[width=1\linewidth, trim=0 0 0 0,clip]{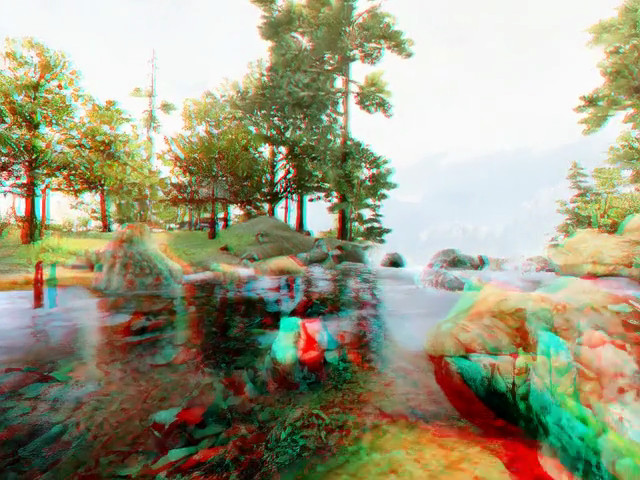}
    \\
    \rotatebox{90}{ \hspace{7mm} \small{t3}} &
    \includegraphics[width=1\linewidth, trim=0 0 0 0,clip]{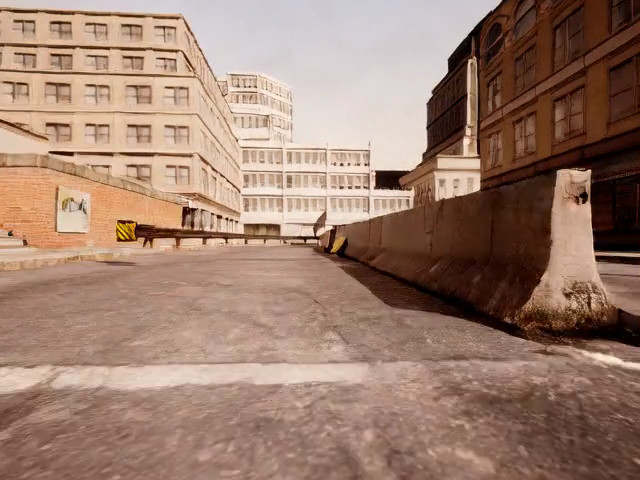} &
    \includegraphics[width=1\linewidth, trim=0 0 0 0,clip]{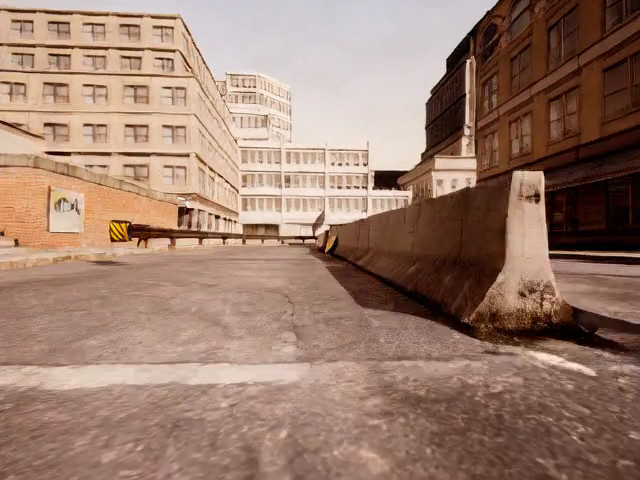} &
    \includegraphics[width=1\linewidth, trim=0 0 0 0,clip]{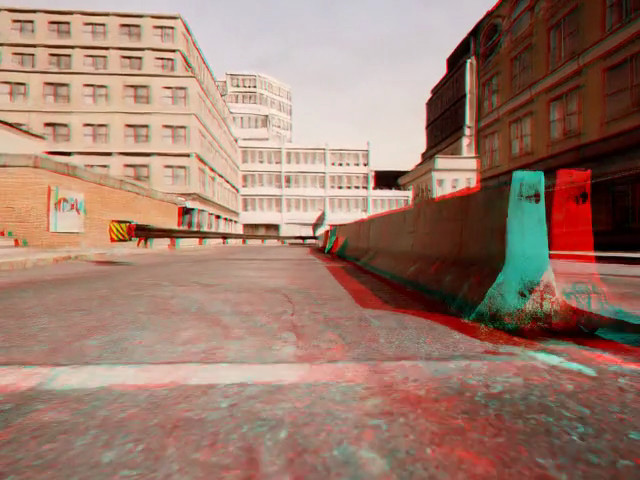} &
    \includegraphics[width=1\linewidth, trim=0 0 0 0,clip]{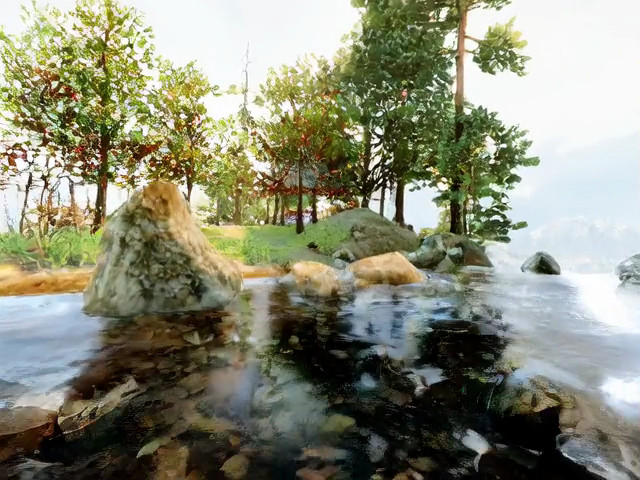} &
    \includegraphics[width=1\linewidth, trim=0 0 0 0,clip]{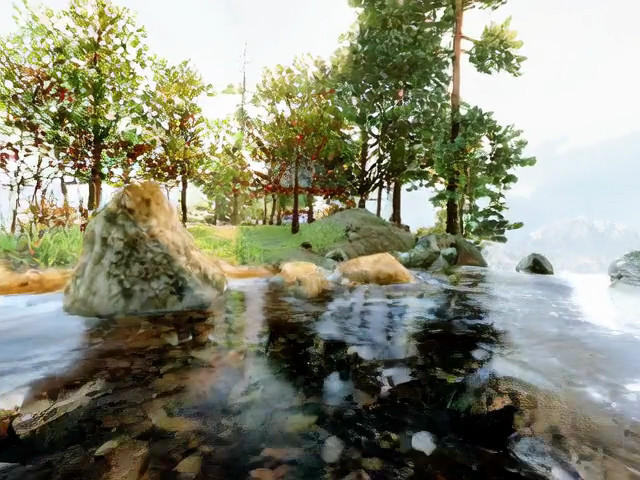} &
    \includegraphics[width=1\linewidth, trim=0 0 0 0,clip]{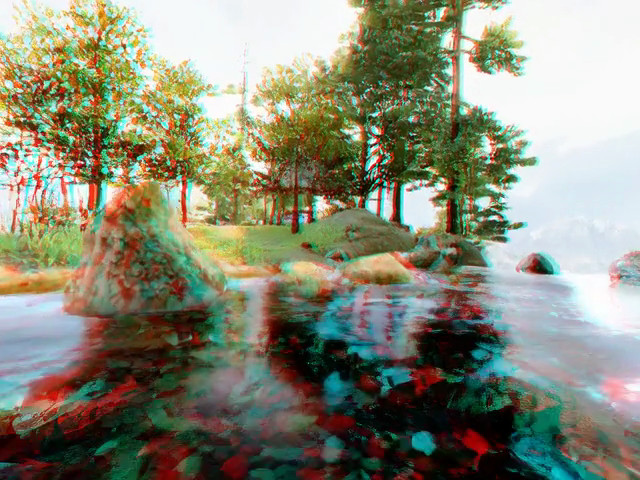}
    
    \end{tabular}
    \end{spacing}
    \vspace{-3mm}
    \caption{More StereoWorld Results with Anaglyph.}
    \label{fig:vr}
\end{figure*}

The binocular videos generated by StereoWorld can be directly utilized in VR/AR applications to deliver immersive experiences. In Fig.~\ref{fig:vr}, we provide additional generated scene examples, together with anaglyph image to demonstrate the diversity and practicality of our approach. We also report the user study results in Fig.~\ref{fig:user_study}, in which we compare our results with baselines in terms of ``Camera Conformity'', ``Temporal Consistency'', ``Image Quality'' and  ``Overall Experience''.

\begin{figure}[t]
    \centering
    \includegraphics[width=0.9\linewidth, trim=50 0 50 0,clip]{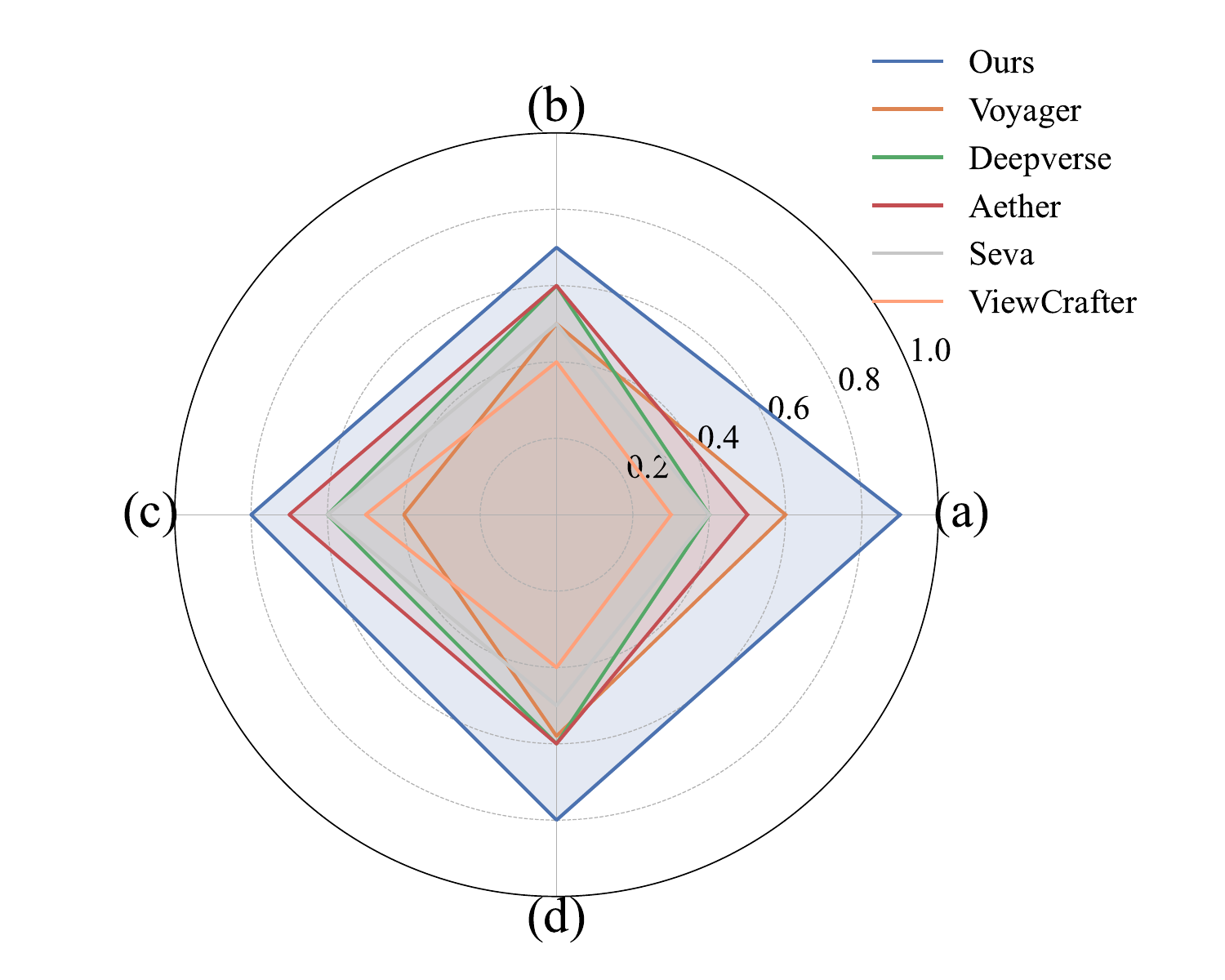}
    \vspace{-3mm}
    \caption{The summary of quantitative feedback in the user study. (a) Camera Conformity (b) Temporal Consistency (c) Image Quality (d) Overall. }
    \vspace{-5mm}
    \label{fig:user_study}
\end{figure}

\subsection{Embodided Scenarios}
\begin{figure}[t]
    \centering
    \setlength{\fboxrule}{0.5pt}
    \setlength{\fboxsep}{-0.01cm}
    \setlength\tabcolsep{0pt}
    \begin{spacing}{1}
    \begin{tabular}{p{0.04\linewidth}<{\centering}p{0.32\linewidth}<{\centering}p{0.32\linewidth}<{\centering}p{0.32\linewidth}<{\centering}}
    
    & Left View & Right View  & Disparity \\
    \rotatebox{90}{ \hspace{5mm} \small{t1}} &
    \includegraphics[width=1\linewidth, trim=0 0 0 0,clip]{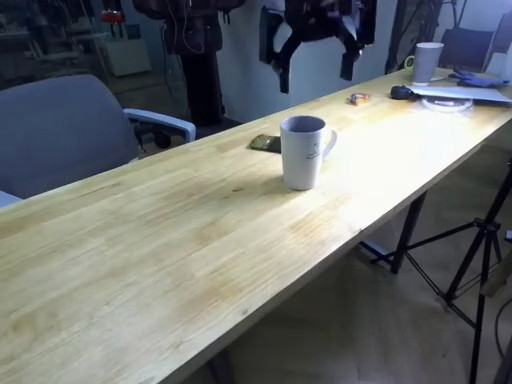} &
    \includegraphics[width=1\linewidth, trim=0 0 0 0,clip]{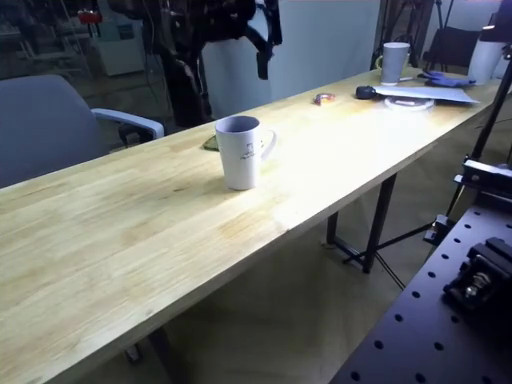} &
    \includegraphics[width=1\linewidth, trim=0 0 0 0,clip]{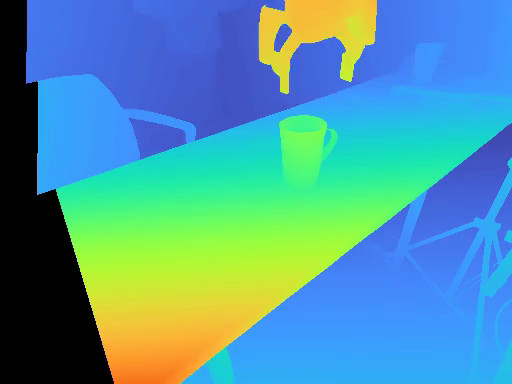}
    \\
    \specialrule{0em}{0pt}{-15pt} \\
    \rotatebox{90}{ \hspace{5mm} \small{t2}} &
    \includegraphics[width=1\linewidth, trim=0 0 0 0,clip]{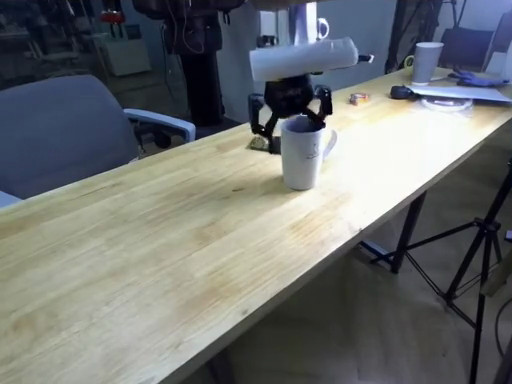} &
    \includegraphics[width=1\linewidth, trim=0 0 0 0,clip]{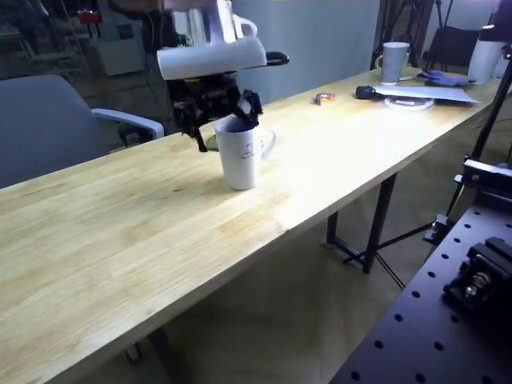} &
    \includegraphics[width=1\linewidth, trim=0 0 0 0,clip]{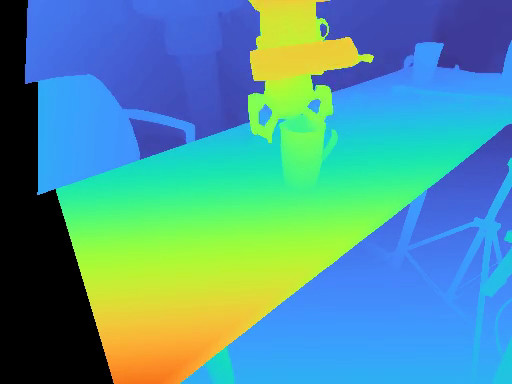}
    \\
    \specialrule{0em}{0pt}{-15pt} \\
    \rotatebox{90}{ \hspace{5mm} \small{t3}} &
    \includegraphics[width=1\linewidth, trim=0 0 0 0,clip]{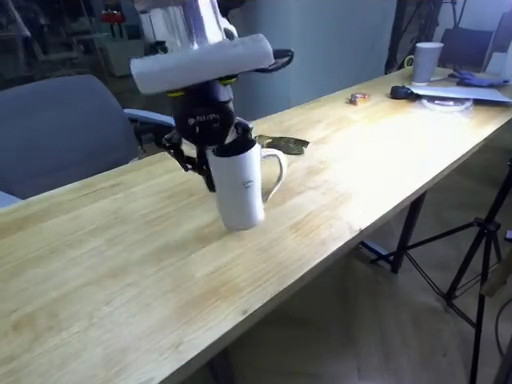} &
    \includegraphics[width=1\linewidth, trim=0 0 0 0,clip]{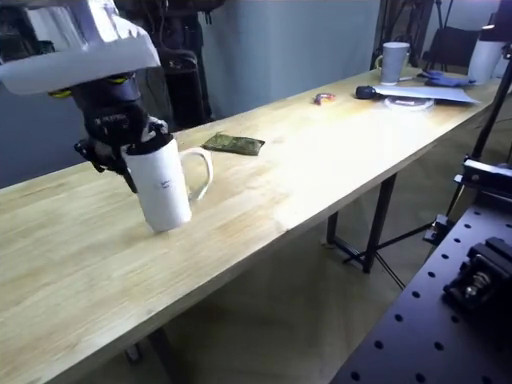} &
    \includegraphics[width=1\linewidth, trim=0 0 0 0,clip]{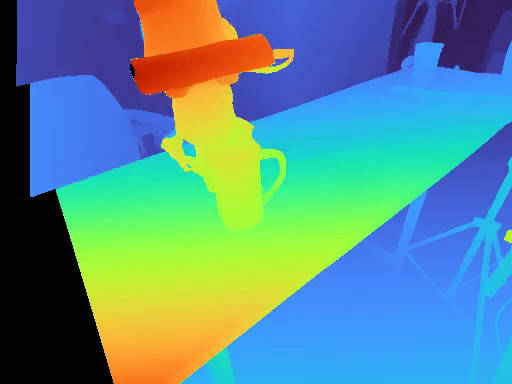} \\
    &  & \textit{``pick up the cup"}  & \\

    \rotatebox{90}{ \hspace{5mm} \small{t1}} &
    \includegraphics[width=1\linewidth, trim=0 0 0 0,clip]{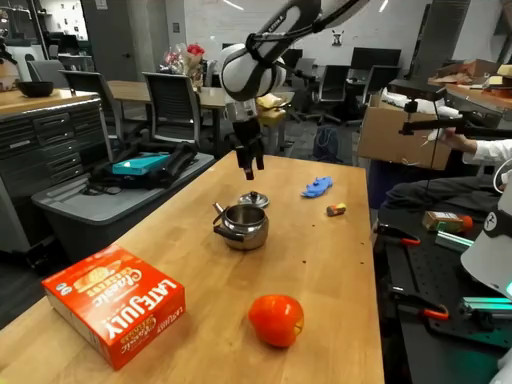} &
    \includegraphics[width=1\linewidth, trim=0 0 0 0,clip]{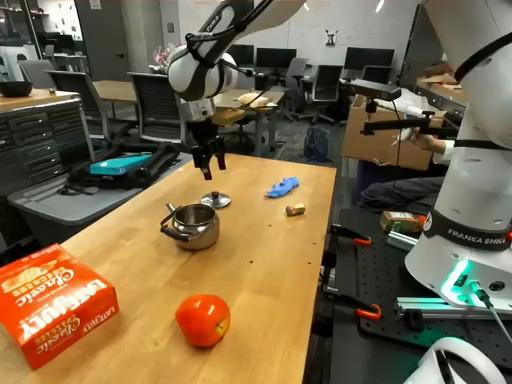} &
    \includegraphics[width=1\linewidth, trim=0 0 0 0,clip]{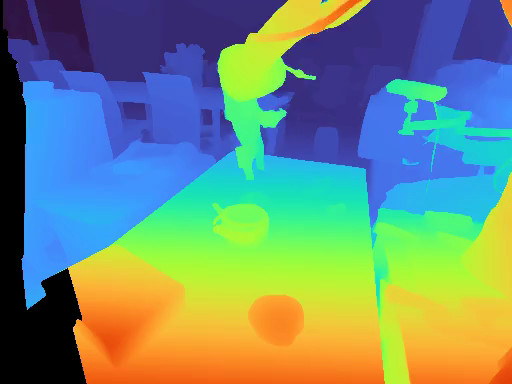}
    \\
    \specialrule{0em}{0pt}{-15pt} \\
    \rotatebox{90}{ \hspace{5mm} \small{t2}} &
    \includegraphics[width=1\linewidth, trim=0 0 0 0,clip]{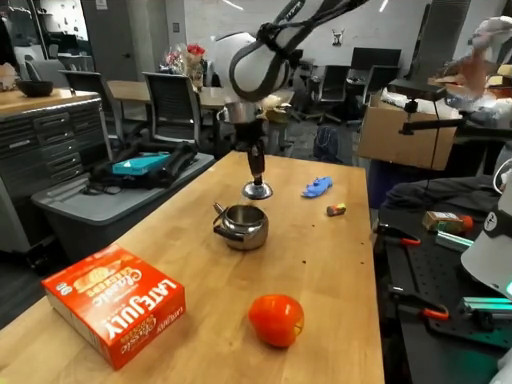} &
    \includegraphics[width=1\linewidth, trim=0 0 0 0,clip]{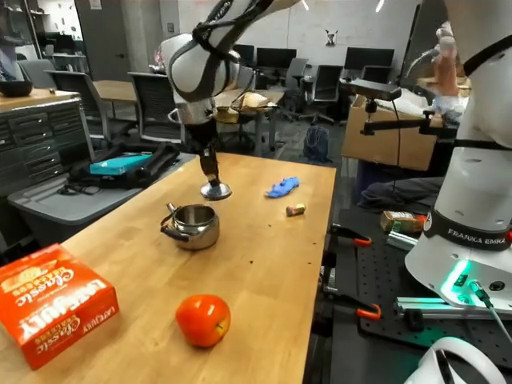} &
    \includegraphics[width=1\linewidth, trim=0 0 0 0,clip]{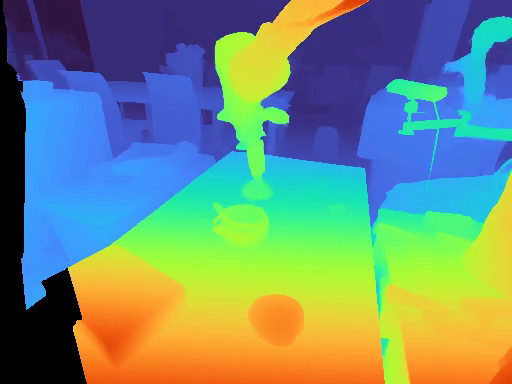}
    \\
    \specialrule{0em}{0pt}{-15pt} \\
    \rotatebox{90}{ \hspace{5mm} \small{t3}} &
    \includegraphics[width=1\linewidth, trim=0 0 0 0,clip]{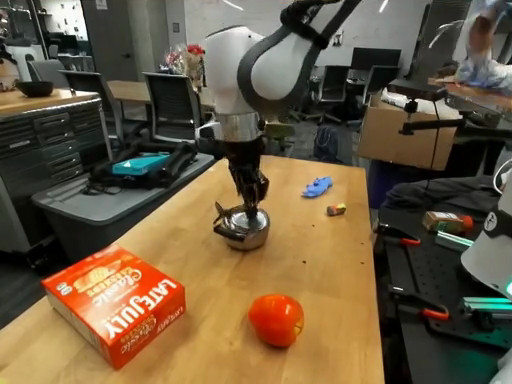} &
    \includegraphics[width=1\linewidth, trim=0 0 0 0,clip]{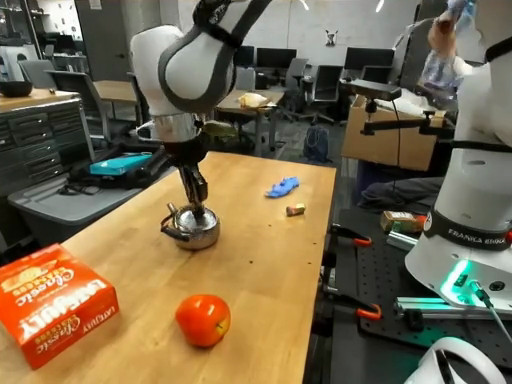} &
    \includegraphics[width=1\linewidth, trim=0 0 0 0,clip]{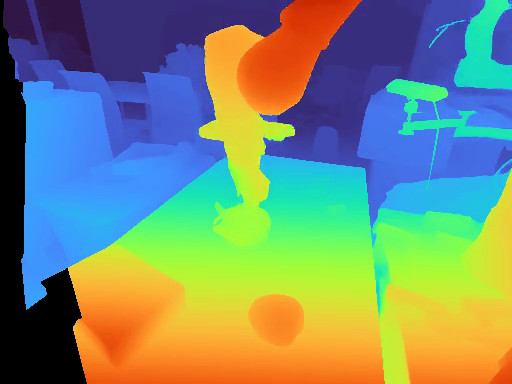} \\
    &  & {\hspace{-13mm}{\textit{``put the lid on the teapot"}}}  &

    \end{tabular}
    \end{spacing}
    \vspace{-3mm}
    \caption{ Stereo Video Generation on Embodided Scenarios.}
    \label{fig:embodided scenarios}
\end{figure}
By fine-tuning our model on binocular robotic arm datasets~\cite{khazatsky2024droid}, our approach can also be applied to embodied scenarios for stereo video generation, supporting downstream tasks such as action planning. As shown in Fig.~\ref{fig:embodided scenarios}, given an action command and the initial stereo frame, our model is able to generate the corresponding subsequent motion sequence. The results demonstrate that the generated videos not only follow the specified action instructions but also maintain high stereo consistency between the left and right views. We further performed disparity estimation on the generated outputs to verify their geometric plausibility and assess their feasibility for action planning.

\subsection{Long Video Distillation}
Our trained model employs a bidirectional attention mechanism, which limits it to relatively short video sequences (49 frames in our setting). In contrast, autoregressive video generation models~\cite{yin2025causvid, huang2025selfforcing} can effectively overcome this limitation and improve efficiency through a rolling KV-cache mechanism.
Inspired by these advancements, we further distill StereoWorld into an autoregressive binocular video generation model, enabling long-horizon video synthesis and improving generation speed.

Following Self-Forcing~\cite{huang2025selfforcing}, we adopt a two-stage paradigm. In the first stage (ODE distillation), we replace the bidirectional attention with a causal attention mechanism and distill the denoising process into four steps. The attention mask is illustrated in Fig~\ref{fig:distillation}, which generates two views at one step. 
In the second stage~\cite{huang2025selfforcing}, we condition each pair of stereo frame’s (or chunks in practice) generation on previously self-generated outputs by performing autoregressive rollout with
KV-cache. In this stage, a distribution matching distillation~\cite{yin2024one} (DMD loss) is applied to address exposure bias via distribution matching.
Unlike monocular autoregressive video generation, our method simultaneously synthesizes binocular views and incorporates camera pose–aware positional encoding. As a result, the KV-cache must be updated with two separate sets of keys and values at each step -- one for the left-view tokens and one for the right-view tokens, each containing our \emph{Unified Camera-Frame RoPE}.

\begin{figure}[t]
    \centering
    \setlength{\fboxrule}{0.5pt}
    \setlength{\fboxsep}{-0.01cm}
    \setlength\tabcolsep{0pt}
    \begin{spacing}{1}
    \begin{tabular}{p{0.04\linewidth}<{\centering}p{0.24\linewidth}<{\centering}p{0.24\linewidth}<{\centering}p{0.24\linewidth}<{\centering}p{0.24\linewidth}<{\centering}}
    & Left View & Right View  & Left View & Right View \\
    \rotatebox{90}{ \hspace{1mm} \small{frame 1}} &
    \includegraphics[width=1\linewidth, trim=0 0 0 0,clip]{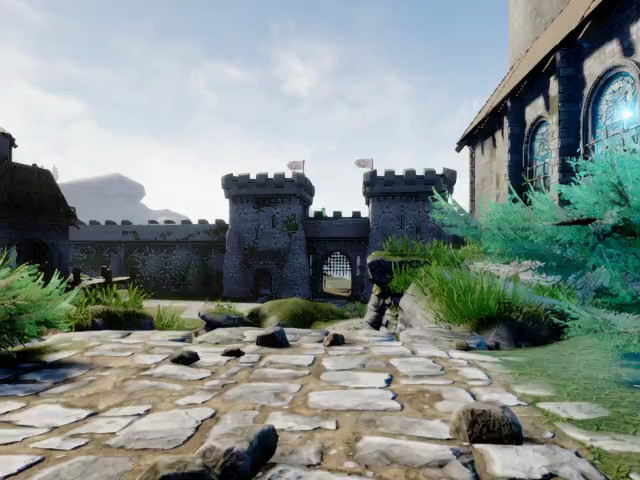} &
    \includegraphics[width=1\linewidth, trim=0 0 0 0,clip]{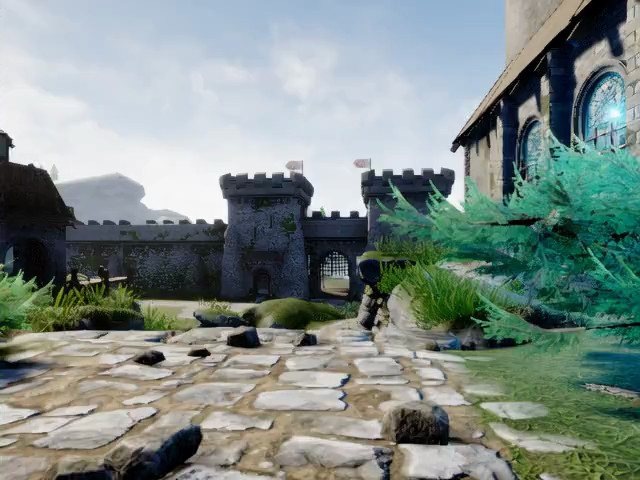} &
    \includegraphics[width=1\linewidth, trim=0 0 0 0,clip]{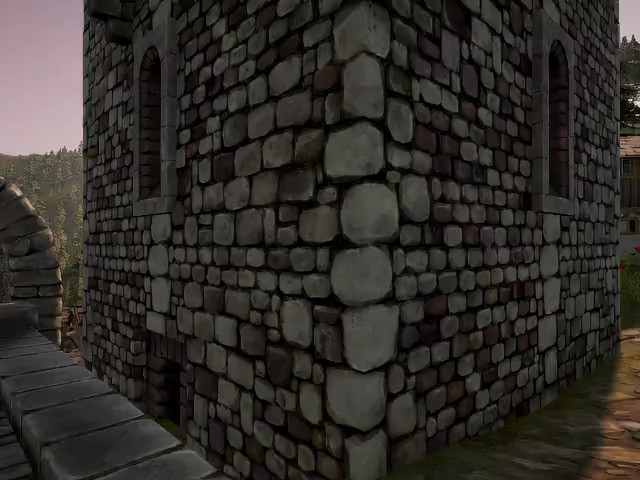} &
    \includegraphics[width=1\linewidth, trim=0 0 0 0,clip]{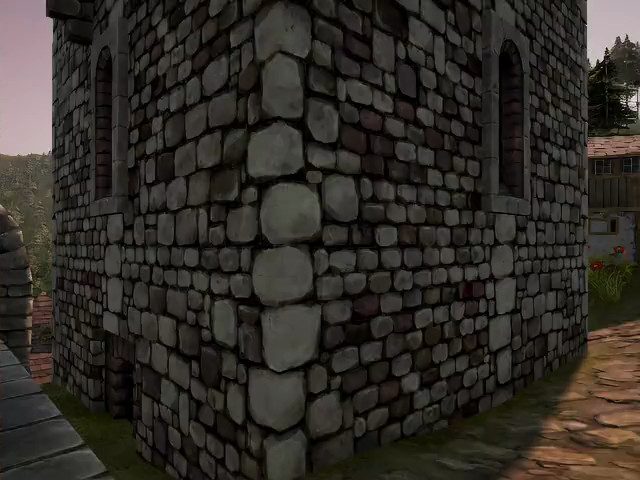} \\
    \specialrule{0em}{0pt}{-15pt} \\
    \rotatebox{90}{ \hspace{0mm} \small{frame 40}} &
    \includegraphics[width=1\linewidth, trim=0 0 0 0,clip]{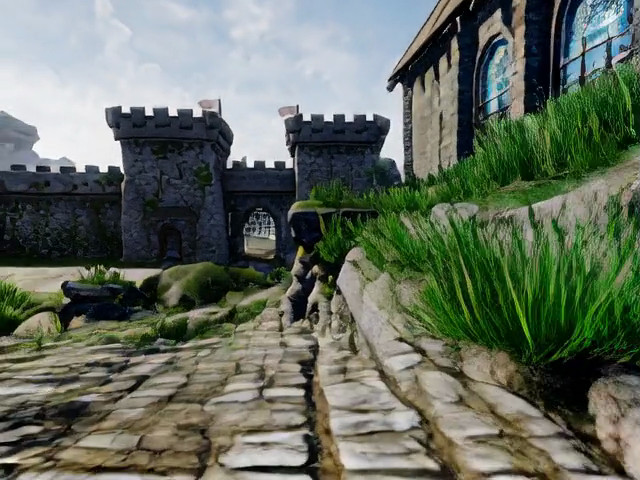} &
    \includegraphics[width=1\linewidth, trim=0 0 0 0,clip]{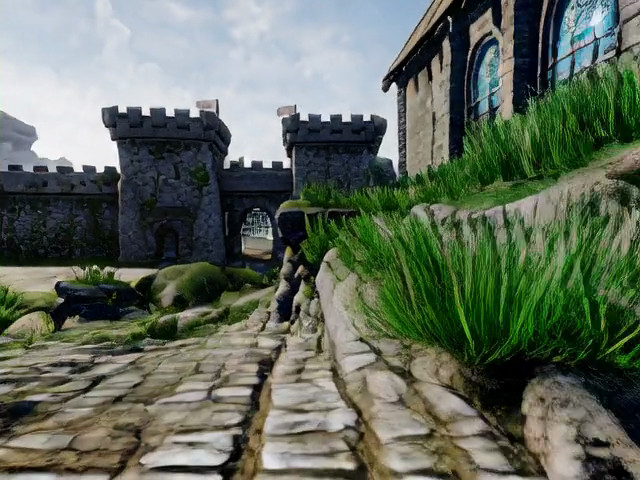} &
    \includegraphics[width=1\linewidth, trim=0 0 0 0,clip]{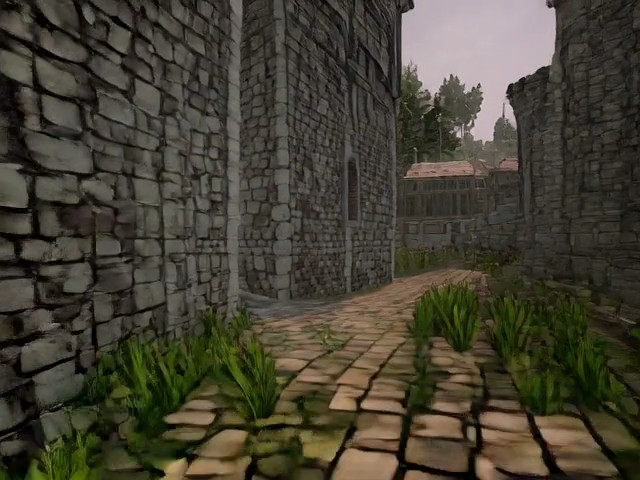} &
    \includegraphics[width=1\linewidth, trim=0 0 0 0,clip]{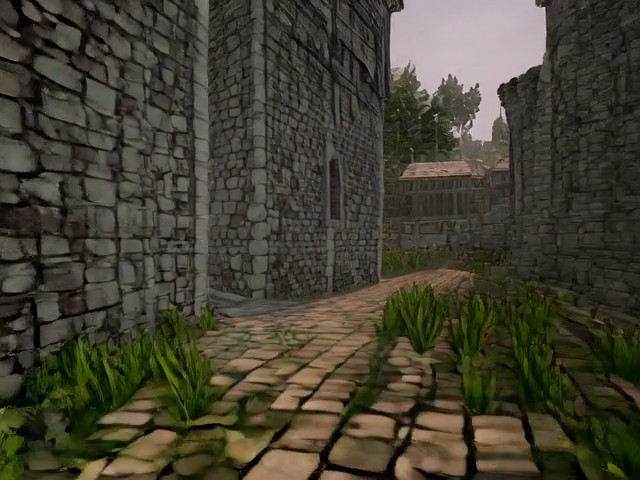} \\
    \specialrule{0em}{0pt}{-15pt} \\
    \rotatebox{90}{ \hspace{0mm} \small{frame 80}} &
    \includegraphics[width=1\linewidth, trim=0 0 0 0,clip]{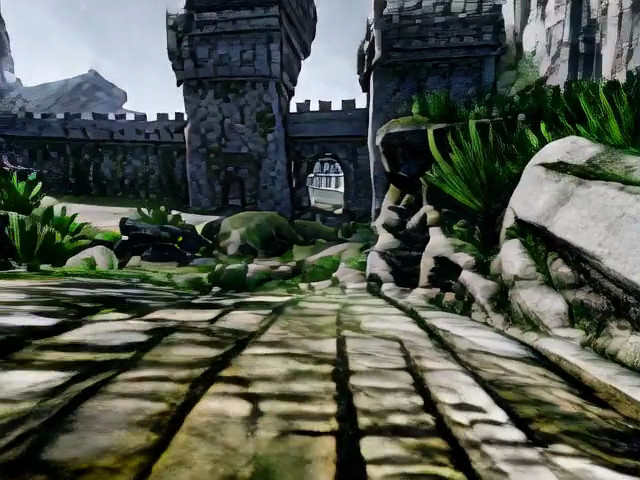} &
    \includegraphics[width=1\linewidth, trim=0 0 0 0,clip]{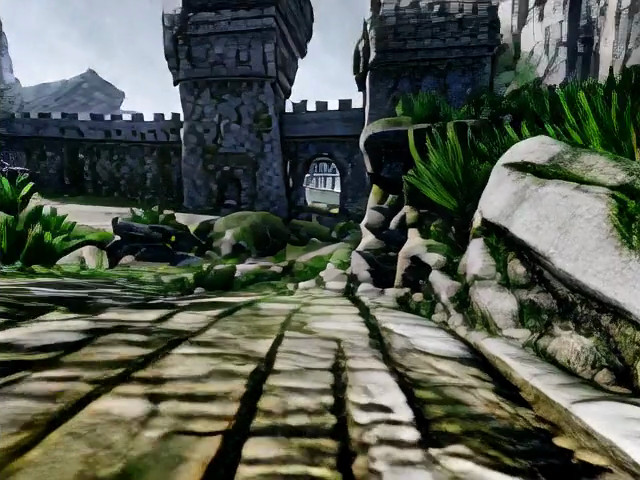} &
    \includegraphics[width=1\linewidth, trim=0 0 0 0,clip]{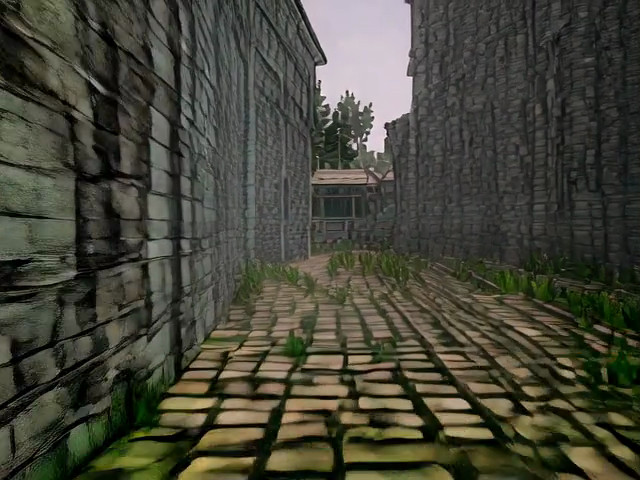} &
    \includegraphics[width=1\linewidth, trim=0 0 0 0,clip]{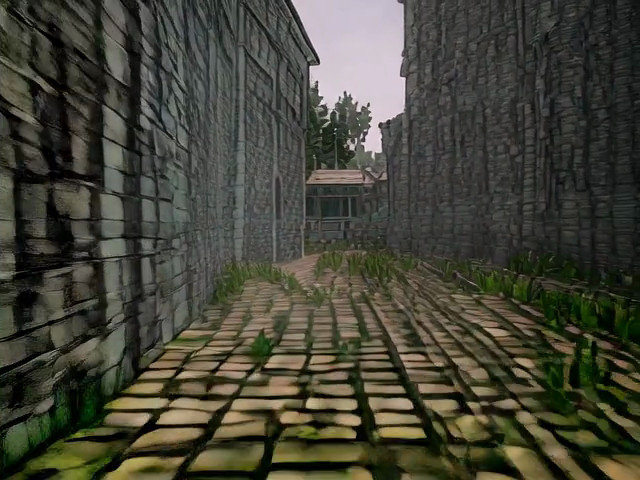} \\
    \specialrule{0em}{0pt}{-15pt} \\
    \rotatebox{90}{ \hspace{-1mm} \small{frame 120}} &
    \includegraphics[width=1\linewidth, trim=0 0 0 0,clip]{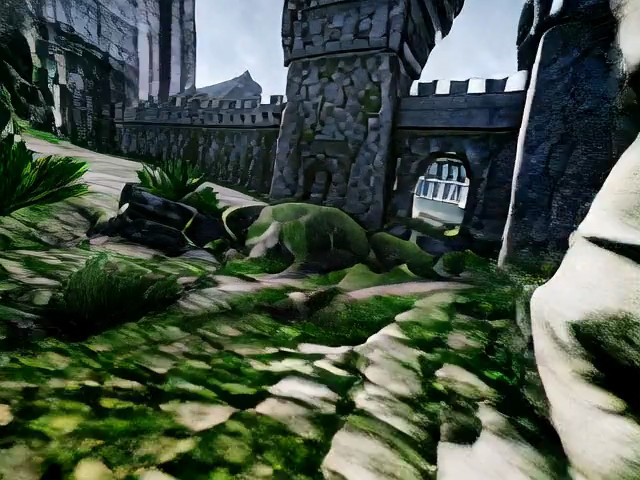} &
    \includegraphics[width=1\linewidth, trim=0 0 0 0,clip]{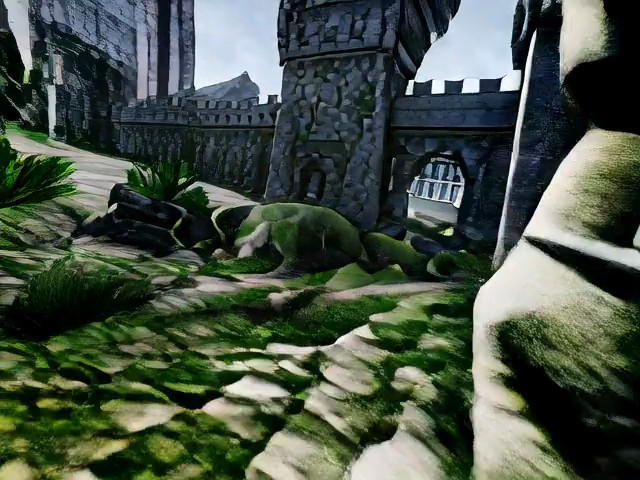} &
    \includegraphics[width=1\linewidth, trim=0 0 0 0,clip]{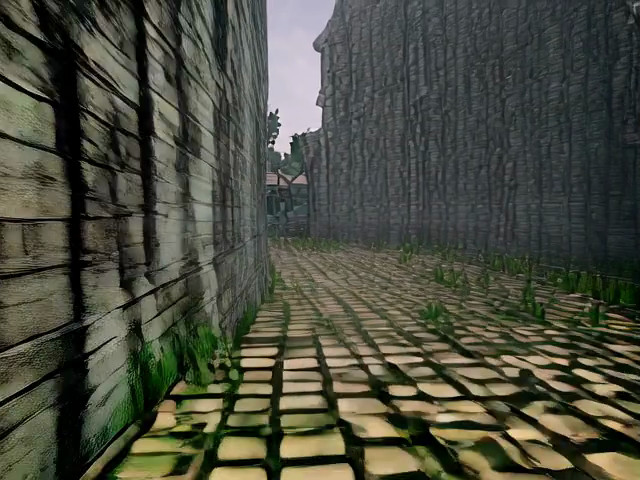} &
    \includegraphics[width=1\linewidth, trim=0 0 0 0,clip]{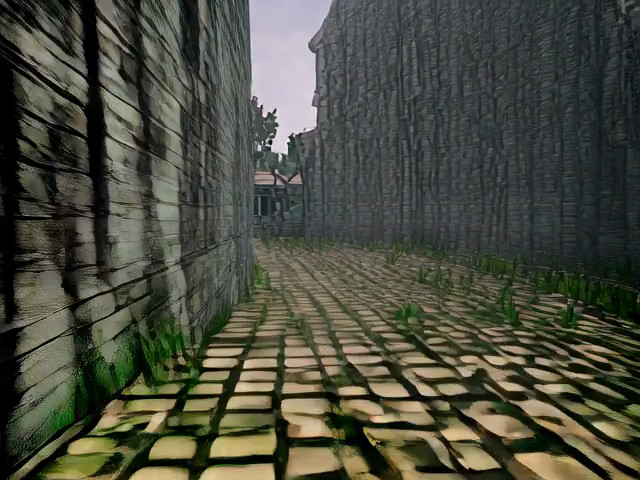} \\
    \specialrule{0em}{0pt}{-15pt} \\
    \rotatebox{90}{ \hspace{-1mm} \small{frame 160}} &
    \includegraphics[width=1\linewidth, trim=0 0 0 0,clip]{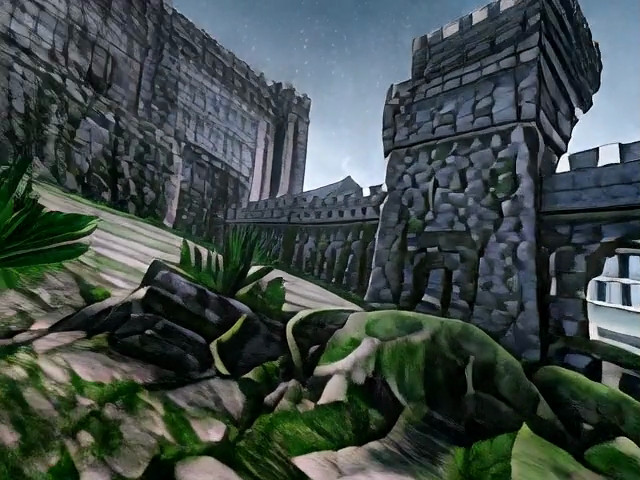} &
    \includegraphics[width=1\linewidth, trim=0 0 0 0,clip]{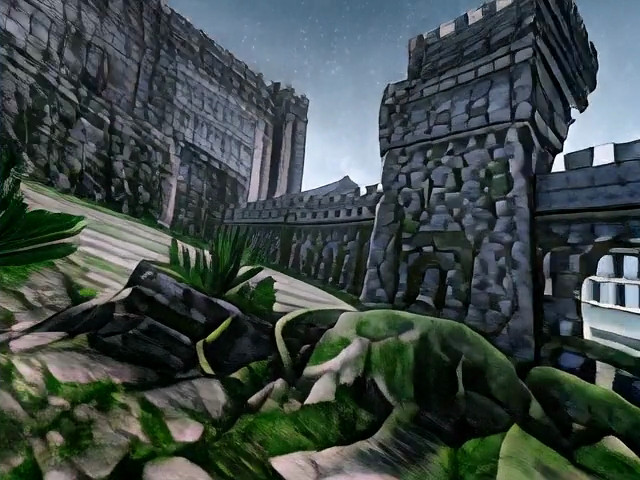} &
    \includegraphics[width=1\linewidth, trim=0 0 0 0,clip]{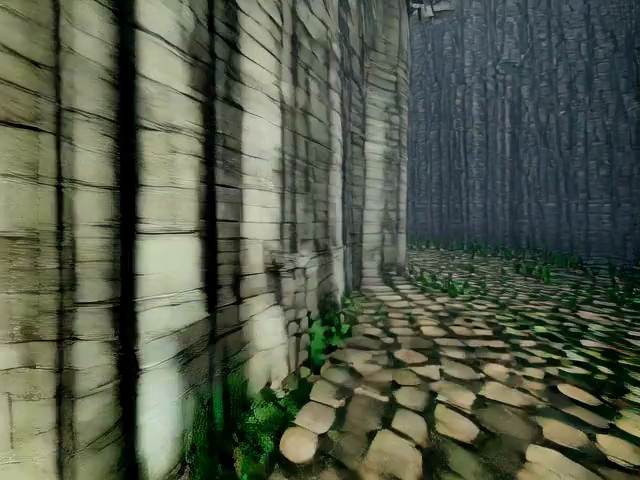} &
    \includegraphics[width=1\linewidth, trim=0 0 0 0,clip]{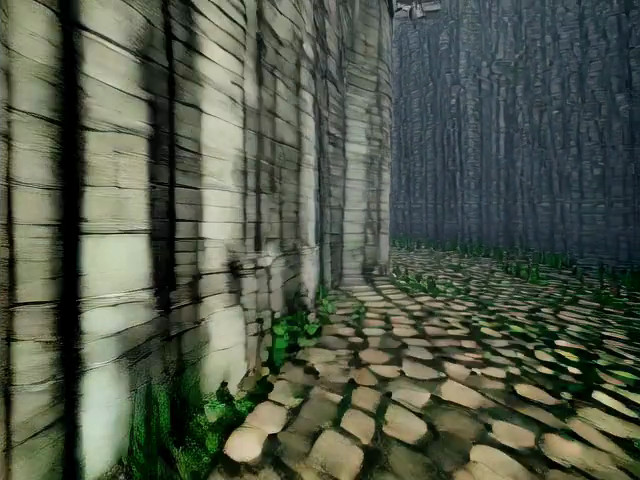} \\
    \specialrule{0em}{0pt}{-15pt} \\
    \rotatebox{90}{ \hspace{-1mm} \small{frame 192}} &
    \includegraphics[width=1\linewidth, trim=0 0 0 0,clip]{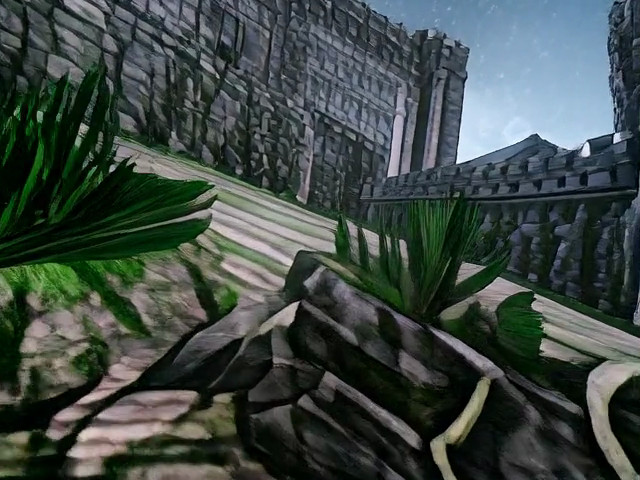} &
    \includegraphics[width=1\linewidth, trim=0 0 0 0,clip]{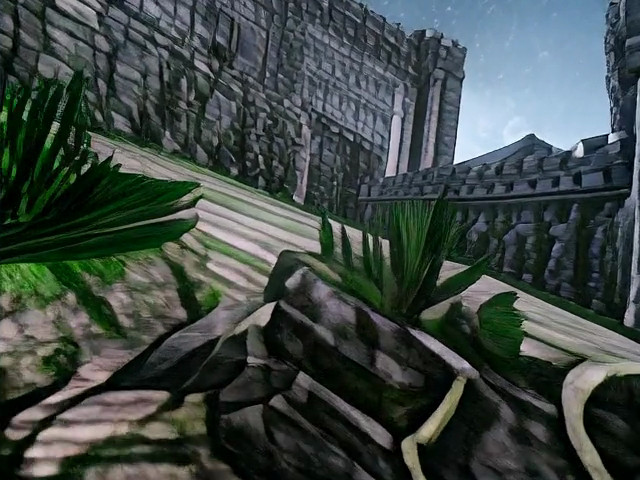} &
    \includegraphics[width=1\linewidth, trim=0 0 0 0,clip]{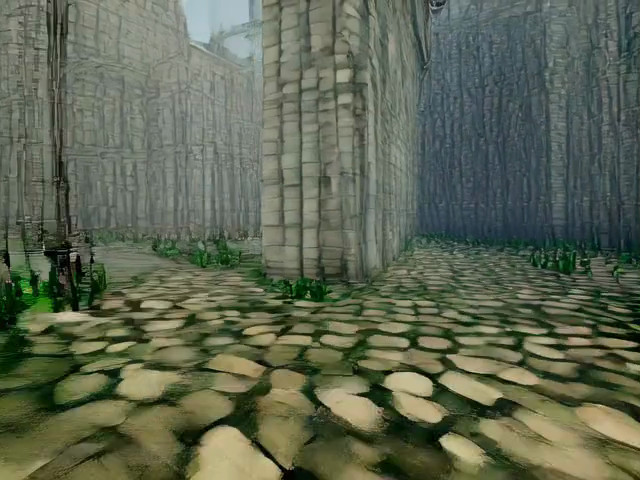} &
    \includegraphics[width=1\linewidth, trim=0 0 0 0,clip]{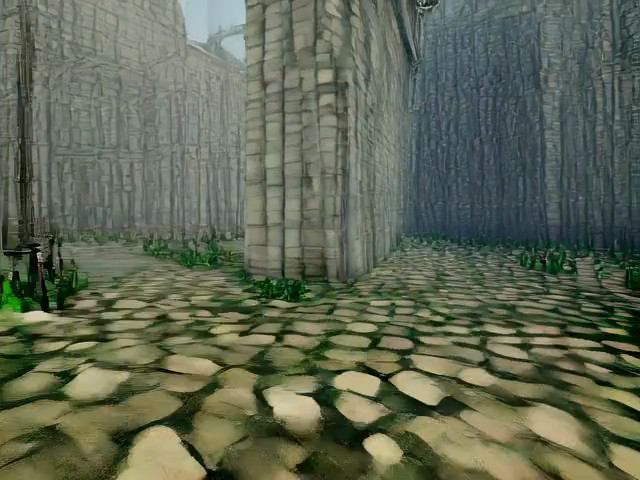} \\
    \end{tabular}
    \end{spacing}
    \vspace{-3mm}
    \caption{ Long Video Distillation Results.}
    \label{fig:long-video}
\end{figure}
The distilled model achieves a significant improvement in binocular video generation speed, increasing from 0.49 FPS to 5 FPS, and is no longer limited to generating video clips of 49 frames. We present the results of long-video distillation in Fig~\ref{fig:long-video}, and in the supplementary video materials.


However, we observe that as the video length increases, the generated results still exhibit noticeable degradation. This issue is also present in prior works such as Self-Forcing. Improving the stability of long-horizon video generation therefore remains an open challenge shared by both monocular and stereo video synthesis.

\section{Monocular \& Stereo Generation Comparison.}
``Ours Monocular" and ``Ours Stereo" in Tab~\ref{tab:stereo_video_comparison} employ the exact same parameter count and compute budget. 
The superior FID for ``Ours Stereo'' is because binocular views provide a physical ``anchor''. As demonstrated below (Fig~\ref{fig:monocular_stereo_comparison}) monocular pipelines relies on a single condition frame and often hallucinate unrealistic structures due to occlusion, whereas stereo settings incorporates additional view and better maintains alignment with real scene by stereo-aware attention.

\begin{figure}[thb]
    \vspace{-3mm}
    \centering
    \includegraphics[width=1\linewidth, trim=0 0 0 0,clip]{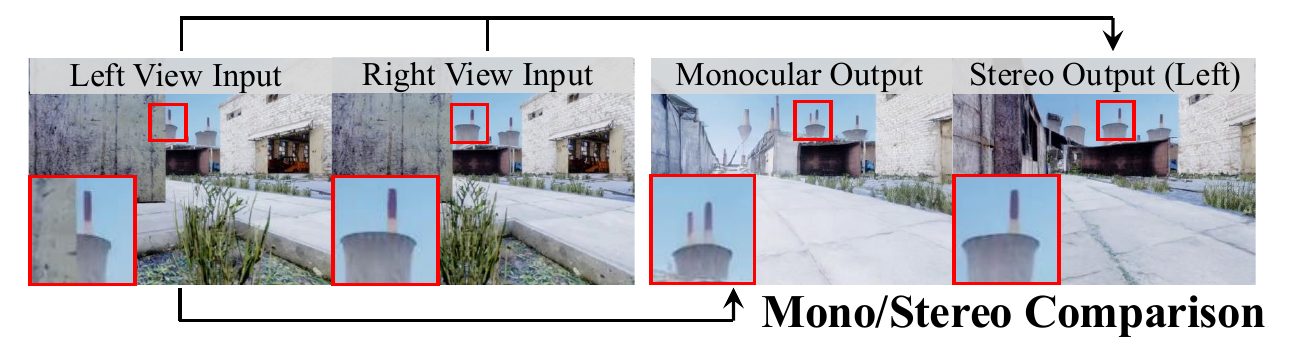}
      \vspace{-3mm}
    \caption{Monocular and stereo generation comparison.}
    \label{fig:monocular_stereo_comparison}
\end{figure}

\section{Large \& Varying Baselines.} 
To evaluate the model's performance under varying baselines, we construct a camera trajectory by expanding the right camera baseline from 0.25m to 0.75m-- well beyond the training distribution (0.063m-0.25m). As illustrated below (Fig~\ref{fig:baseline_effect}), StereoWorld maintains geometric plausibility and achieves accurate metric-scale recovery up to 0.42m, outperforming SOTA like DepthAnything V2. This confirms our Unified Camera-Frame RoPE performs genuine geometric reasoning rather than simple image stretching, also demonstrating robust generalization to unseen camera trajectories and baseline configurations.

\begin{figure}[htb]
    \vspace{-3mm}
    \centering
    \includegraphics[width=1\linewidth, trim=0 0 0 0,clip]{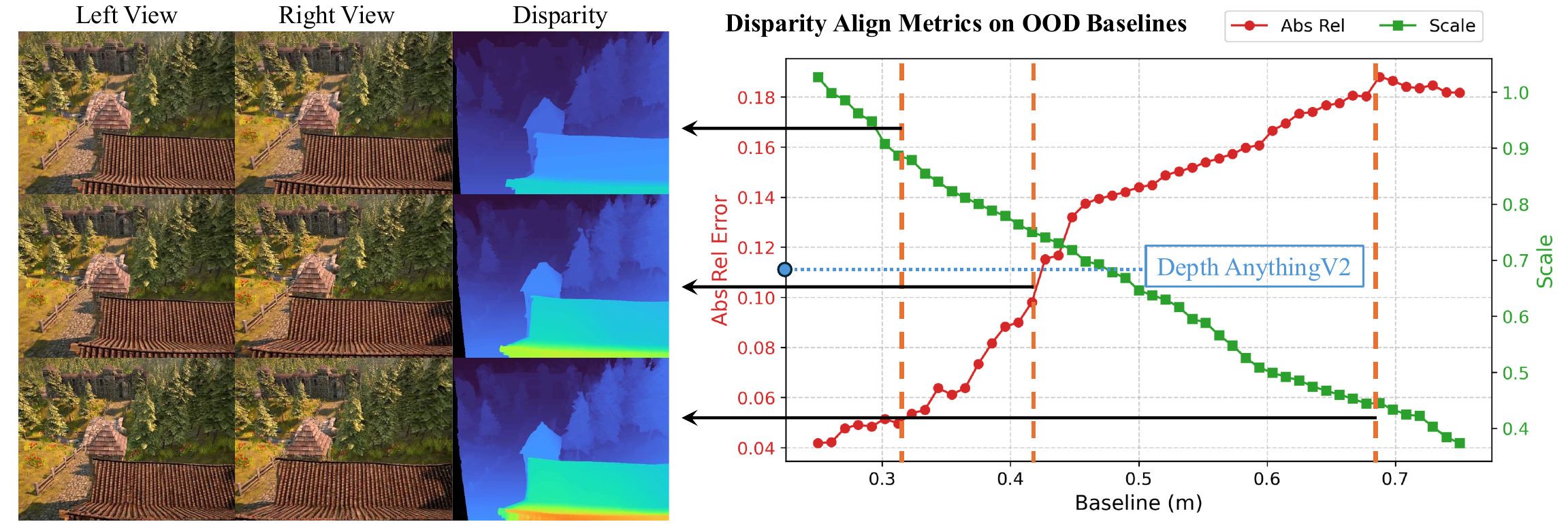}
    \caption{Effect of different baselines on StereoWorld.}
    \label{fig:baseline_effect}
\end{figure}

\section{Discussion}

\begin{figure}[t]
    \centering
    \setlength{\fboxrule}{0.5pt}
    \setlength{\fboxsep}{-0.01cm}
    \setlength\tabcolsep{0pt}
    \begin{spacing}{1}
    \begin{tabular}{p{0.04\linewidth}<{\centering}p{0.32\linewidth}<{\centering}p{0.32\linewidth}<{\centering}p{0.32\linewidth}<{\centering}}
    
    & t$_1$ & t$_2$  & t$_3$ \\
    \rotatebox{90}{ \hspace{1mm} \small{Left View}} &
    \includegraphics[width=1\linewidth, trim=0 0 0 0,clip]{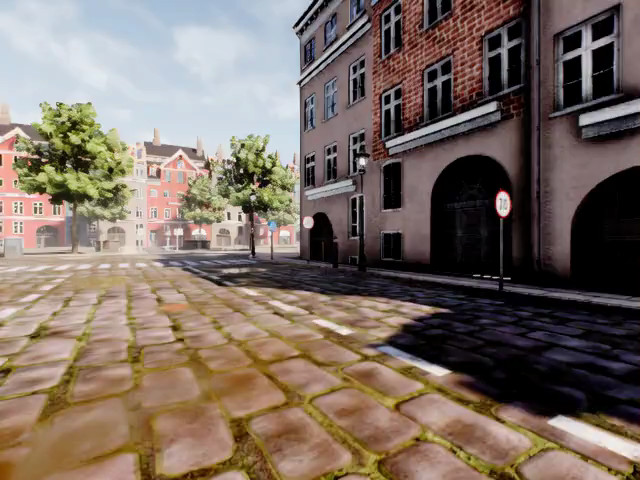} &
    \includegraphics[width=1\linewidth, trim=0 0 0 0,clip]{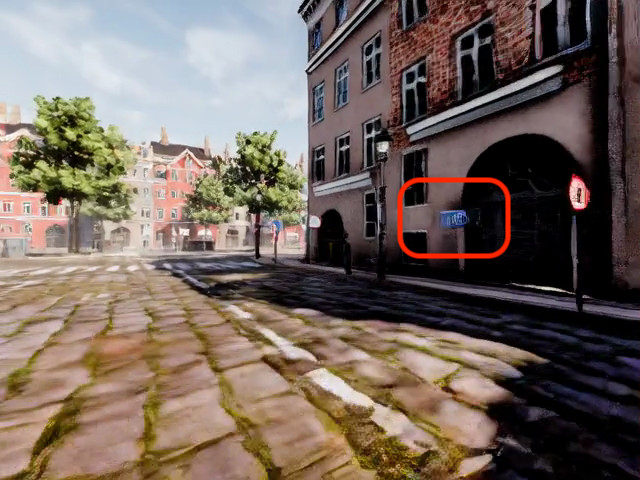} &
    \includegraphics[width=1\linewidth, trim=0 0 0 0,clip]{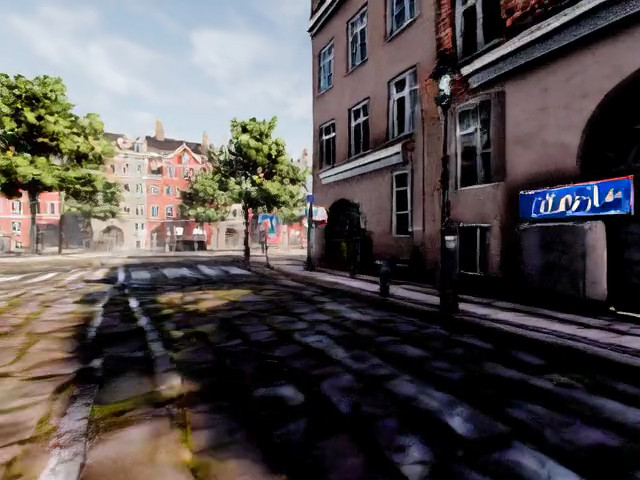}
    \\
    \rotatebox{90}{ \hspace{1mm} \small{Right View}} &
    \includegraphics[width=1\linewidth, trim=0 0 0 0,clip]{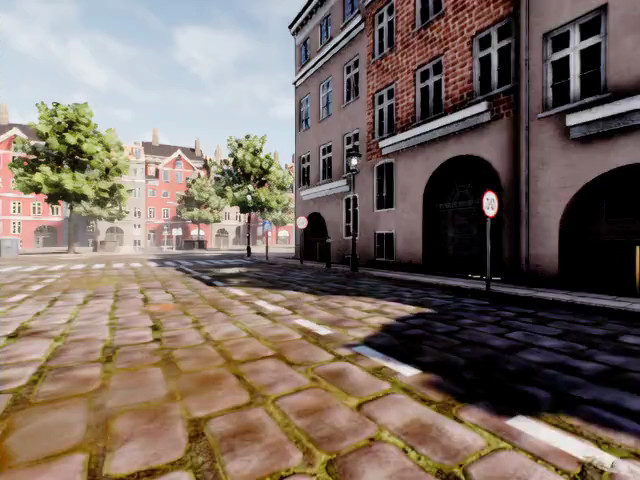} &
    \includegraphics[width=1\linewidth, trim=0 0 0 0,clip]{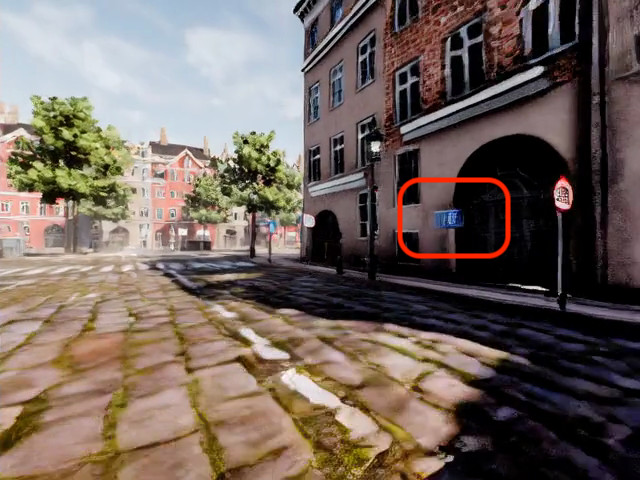} &
    \includegraphics[width=1\linewidth, trim=0 0 0 0,clip]{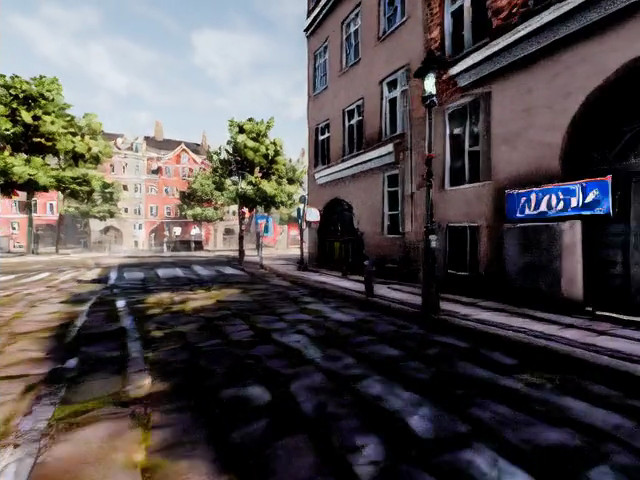}
    \\
    \end{tabular}
    \end{spacing}
    \vspace{-3mm}
    \caption{Failure Case. Note that the blue road sign does not exist at the beginning of the sequence; however, as the viewpoint advances, it gradually emerges and increases in size.}
    \label{fig:failure cases}
\end{figure}

Our method currently does not incorporate any explicit constraints on scene-level consistency. Although it handles most cases well, certain examples may still exhibit spatial inconsistencies across video frames, as illustrated in Fig~\ref{fig:failure cases}. This issue may be alleviated by introducing a spatial memory mechanism~\cite{li2025vmem, wu2025spmem}. Since stereo video generation inherently provides geometric information about the scene, our approach can be naturally integrated with methods such as VMem~\cite{li2025vmem} or SPMem~\cite{wu2025spmem}, replacing their additional reconstruction modules and maintaining consistency through a dedicated spatial memory.


We also note that our method predominantly generates static scenes. This is primarily due to the limited availability of binocular video data for training stereo models. Most of our training corpus consists of static, rendered scenes, which restricts the model’s ability to synthesize dynamic environments. Exploring strategies for collecting more dynamic stereo video data, or leveraging richer monocular dynamic video datasets, represents a highly promising direction for future work. Scaling the training to substantially larger datasets may also help mitigate the aforementioned consistency issues.

Moreover, since the stereo world model generates binocular videos simultaneously, it inherently models fewer frames compared to monocular methods. Although distillation into autoregressive frameworks enables the generation of longer videos, we still observe noticeable degradation in the later stages of video generation, similar as reported in self-forcing~\cite{huang2025selfforcing}. Developing approaches to robustly distill stereo video models into long-term video generators will therefore be a key focus for our future work.

\end{document}